\documentclass{article}

\PassOptionsToPackage{numbers, compress}{natbib}
\usepackage[preprint]{neurips_2026}

\usepackage[utf8]{inputenc} 
\usepackage[T1]{fontenc}    
\usepackage{hyperref}       
\usepackage{url}            
\usepackage{booktabs}       
\usepackage{amsfonts}       
\usepackage{nicefrac}       
\usepackage{microtype}      
\usepackage{xcolor}         
\usepackage{graphicx}
\usepackage{wrapfig}
\usepackage{colortbl}
\usepackage{multirow}
\usepackage{array}
\usepackage[most]{tcolorbox}
\usepackage{microtype}
\usepackage{graphicx}
\usepackage{subcaption}
\usepackage{booktabs}
\usepackage{tabularx}
\usepackage{array}
\usepackage[table]{xcolor} 
\usepackage{xcolor}
\usepackage{colortbl}
\usepackage{amssymb}
\usepackage{multirow}
\usepackage{pgf}
\usepackage{enumitem}

\title{VISTA-Bench: Do Vision-Language Models 
  Really Understand Visualized Text as Well as Pure Text?}

\author{%
  {\bfseries
  Qing'an Liu$^{1*}$,
  Juntong Feng$^{1*}$,
  Yuhao Wang$^{1*}$,
  Xinzhe Han$^{1}$,
  Yujie Cheng$^{1}$,
  Yue Zhu$^{1}$}\\
  {\bfseries
  Haiwen Diao$^{2\dagger}$,
  Yunzhi Zhuge$^{1\dagger}$,
  Huchuan Lu$^{1}$}\\[0.35em]
  {\normalfont
  $^{1}$Dalian University of Technology,
  $^{2}$Nanyang Technological University}
}

\begin{document}

\maketitle

\begingroup
\renewcommand{\thefootnote}{}
\footnotetext{$^{*}$Equal contribution. $^{\dagger}$Correspondence to \texttt{haiwen.diao@ntu.edu.sg}, \texttt{zgyz@dlut.edu.cn}.}
\endgroup

\begin{abstract}
  Vision–Language Models (VLMs) have achieved impressive performance in cross-modal understanding across textual and visual inputs, yet existing benchmarks predominantly focus on pure-text queries. In real-world scenarios, language also frequently appears as visualized text embedded in images, raising the question of whether current VLMs handle such input requests comparably.
    We introduce \textbf{VISTA-Bench}, a systematic benchmark from multimodal perception, reasoning, to unimodal understanding domains. It evaluates visualized text understanding by contrasting pure-text and visualized-text questions under controlled rendering conditions. Extensive evaluation of over 30 representative VLMs reveals a pronounced modality gap: models that perform well on pure-text queries often degrade substantially when equivalent semantic content is presented as visualized text. This gap is further amplified by increased perceptual difficulty, highlighting sensitivity to rendering variations despite unchanged semantics. Overall, VISTA-Bench provides a principled evaluation framework to diagnose this limitation and to guide progress toward more unified language representations across tokenized text and pixels. The source dataset and code are publicly available at \href{https://github.com/QingAnLiu/VISTA-Bench}{\textbf{VISTA-Bench}}.

\end{abstract}

\section{Introduction}

\begin{figure}[t]
  \centering
  \includegraphics[width=\linewidth]{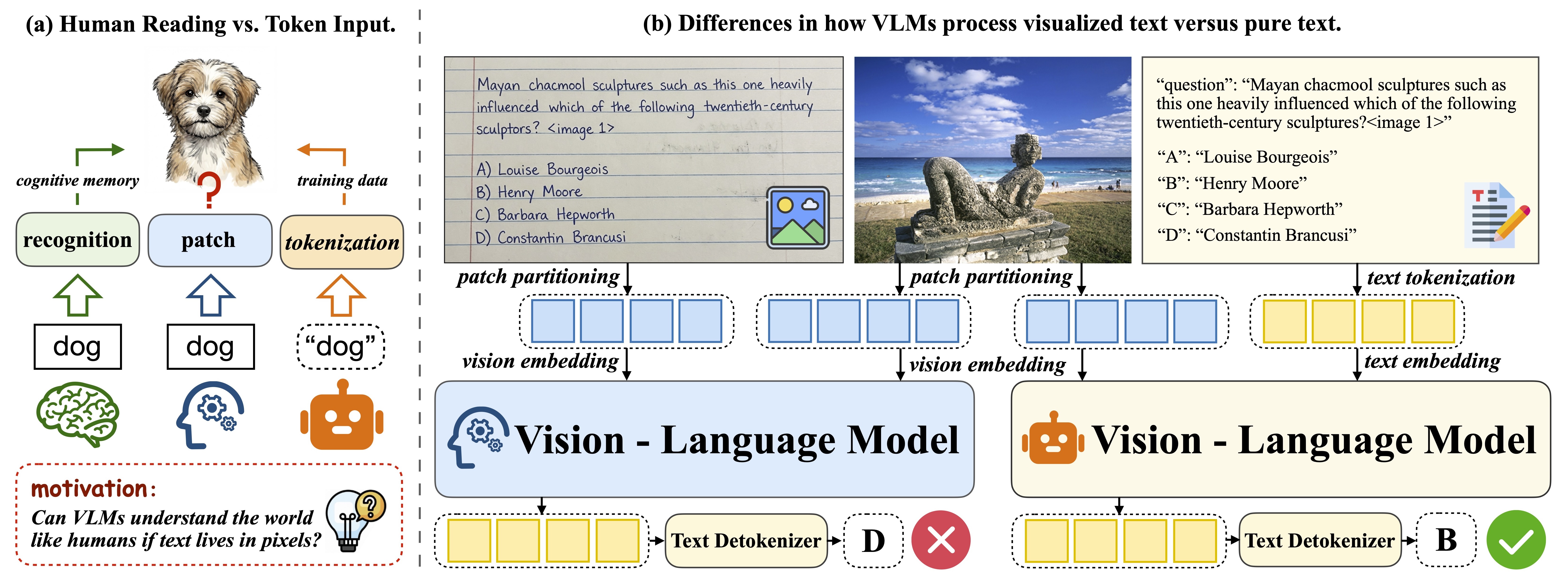}
  \caption{(a) Humans integrate visual context with embedded text, while standard VLM evaluation typically provides language as discrete tokens. (b) When language is presented as visualized text, VLMs can behave differently from the pure text, inducing a modality gap.}
  \label{fig:motivation}
\end{figure}

Recent vision–language models (VLMs)~\cite{bai2025qwen3vltechnicalreport,diao2025pixels,wang2025internvl3} extend large language models (LLMs)~\cite{cai2024internlm2,grattafiori2024llama,yang2025qwen3} to diverse tasks by treating visual tokens and text queries as inputs. Accordingly, existing benchmarks predominantly assume pure-text queries.
However, real-world scenarios often embed language within visual content, as shown in Figure~\ref{fig:motivation}(a), raising a fundamental question:
\textbf{Do vision–language models truly understand visualized text as well as pure text?}
Recent explorations such as DeepSeek-OCR~\cite{wei2025deepseek} and Glyph~\cite{cheng2025glyph} point to a growing \emph{text-as-pixels} paradigm, in which text is rendered into images to reduce token overhead in long contexts and to establish a unified perceptual interface across modalities. 
Crucially, replacing pure-text queries with visualized text fundamentally alters the input pathway in Figure~\ref{fig:motivation}(b), forcing all information to be processed through a unified modality.
These developments expose an interesting and underexplored challenge: whether foundation models can preserve semantic fidelity and cross-modal alignment when language is represented as pixels, especially for visual–language understanding and unified model architecture design in the future.

Notably, existing VLM evaluation protocols remain text-centric, spanning both standard language reasoning benchmarks~\cite{hendrycks2020measuring} and widely used multimodal benchmarks~\cite{liu2024mmbench, yue2024mmmu, li2024seed}. This blind spot overlooks the perceptual challenge of reading language from pixels, leaving it unclear whether model behavior remains stable when language isn't conveyed symbolically.
Fortunately, concurrent works start to expose this gap.
VTCBench~\cite{zhao2025vtcbench} reports significant performance degradation when text-only tasks are converted to visualized text.
Impressively, Gemini-3-Pro~\cite{google2025gemini3pro} exhibits a tiny modality gap, highlighting the potential for modality-equivalent representations.
However, existing analyses are largely confined to unimodal settings. It therefore remains unclear whether similar phenomena persist in multimodal tasks from perception to reasoning and how they manifest in such scenarios.

To address this question, we uncover a pronounced modality gap between pure-text and visualized-text inputs through preliminary experiments. We then introduce \textbf{VISTA-Bench}, a benchmark of 1,500 carefully filtered samples designed for rigorous comparison under strictly matched pure-text and visualized-text evaluation conditions.
VISTA-Bench is built through a three-stage pipeline that curates high-quality questions, renders text into diverse visual layouts, and enforces rendering fidelity via VLM-based verification and manual review. It contains multiple-choice questions and organizes them under a hierarchical capability taxonomy spanning perception, reasoning, and knowledge.

Evaluating over 30 state-of-the-art VLMs, we uncover a pervasive modality gap that is consistently amplified by perceptually challenging renderings. 
Our analysis attributes this gap primarily to limited perceptual robustness, while additional visual grounding offers only partial mitigation. 
Notably, MiMo-VL-7B-RL~\cite{coreteam2025mimovltechnicalreport} stands out as a rare exception, exhibiting markedly stronger robustness under visualized-text inputs.
Overall, VISTA-Bench provides a principled testbed for diagnosing modality gaps under realistic text-as-pixels inputs and underscores the need for more robust unified representations in future VLMs.
Our main contributions are summarized as follows:

\begin{itemize}

\item \textbf{Exposing a modality non-equivalence phenomenon.}
We uncover comprehensive analyses across distinct multi-modality tasks when text is presented as pixels rather than symbols.

\item \textbf{Defining a systematic visualized-text benchmark.}
VISTA-Bench is a rigorously controlled and easy-to-deploy benchmark spanning perception, reasoning, and knowledge capabilities, filling a blind spot in evaluating the gap between pure-text and visualized-text inputs.

\item \textbf{Providing a text-as-pixels diagnosis and outlook.}
Large-scale analyses identify limited perceptual robustness as the core driver of the modality gap, and VISTA-Bench elevates text-as-pixels to a foundational challenge for unified vision–language representations.

\end{itemize}

\section{Related Work}

\subsection{Multimodal Understanding Benchmarks} 

The rapid advancement of Vision-Language Models (VLMs) has led to significant breakthroughs in areas such as fine-grained OCR, text retrieval and text-image understanding. To comprehensively evaluate these capabilities, a variety of benchmarks have been proposed. Early efforts, including DocVQA~\cite{mathew2021docvqa} and OCRBench-v2~\cite{fu2024ocrbench}, primarily concentrate on localized text recognition and document-level OCR tasks. Subsequent benchmarks have expanded their scope to encompass higher-level visual cognition, reasoning and the integration of external knowledge~\cite{li2024seed, liu2024mmbench, yue2024mmmu}. Recent benchmarks also study fine-grained visual perception, advanced visual reasoning, and agentic visual research, such as V-FAT~\cite{wang2026v}, VisuLogic~\cite{xu2025visulogic}, and Vision-DeepResearch Benchmark~\cite{zeng2026vision}. Despite this progress, most existing evaluation paradigms predominantly focus on pure-text. Following the philosophies of comprehensive benchmarks, we introduce VISTA-Bench, a systematic benchmark from multimodal perception, reasoning, to unimodal understanding domains based on visualized-text input. VISTA-Bench serves as a necessary supplement that reveals the modality gap between pure and visualized text and provides a critical entry point for research on unified multimodal representations.

\subsection{Emerging Paradigms in Visualized Text} 

Recent advancements, such as DeepSeek-OCR~\cite{wei2025deepseek} and Glyph \cite{cheng2025glyph}, have revitalized the text-as-pixel paradigm, demonstrating that visualized text can significantly enhance text compression and inference efficiency while maintaining accuracy~\cite{li2025text,xing2025see}. A concurrent work, VTCBench~\cite{zhao2025vtcbench}, renders text-only tasks into images, revealing that most models suffer from a significant modality gap, whereas Gemini~\cite{google2025gemini3pro} exhibits much higher robustness to such transitions, indicating strong potential for unified visual-textual representations. However, this benchmark is largely confined to single-modality tasks and falls short of covering the diverse modality and broad capability required in real-world applications.
In the multimodal domain, several efforts merge images and questions into a single visual input~\cite{an2025voqa,li2023text}. However, this all-in-one design offers limited flexibility, especially for text-dense images, where overlaying queries can introduce visual interference. In contrast, VISTA-Bench adopts a decoupled formulation that integrates multimodal and unimodal tasks while separating original images from question-rendered images. This design enables a more comprehensive benchmark and allows precise, efficient evaluation of the modality gap between pure-text and visualized-text inputs.

\section{VISTA-Bench}

\begin{wrapfigure}[7]{r}{0.46\textwidth}
  \vspace{-0.40in}
  \centering
  \includegraphics[width=0.44\textwidth]{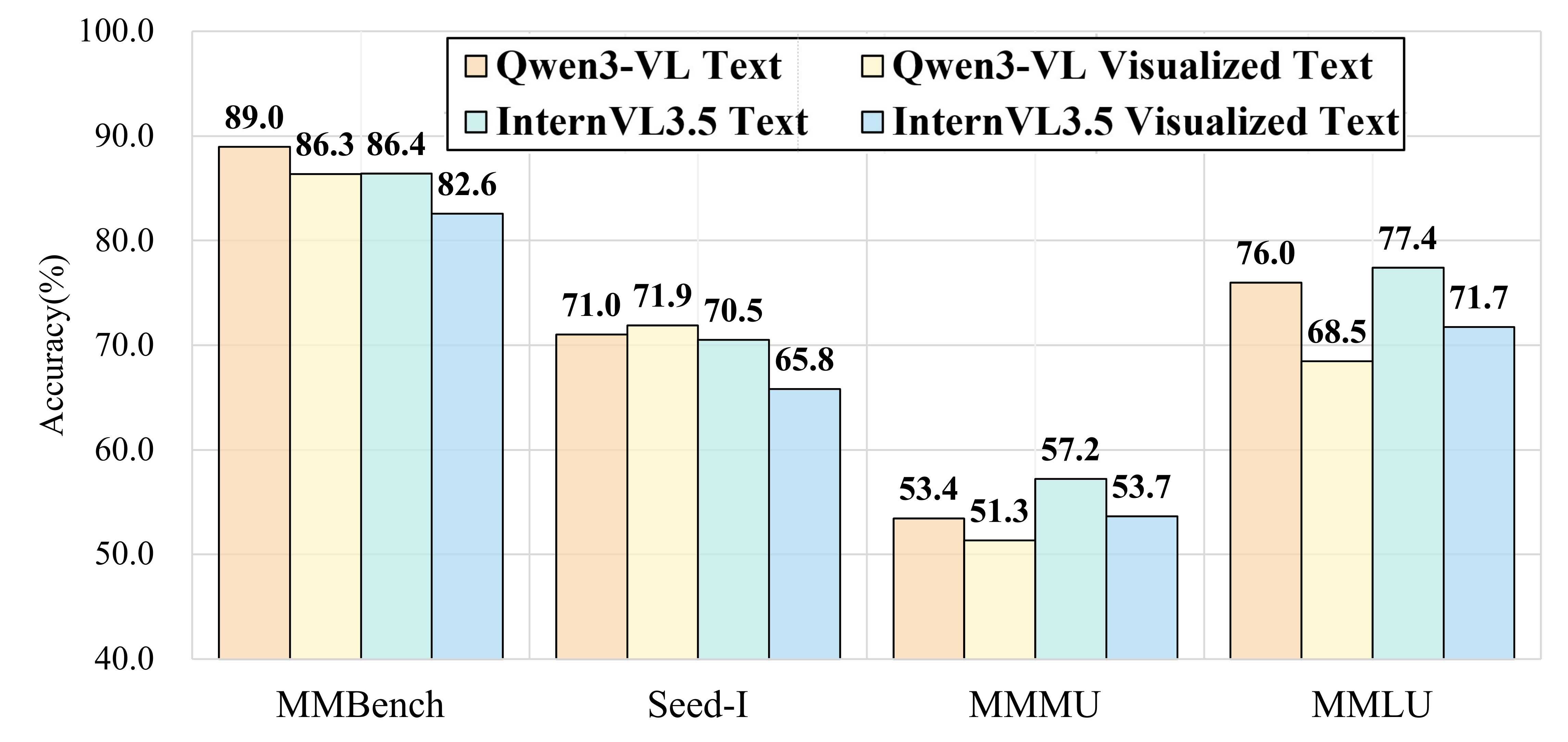}
  \caption{Comparison between Text and Visualized Text Inputs.}
  \label{fig:visualized_text}
  \vspace{-0.15in}
\end{wrapfigure}

We introduce VISTA-Bench, the first benchmark that uses visualized text to systematically evaluate the visual understanding capabilities of VLMs. In this section, we first present the task settings and preliminary experiments to reveal three key findings. Based on these insights, we detail the construction of VISTA-Bench in three steps.

\subsection{Task Settings}

We study the impact of replacing pure text with visualized text under both unimodal and multimodal tasks. Each question is evaluated in two forms: pure-text questions are provided as textual prompts, whereas visualized-text questions are processed through the vision encoder. For the unimodal task, we use MMLU~\cite{hendrycks2020measuring}. For the multimodal task, we evaluate three benchmarks, including MMBench~\cite{liu2024mmbench}, Seed-Bench~\cite{li2024seed}, and MMMU~\cite{yue2024mmmu}. 
Experiments are conducted on two VLMs, Qwen3-VL-8B-Instruct~\cite{bai2025qwen3vltechnicalreport} and InternVL3.5-8B~\cite{wang2025internvl3}. Unless otherwise specified, visualized text is rendered in \textit{Arial} at 16pt, with a fixed width of 800 pixels and adaptive height. With the above settings, we can fully explore the modality gap introduced by visualized text across both task paradigms.

\subsection{Preliminary experiments}
\label{sec:preliminary_experiments}
To investigate the modality gap between text and visualized text, we conduct preliminary experiments. Here are three key findings we summarized from the experiments.

\definecolor{mydarkblue}{RGB}{0, 20, 115}
{\textbf{\textcolor{mydarkblue}{Finding 1}: Replacing pure text with visualized text reveals a persistent modality gap.}}  
\textit{Visualized text consistently underperforms pure text.} 
As shown in Figure~\ref{fig:visualized_text}, in unimodal tasks, the drop is largest, with Qwen3-VL-8B-Instruct accuracy falling from 75.99\% to 68.46\% on MMLU. Multimodal tasks show smaller declines, indicating that image input partially compensates for visualized text. Overall, Qwen3-VL-8B-Instruct outperforms InternVL3.5-8B.

{\textbf{\textcolor{mydarkblue}{Finding 2}: Visualized text processing is governed by both perceptual robustness and instruction sensitivity.}}  
\textit{Appropriate settings improve accuracy, while poor choices can sharply degrade it.} From the visualized text and prompt, we explore three key factors that directly affect the model's processing of visualized text. Details are as follows.

\begin{figure*}[!t]
  \centering
  \includegraphics[width=\textwidth]{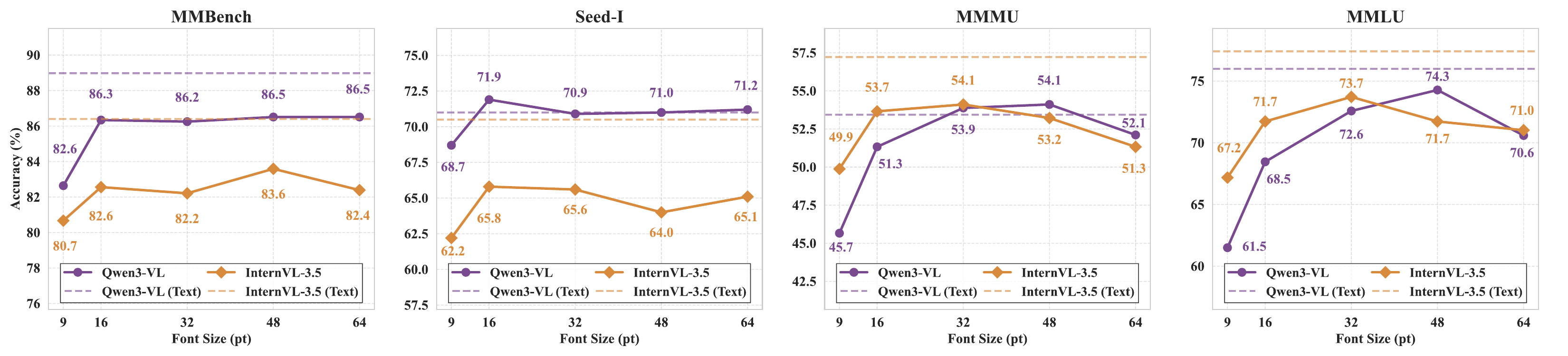}
  \includegraphics[width=\textwidth]{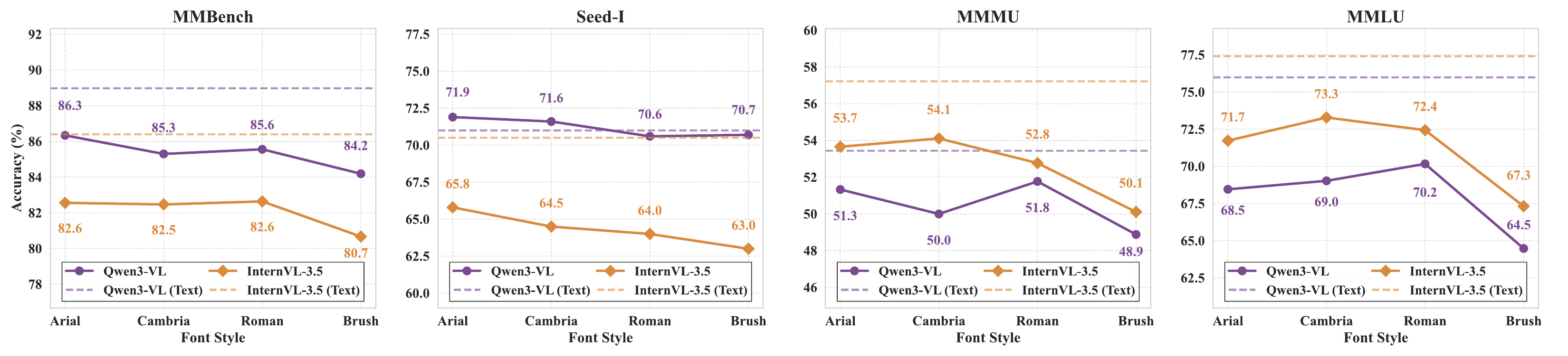}
  \caption{\textbf{Perceptual factor impact.} \textbf{Top:} Font Size (9, 16, 32, 48, 64). \textbf{Bottom:} Font Style (\textit{Arial, Cambria, Roman, Brush}).}
  \label{fig:overall_impact}
  \vspace{-1.8em}
\end{figure*}

\textbf{Font Size.} Text are rendered in \textit{Arial} at 9pt, 16pt, 32pt, 48pt, and 64pt to examine how font size affects the modality gap. As shown in Figure~\ref{fig:overall_impact} top row, font size substantially influences performance across models and tasks: very small fonts yield the lowest accuracy due to poor legibility, with a larger impact in the unimodal setting where all semantics must be recovered from pixels. Increasing the font size to 32--48pt consistently improves accuracy, with the largest gains on MMLU. At 64pt, performance saturates or slightly declines because larger text induces more line wrapping and reduces the effective context per image. Overall, even under the best font sizes, visualized text remains worse than pure text, confirming a persistent modality gap.

\textbf{Font Style.} Building on the font size analysis, we fix the rendering size at 16 pt to examine font style effects on the modality gap. We compare a sans-serif font, \textit{Arial}; two serif fonts, \textit{Times New Roman} and \textit{Cambria}; and a handwritten-style font, \textit{Brush Script MT}. 
Across both models and tasks, standard fonts yield similar performance, with only minor variations from the \textit{Arial} baseline. The handwritten-style font consistently degrades accuracy across all benchmarks, with the largest impact on MMLU, where Qwen3-VL-8B-Instruct drops from 68.5\% (\textit{Arial}) to 64.5\% (\textit{Brush}) in Figure~\ref{fig:overall_impact}. Overall, the gap is minimal with standard fonts but substantial with handwritten-style fonts.

\textbf{Prompt Design.} We compare five prompt variants to assess prompt sensitivity, with details provided in Appendix Figure~\ref{fig:app_prompt_design}.
Moderate-length prompts with semantic guidance slightly improve accuracy and reduce the modality gap in several settings. However, prompt effects are model- and task-dependent rather than uniformly beneficial. CoT brings limited changes for Qwen3-VL-8B-Instruct, but substantially improves InternVL3.5-8B on MMMU and MMLU, while remaining comparable on MMBench and Seed-Bench. Overall, Qwen3-VL-8B-Instruct is relatively stable across prompts, whereas InternVL3.5-8B shows larger fluctuations, indicating stronger instruction sensitivity.

{\textbf{\textcolor{mydarkblue}{Finding 3}: The text recognition capability of VLMs is closely related to the modality gap.}}
\textit{Models with stronger text recognition abilities show a smaller modality gap.}
In the experiments above, we observe that Qwen3-VL-8B-Instruct has a smaller gap on visualized text, while InternVL3.5-8B shows a larger gap. We attribute this difference to stronger visual text recognition. On OCR-related benchmarks, Qwen3-VL-8B-Instruct scores 96.1 on DocVQA\_test~\cite{mathew2021docvqa} and 896 on OCRBench~\cite{liu2024ocrbench}, whereas InternVL3.5-8B scores 92.3 and 832. These results suggest that better OCR ability helps Qwen3-VL-8B-Instruct interpret visualized text more accurately, which in turn reduces the gap.

Building on these findings, we observe that visualized-text inputs induce a consistent modality gap across models and tasks, whose magnitude depends on recognition robustness and is modulated by rendering and prompting choices. This motivates the construction of a dedicated benchmark to systematically evaluate VLMs under visualized-text conditions.

\begin{figure*}[t]
    \centerline{\includegraphics[width=\textwidth]{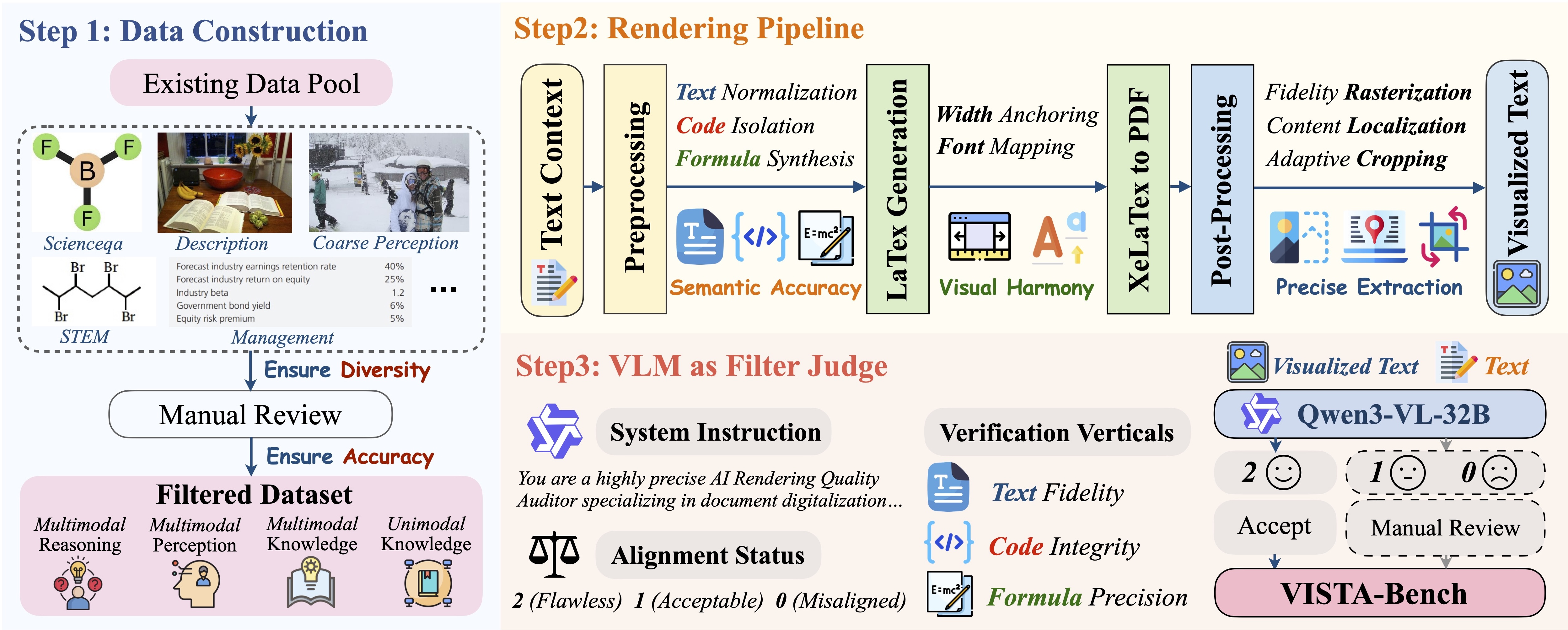}}
    \caption{
      Overview of the construction. First, we extract filtered dataset from existing data rely on diversity and accuracy. Second, we transform text into visualized text through the rendering pipeline. We then validate the precision of visualized text using a VLM-based filter judge and manual review. Finally, we establish VISTA-Bench, supported by a sophisticated rendering pipeline.
    }
    \label{fig:Construction_of_the_VISTABench_framework}
\end{figure*}

\subsection{VISTA-Bench Construction}

We propose \textbf{VISTA-Bench}, a benchmark designed to evaluate the capability of models to process visualized text. As shown in Figure \ref{fig:Construction_of_the_VISTABench_framework}, construction of VISTA-Bench consists of three steps:  \textbf{\textit{(i)}} Data Construction, \textbf{\textit{(ii)}} Rendering Pipeline, and \textbf{\textit{(iii)}} VLM as Filter Judge. Below, we provide details of the construction and a brief introduction to the benchmark.

\textbf{Step 1: Data Construction.} We assemble a data pool by collecting various samples across distinct categories from existing benchmarks~\cite{liu2024mmbench, yue2024mmmu, li2024seed,hendrycks2020measuring}. To guarantee diversity, our initial data extraction was guided by the underlying categories present in the data pool. Subsequently, the extracted samples were subjected to a rigorous manual review to ensure their accuracy. Exclusively correct samples populate the filtered dataset prior to the rendering process.

\begin{wrapfigure}[17]{r}{0.46\textwidth}
    \vspace{-0.17in}
    \includegraphics[width=0.44\textwidth]{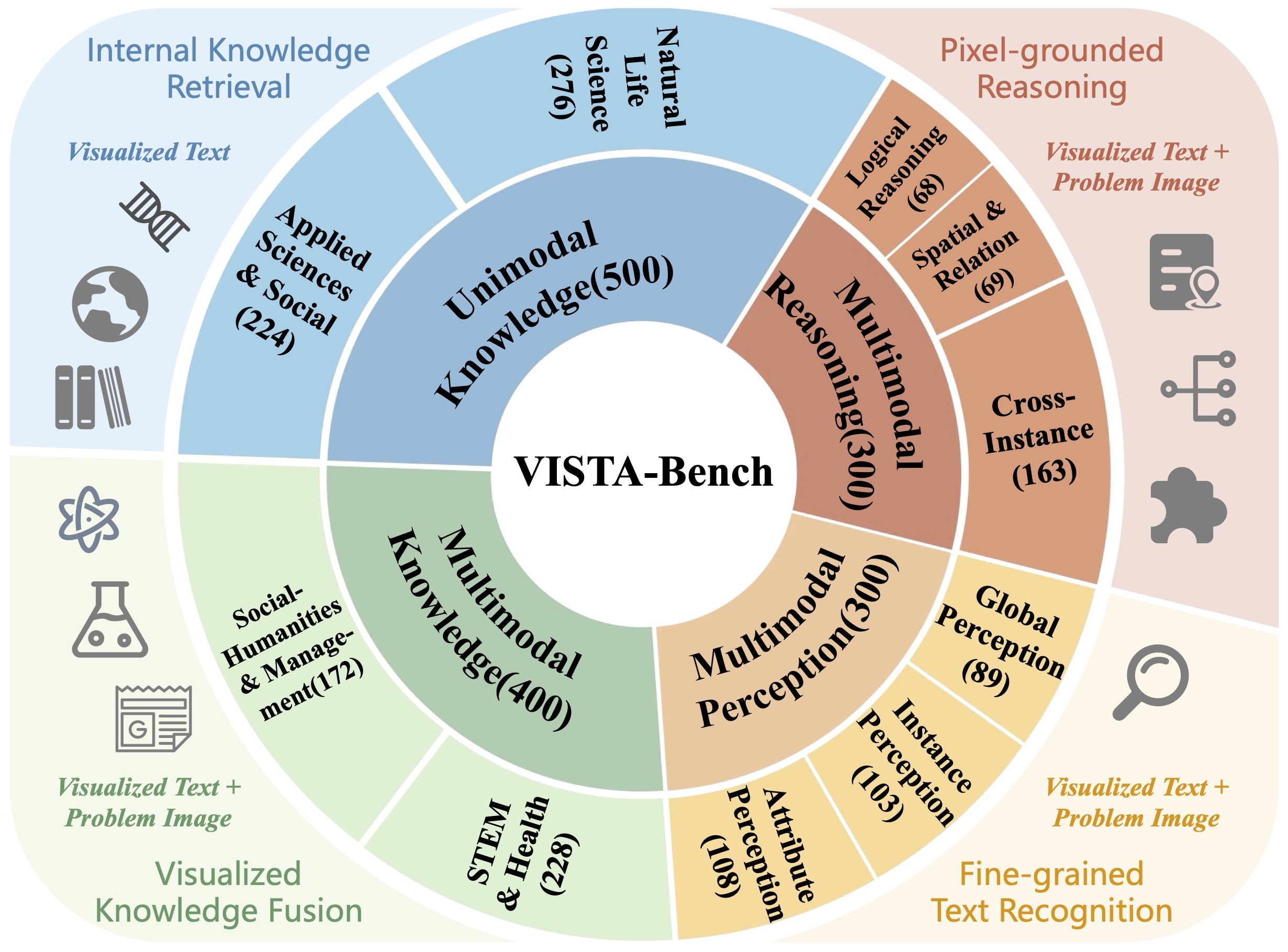}
    \caption{\textbf{Ability dimensions in VISTA-Bench.} VISTA-Bench includes two main levels of dimensions based on inherent modality and cognitive dimension, with 10 sub-tasks.
    }
    \label{fig:ability_dimensions}
\end{wrapfigure}

\textbf{Step 2: Rendering Pipeline.} 
Since the filtered dataset preserves structured code and LaTeX formulas, naive rendering without specialized handling can produce unreadable text and visual artifacts. To mitigate this, we propose a LaTeX-based rendering pipeline. During preprocessing, we apply dedicated treatment to text, code, and formulas to preserve semantic accuracy.
Width anchoring and font mapping in the generation stage maintain visual harmony and font diversity. 
In post-processing, we perform fidelity-preserving rasterization, content localization, and adaptive cropping to obtain the precisely extracted visualized text regions. 
Appendix~\ref{app:rendering_layout_sensitivity} further verifies that varying layout-level rendering parameters such as width and margins changes absolute visualized-text accuracy but does not overturn the modality gap.

\textbf{Step 3: VLM as Filter Judge.}
To ensure the quality of visualized text, we use Qwen3-VL-32B-Instruct~\cite{bai2025qwen3vltechnicalreport} as a VLM-based filter judge to check rendering fidelity. Guided by system instructions, the judge evaluates whether the rendered text, code, and formulas are faithfully preserved. A three-tier hierarchical alignment status is employed to categorize rendering quality, and samples scoring below 2 are further sent for manual verification. Through this procedure, we establish VISTA-Bench.

\textbf{VISTA-Bench} comprises 1,500 instances, predominantly featuring Multiple Choice Questions alongside a selective set of open-ended queries. Each instance provides dual data formats: visualized text and their corresponding text counterparts.
As illustrated in Figure \ref{fig:ability_dimensions}, we establish a hierarchical evaluation framework to systematically assess the capabilities of VLMs in processing visualized text. Our benchmark is structured into three levels, comprising 4 primary tasks, 10 sub-tasks and 27 fine-grained dimensions. The design motivations for the four primary tasks are detailed below:

\noindent $\bullet$ \textbf{Multimodal Perception.} This task contains 300 instances and evaluates perceptual grounding under visualized-text inputs. It covers three sub-dimensions: \textbf{\textit{(i)}} Global Perception for holistic scene understanding, \textbf{\textit{(ii)}} Instance Perception for locating or identifying specific entities and \textbf{\textit{(iii)}} Attribute Perception for recognizing fine-grained properties. Together, it measures how well models extract and ground information when textual cues are presented as dense pixels.

\noindent $\bullet$ \textbf{Multimodal Reasoning.} This task contains 300 instances and evaluates reasoning over visualized-text inputs with accompanying visual context. It spans three sub-dimensions: \textbf{\textit{(i)}} Logical Reasoning for multi-step inference, \textbf{\textit{(ii)}} Spatial \& Relation Understanding for relative positions and relations, and \textbf{\textit{(iii)}} Cross-Instance Reasoning for aggregating evidence across multiple visual elements. It measures whether VLMs can effectively execute non-trivial reasoning when key cues are rendered as pixels.

\noindent $\bullet$ \textbf{Multimodal Knowledge.} This task includes 400 knowledge-intensive instances and evaluates knowledge application when questions are presented as visualized text with visual evidence. We organize queries into two domains: \textbf{\textit{(i)}} STEM \& Health, covering scientific and medical concepts and \textbf{\textit{(ii)}} Social-Humanities \& Management, covering cultural and organizational topics. It measures whether models can retrieve and apply knowledge when linguistic cues must be read from pixels.

\noindent $\bullet$ \textbf{Unimodal Knowledge.} This task includes 500 instances and isolates knowledge retrieval from visual scene understanding by removing external images. Each instance renders a knowledge prompt as visualized text, requiring all evidence to be read from pixels. We cover two domains: \textbf{\textit{(i)}} Natural \& Life Sciences and \textbf{\textit{(ii)}} Social \& Applied Sciences. This setup diagnoses whether performance is primarily limited by decoding visualized text, rather than by the availability of visual context.

\definecolor{txtgray}{RGB}{232, 232, 232}    
\definecolor{myblue}{RGB}{235, 245, 255}
\definecolor{lighttext}{RGB}{90,90,90} 
\definecolor{gapgreen}{RGB}{200, 0, 0}   
\definecolor{gapred}{RGB}{0, 153, 0}     

\newcommand{\gapdown}[1]{\textcolor{gapgreen}{#1}}
\newcommand{\gapup}[1]{\textcolor{gapred}{#1}}


\newcommand{\GapMin}{-32}
\newcommand{\GapMax}{0}

\definecolor{gaplow}{RGB}{252,252,252}
\definecolor{gaphighred}{RGB}{220,170,170}
\definecolor{gaphighgreen}{RGB}{198,225,205} 

\newcommand{\gapheat}[2]{%
  \begingroup
  \pgfmathsetmacro{\gg}{max(\GapMin, min(\GapMax, #1))}%
  \pgfmathsetmacro{\ppfloat}{100*(\gg-\GapMax)/(\GapMin-\GapMax)}%
  \pgfmathtruncatemacro{\pp}{round(\ppfloat)}%
  \edef\gapcol{gaphighred!\pp!gaplow}%
  \expandafter\cellcolor\expandafter{\gapcol}#2%
  \endgroup
}

\newcommand{\gapheatdown}[2]{\gapheat{#1}{\gapdown{#2}}}
\newcommand{\gapheatup}[2]{\gapheat{0}{\gapup{#2}}}

\newcommand{\gapheatspecialup}[1]{%
  \begingroup
  \edef\gapcol{gaphighgreen!35!gaplow}
  \expandafter\cellcolor\expandafter{\gapcol}\gapup{#1}%
  \endgroup
}

\newcommand{\txtcell}[1]{%
  \begingroup
  \setlength{\fboxsep}{0.45pt}%
  \colorbox{txtgray}{\strut\hspace{0.6pt}#1\hspace{0.6pt}}%
  \endgroup
}

\newcommand{\na}{\phantom{0.0}}
\newcommand{\nagap}{\phantom{↓ -0.0}}

\newcolumntype{M}{>{\raggedright\arraybackslash}p{0.301\textwidth}} 
\newcolumntype{V}{>{\centering\arraybackslash}p{0.039\textwidth}}  
\newcolumntype{T}{>{\centering\arraybackslash}p{0.039\textwidth}}  
\newcolumntype{G}{>{\centering\arraybackslash}p{0.064\textwidth}}  

\begin{table*}[!t]
  \caption{Comparison of different VLMs on our benchmark. Results are reported under \textbf{Visualized Text (VT)} and \textbf{Text} inputs for each metric. \textbf{Text} cells are shaded in light gray. The \textbf{best} result per column is bolded and the \underline{second best} is underlined. The ↓Gap column denotes the overall performance drop when switching from Text to Visualized Text. All metrics are reported as percentages (\%).}
  \label{tab:modality_comparison_compact}
  \centering
  \begin{small}
  \setlength{\tabcolsep}{2.6pt}
  \renewcommand{\arraystretch}{1.15}

  \begin{tabular}{M|V T|V T|V T|V T|V T|G}
    \toprule

    \multirow{3}{*}{\textbf{Model}} &
    \multicolumn{2}{c|}{\textbf{Multimodal}} &
    \multicolumn{2}{c|}{\textbf{Multimodal}} &
    \multicolumn{2}{c|}{\textbf{Multimodal}} &
    \multicolumn{2}{c|}{\textbf{Unimodal}} &
    \multicolumn{2}{c|}{\multirow{2}{*}{\textbf{Overall}}} &
    \multirow{3}{*}{\textbf{↓ Gap}} \\
    & \multicolumn{2}{c|}{\textbf{Perception}} &
      \multicolumn{2}{c|}{\textbf{Reasoning}} &
      \multicolumn{2}{c|}{\textbf{Knowledge}} &
      \multicolumn{2}{c|}{\textbf{Knowledge}} &
      \multicolumn{2}{c|}{} &
      \\
    \cline{2-11}
    & \textbf{VT} & \txtcell{\textbf{Text}}
    & \textbf{VT} & \txtcell{\textbf{Text}}
    & \textbf{VT} & \txtcell{\textbf{Text}}
    & \textbf{VT} & \txtcell{\textbf{Text}}
    & \textbf{VT} & \txtcell{\textbf{Text}}
    & \\
    \midrule

    \rowcolor{myblue}
    \multicolumn{12}{l}{\textbf{$\blacktriangledown$ Closed-source Vision-Language Models}} \\

    Gemini-3.1-Pro-Preview~\cite{google2025gemini3pro}
    & \textbf{72.0} & \txtcell{\textbf{74.0}}
    & \textbf{73.3} & \txtcell{\textbf{76.0}}
    & \textbf{80.5} & \txtcell{\textbf{82.3}}
    & \textbf{89.4} & \txtcell{\textbf{90.0}}
    & \textbf{80.3} & \txtcell{\textbf{81.9}}
    & \gapheatdown{-1.6}{↓ -1.6} \\

    GPT-5.2~\cite{singh2025openai}
    & \underline{57.7} & \txtcell{\underline{67.7}}
    & \underline{53.0} & \txtcell{\underline{63.0}}
    & \underline{46.0} & \txtcell{\underline{58.8}}
    & \underline{70.2} & \txtcell{\underline{83.2}}
    & \underline{57.8} & \txtcell{\underline{69.5}}
    & \gapheatdown{-11.7}{↓ -11.7} \\

    \midrule

    \rowcolor{myblue}
    \multicolumn{12}{l}{\textbf{$\blacktriangledown$ Large-scale Vision-Language Models}} \\

    Qwen3.5-122B-A10B~\cite{qwen3.5}
    & \textbf{71.7} & \txtcell{\textbf{74.0}}
    & \textbf{72.3} & \txtcell{\textbf{74.7}}
    & \textbf{67.8} & \txtcell{\textbf{78.8}}
    & \textbf{86.0} & \txtcell{\textbf{86.6}}
    & \textbf{75.5} & \txtcell{\textbf{79.6}}
    & \gapheatdown{-4.1}{↓ -4.1} \\

    GLM-4.6V~\cite{hong2025glm}
    & \underline{70.3} & \txtcell{\underline{71.0}}
    & \underline{67.0} & \txtcell{\underline{68.7}}
    & \underline{63.3} & \txtcell{\underline{69.0}}
    & \underline{80.8} & \txtcell{\underline{82.6}}
    & \underline{71.3} & \txtcell{\underline{73.9}}
    & \gapheatdown{-2.6}{↓ -2.6} \\

    InternVL3.5-241B-A28B~\cite{wang2025internvl3}
    & 68.0 & \txtcell{\textbf{74.0}}
    & 56.3 & \txtcell{67.0}
    & 43.5 & \txtcell{61.5}
    & 65.2 & \txtcell{82.0}
    & 58.2 & \txtcell{71.9}
    & \gapheatdown{-13.7}{↓ -13.7} \\

    \midrule

    \rowcolor{myblue}
    \multicolumn{12}{l}{\textbf{$\blacktriangledown$ Vision-Language Models (30B)}} \\

    Qwen3-VL-30B-A3B-Instruct~\cite{bai2025qwen3vltechnicalreport}
    & \underline{64.3} & \txtcell{\underline{71.0}} & \underline{51.0} & \txtcell{60.3} & 35.8 & \txtcell{\textbf{58.0}} & 54.6 & \txtcell{\underline{75.0}} & 50.8 & \txtcell{\textbf{66.7}} & \gapheatdown{-15.9}{↓ -15.9} \\

    InternVL3.5-30B-A3B~\cite{wang2025internvl3}
    & \underline{64.3} & \txtcell{70.3} & 50.3 & \txtcell{\underline{61.7}} & \textbf{41.8} & \txtcell{\underline{52.5}} & \textbf{61.8} & \txtcell{\textbf{75.2}} & \underline{54.7} & \txtcell{\underline{65.5}} & \gapheatdown{-10.8}{↓ -10.8} \\

    Kimi-VL-A3B-Thinking~\cite{team2025kimi}
    & \textbf{70.0} & \txtcell{\textbf{71.3}} & \textbf{52.7} & \txtcell{\textbf{66.0}} & \textbf{41.8} & \txtcell{43.5} & \underline{59.4} & \txtcell{68.8} & \textbf{55.5} & \txtcell{62.0} & \gapheatdown{-6.5}{↓ -6.5} \\

    \midrule

    \rowcolor{myblue}
    \multicolumn{12}{l}{\textbf{$\blacktriangledown$ Vision-Language Models (8B)}} \\

    GLM-4.1V-9B-Thinking~\cite{hong2025glm}
    & \textbf{70.7} & \txtcell{\underline{71.3}} & 58.7 & \txtcell{61.7} & \textbf{51.3} & \txtcell{\textbf{57.5}} & \textbf{73.8} & \txtcell{\underline{75.8}} & \textbf{64.1} & \txtcell{\textbf{67.2}} & \gapheatdown{-3.1}{↓ -3.1} \\

    Ovis2.5-9B~\cite{lu2025ovis2}
    & 68.3 & \txtcell{69.3} & 56.3 & \txtcell{\textbf{65.7}} & 40.3 & \txtcell{\underline{56.0}} & 66.0 & \txtcell{73.4} & 57.7 & \txtcell{\underline{66.4}} & \gapheatdown{-8.7}{↓ -8.7} \\

    MiMo-VL-7B-SFT~\cite{coreteam2025mimovltechnicalreport}
    & \underline{68.7} & \txtcell{69.3} & \textbf{61.3} & \txtcell{\underline{63.0}} & \underline{46.5} & \txtcell{47.5} & \underline{71.4} & \txtcell{\textbf{76.0}} & \underline{62.2} & \txtcell{64.5} & \gapheatdown{-2.3}{↓ -2.3} \\

    SAIL-VL2-8B~\cite{yin2025sail}
    & \underline{68.7} & \txtcell{70.0} & 54.3 & \txtcell{60.7} & 37.8 & \txtcell{45.8} & 58.0 & \txtcell{71.0} & 54.0 & \txtcell{62.0} & \gapheatdown{-8.0}{↓ -8.0} \\

    MiMo-VL-7B-RL~\cite{coreteam2025mimovltechnicalreport}
    & \textbf{70.7} & \txtcell{69.3} & \underline{59.0} & \txtcell{62.0} & 45.8 & \txtcell{44.0} & 70.4 & \txtcell{71.8} & 61.6 & \txtcell{61.9} & \gapheatdown{-0.3}{↓ -0.3} \\

    LLaVA-OneVision-1.5-8B~\cite{an2025llava}
    & 62.7 & \txtcell{68.7} & 46.3 & \txtcell{59.0} & 34.0 & \txtcell{44.8} & 57.4 & \txtcell{67.8} & 50.0 & \txtcell{60.1} & \gapheatdown{-10.1}{↓ -10.1} \\

    NEO-9B-SFT~\cite{diao2025pixels}
    & 32.7 & \txtcell{69.0} & 29.0 & \txtcell{58.0} & 25.0 & \txtcell{41.5} & 28.6 & \txtcell{69.2} & 28.5 & \txtcell{59.5} & \gapheatdown{-31.0}{↓ -31.0} \\

    InternVL3.5-8B~\cite{wang2025internvl3}
    & 61.3 & \txtcell{64.3} & 45.7 & \txtcell{52.3} & 36.0 & \txtcell{45.8} & 57.6 & \txtcell{71.2} & 50.2 & \txtcell{59.3} & \gapheatdown{-9.1}{↓ -9.1} \\

    Qwen3-VL-8B-Instruct~\cite{bai2025qwen3vltechnicalreport}
    & 65.3 & \txtcell{67.3} & 49.0 & \txtcell{49.3} & 37.8 & \txtcell{48.5} & 58.2 & \txtcell{68.4} & 52.3 & \txtcell{59.1} & \gapheatdown{-6.7}{↓ -6.7} \\

    Ovis2-8B~\cite{lu2024ovis}
    & 66.7 & \txtcell{71.0} & 47.7 & \txtcell{60.0} & 29.5 & \txtcell{41.0} & 50.8 & \txtcell{65.4} & 47.7 & \txtcell{58.9} & \gapheatdown{-11.3}{↓ -11.3} \\

    Qwen2.5-VL-7B-Instruct~\cite{bai2025qwen2}
    & 65.7 & \txtcell{65.3} & 52.7 & \txtcell{53.0} & 27.0 & \txtcell{37.5} & 62.4 & \txtcell{62.0} & 51.7 & \txtcell{54.3} & \gapheatdown{-2.7}{↓ -2.7} \\

    MiniCPM-V-4.5~\cite{yu2025minicpm}
    & 64.3 & \txtcell{\textbf{71.6}} & 45.7 & \txtcell{60.3} & 31.5 & \txtcell{36.0} & 50.4 & \txtcell{55.0} & 47.2 & \txtcell{54.3} & \gapheatdown{-7.1}{↓ -7.1} \\

    LLaVA-OneVision-7B~\cite{li2024llava}
    & 40.3 & \txtcell{66.0} & 27.0 & \txtcell{56.3} & 20.8 & \txtcell{35.8} & 27.0 & \txtcell{58.6} & 28.0 & \txtcell{53.5} & \gapheatdown{-25.5}{↓ -25.5} \\

    LLaVA-1.5-7B~\cite{liu2024improved}
    & 33.0 & \txtcell{58.7} & 28.7 & \txtcell{44.3} & 27.3 & \txtcell{28.0} & 24.0 & \txtcell{48.8} & 27.6 & \txtcell{44.3} & \gapheatdown{-16.7}{↓ -16.7} \\

    \midrule

    \rowcolor{myblue}
    \multicolumn{12}{l}{\textbf{$\blacktriangledown$ Vision-Language Models (2B / 3B)}} \\

    Ovis2.5-2B~\cite{lu2025ovis2}
    & \textbf{66.3} & \txtcell{\textbf{70.0}} & \textbf{51.7} & \txtcell{\textbf{58.7}} & 29.5 & \txtcell{\textbf{43.3}} & \underline{51.8} & \txtcell{\textbf{60.4}} & \underline{48.7} & \txtcell{\textbf{57.4}} & \gapheatdown{-8.7}{↓ -8.7} \\

    SAIL-VL2-2B~\cite{yin2025sail}
    & \underline{65.3} & \txtcell{\underline{69.7}} & \underline{47.3} & \txtcell{\underline{57.7}} & \underline{32.8} & \txtcell{40.5} & 43.6 & \txtcell{54.8} & 45.8 & \txtcell{\underline{54.5}} & \gapheatdown{-8.7}{↓ -8.7} \\

    InternVL3.5-2B~\cite{wang2025internvl3}
    & 56.0 & \txtcell{66.3} & 39.3 & \txtcell{50.3} & 30.8 & \txtcell{\underline{41.0}} & 45.4 & \txtcell{\underline{57.0}} & 42.4 & \txtcell{53.3} & \gapheatdown{-10.9}{↓ -10.9} \\

    Qwen2.5-VL-3B-Instruct~\cite{bai2025qwen2}
    & 65.0 & \txtcell{67.7}
    & 43.3 & \txtcell{54.3}
    & \textbf{33.3} & \txtcell{36.3}
    & \textbf{54.8} & \txtcell{56.6}
    & \textbf{48.8} & \txtcell{52.9}
    & \gapheatdown{-4.1}{↓ -4.1} \\

    NEO-2B-SFT~\cite{diao2025pixels}
    & 40.0 & \txtcell{68.3} & 31.3 & \txtcell{49.3} & 25.5 & \txtcell{38.3} & 29.4 & \txtcell{53.4} & 30.9 & \txtcell{51.5} & \gapheatdown{-20.7}{↓ -20.7} \\

    Qwen3-VL-2B-Instruct~\cite{bai2025qwen3vltechnicalreport}
    & 56.7 & \txtcell{69.0} & 41.7 & \txtcell{49.7} & 26.5 & \txtcell{27.0} & 49.0 & \txtcell{56.0} & 43.1 & \txtcell{49.6} & \gapheatdown{-6.5}{↓ -6.5} \\

    Ovis2-2B~\cite{lu2024ovis}
    & 58.3 & \txtcell{66.7} & 39.7 & \txtcell{52.3} & 27.8 & \txtcell{31.5} & 36.0 & \txtcell{50.8} & 39.0 & \txtcell{49.1} & \gapheatdown{-10.1}{↓ -10.1} \\

    DeepSeek-VL2-Tiny~\cite{wu2024deepseek}
    & 44.3 & \txtcell{64.0} & 31.3 & \txtcell{41.0} & 27.8 & \txtcell{29.8} & 27.6 & \txtcell{41.8} & 31.7 & \txtcell{42.9} & \gapheatdown{-11.1}{↓ -11.1} \\

    \bottomrule
  \end{tabular}
  \end{small}
  \vskip -0.1in
\end{table*}

\section{Experiments}
\subsection{Setup}
\textbf{Models and Evaluation.}
We evaluate a broad suite of vision-language models spanning both open-source and closed-source systems.
For open-source models, we cover several representative scale regimes:
\textbf{\textit{(i)}} small models around 2--3B parameters,
\textbf{\textit{(ii)}} mid-sized models around 7--9B parameters,
\textbf{\textit{(iii)}} MoE-based 30B-A3B models,
and \textbf{\textit{(iv)}} larger-scale recent models with substantially stronger visual-text understanding capabilities.
Our open-source suite covers major model families, including InternVL3.5~\cite{wang2025internvl3},
Qwen2.5/3-VL~\cite{bai2025qwen2,bai2025qwen3vltechnicalreport},
LLaVA (1.5, OneVision and OneVision-1.5)~\cite{liu2024improved,li2024llava,an2025llava},
MiMo-VL~\cite{coreteam2025mimovltechnicalreport},
GLM-4.1V/4.6V~\cite{hong2025glm},
NEO~\cite{diao2025pixels},
DeepSeek-VL2~\cite{wu2024deepseek},
Ovis2/2.5~\cite{lu2024ovis, lu2025ovis2},
SAIL-VL2~\cite{yin2025sail},
Kimi-VL~\cite{team2025kimi},
MiniCPM-V-4.5~\cite{yu2025minicpm},
Qwen3.5-122B-A10B~\cite{qwen3.5},
and InternVL3.5-241B-A28B~\cite{wang2025internvl3}.
To further examine whether the visualized-text modality gap persists in frontier proprietary systems, we additionally evaluate two closed-source models, GPT-5.2~\cite{singh2025openai} and Gemini-3.1-Pro-Preview~\cite{google2025gemini3pro}.
All open-source evaluations are performed using VLMEvalKit~\cite{duan2024vlmevalkit} with the default decoding settings released by each official repository, ensuring consistent comparisons.
Experiments for open-source models are conducted in BF16 on NVIDIA A800 GPUs, while closed-source models are evaluated through their official API interfaces under the same benchmark protocol.

\subsection{Main Results}


\textbf{The modality gap is pervasive but model-dependent.}
As shown in Table~\ref{tab:modality_comparison_compact}, most VLMs exhibit accuracy drops when shifting from pure-text to visualized-text inputs, confirming that the modality gap remains a common challenge.
The gap is severe for some models, such as NEO-9B-SFT (-31.0), LLaVA-OneVision-7B (-25.5), and NEO-2B-SFT (-20.7), but much smaller for recent stronger models, including MiMo-VL-7B-RL (-0.3), MiMo-VL-7B-SFT (-2.3), Qwen2.5-VL-7B-Instruct (-2.7), and GLM-4.1V-9B-Thinking (-3.1).
Closed-source models also show clear heterogeneity: Gemini-3.1-Pro-Preview has only a 1.6-point drop, whereas GPT-5.2 still drops by 11.7 points.
These results suggest that visualized-text robustness is not guaranteed by stronger pure-text reasoning or larger scale, but depends on visual text recognition and cross-modal alignment.

\textbf{Within multimodal tasks, reasoning and knowledge exhibit amplified modality gaps.}
The subtask-level results in Table~\ref{tab:modality_comparison_compact} show that the visualized-text gap is not uniformly distributed within multimodal tasks.
For several strong models, perception remains relatively stable, while reasoning and knowledge-oriented subtasks degrade more sharply.
Ovis2.5-9B drops by only 1.0 point on perception, but by 9.4 and 15.7 points on reasoning and multimodal knowledge; Kimi-VL-A3B-Thinking similarly shows a 1.3-point perception gap but a 13.3-point reasoning gap.
This indicates that visualized-text inputs are not merely an OCR-like recognition challenge: small perception errors can propagate through multi-step inference and become amplified during knowledge retrieval or grounding.
The effect is model-dependent.
Gemini-3.1-Pro-Preview remains comparatively stable across multimodal subtasks, suggesting stronger visual-text alignment.
In contrast, NEO-9B-SFT suffers broad degradation across perception, reasoning, and knowledge, with drops of 36.3, 29.0, and 16.5 points, respectively.
This across-the-board failure suggests limited exposure to text-as-pixels inputs and weaker robustness to rendering-induced distribution shifts.

\textbf{Unimodal knowledge directly exposes the visualized-text processing bottleneck.}
This split removes image-content reasoning and mainly tests whether models can use knowledge-intensive questions presented as pixels.
Table~\ref{tab:modality_comparison_compact} shows a highly polarized pattern.
Some models remain near parity, such as Qwen2.5-VL-7B-Instruct (+0.4), Gemini-3.1-Pro-Preview (-0.6), MiMo-VL-7B-RL (-1.4), and GLM-4.1V-9B-Thinking (-2.0), suggesting that robust text-as-pixels understanding is achievable.
In contrast, NEO-9B-SFT drops by 40.6 points despite strong pure-text accuracy, indicating that its language-side knowledge is not reliably activated from rendered text.
The LLaVA family shows similar progress across iterations: LLaVA-OneVision-7B drops by 31.6 points, while LLaVA-OneVision-1.5-8B reduces the gap to 10.4 points.
Overall, unimodal knowledge serves as a clean stress test for whether models can convert rendered text into semantic representations that support knowledge access, rather than merely possessing the required knowledge in text-token form.

\subsection{Fine-grained Analysis}


\begin{table}[t]
\centering
\caption{Performance comparison across pure-text, visualized-text, and OCR-converted inputs. The $\downarrow$ Gap column denotes the performance drop relative to the pure-text setting of the same model.}
\label{tab:ocr_baseline}
\vspace{0.5em}
\setlength{\tabcolsep}{4.8pt}
\renewcommand{\arraystretch}{1.12}
\small
\begin{tabular}{lcccc|cc}
\toprule
\textbf{Input Setting}
& \textbf{Multi-Per.}
& \textbf{Multi-Rea.}
& \textbf{Multi-Kno.}
& \textbf{Uni-Kno.}
& \textbf{Overall}
& \textbf{$\downarrow$ Gap} \\
\midrule

\multicolumn{7}{l}{\cellcolor{blue!8}\textbf{Qwen3-VL-8B-Instruct~\cite{bai2025qwen3vltechnicalreport}}} \\

Pure Text
& 67.3 & 49.3 & 48.5 & 68.4
& 59.1
& -- \\

Visualized Text
& 65.3 & 49.0 & 37.8 & 58.2
& 52.3
& \textcolor{red}{-6.8} \\

Tesseract OCR~\cite{smith2007overview}
& 41.3 & 25.7 & 27.3 & 37.6
& 33.2
& \cellcolor{red!12}\textcolor{red}{-25.9} \\

DeepSeek-OCR~\cite{wei2025deepseek}
& 62.0 & 46.0 & 44.8 & 65.4
& 55.3
& \cellcolor{red!6}\textcolor{red}{-3.8} \\

PaddleOCR-VL-1.5~\cite{cui2026paddleocr}
& 63.7 & 47.0 & 44.8 & 65.8
& 56.0
& \cellcolor{red!6}\textcolor{red}{-3.1} \\

\midrule

\multicolumn{7}{l}{\cellcolor{blue!8}\textbf{InternVL3.5-8B~\cite{wang2025internvl3}}} \\

Pure Text
& 64.3 & 52.3 & 45.8 & 71.2
& 59.3
& -- \\

Visualized Text
& 61.3 & 45.7 & 36.0 & 57.6
& 50.2
& \textcolor{red}{-9.1} \\

Tesseract OCR~\cite{smith2007overview}
& 40.3 & 29.0 & 27.3 & 42.8
& 35.4
& \cellcolor{red!12}\textcolor{red}{-23.9} \\

DeepSeek-OCR~\cite{wei2025deepseek}
& 62.0 & 45.3 & 43.5 & 66.6
& 55.3
& \cellcolor{red!6}\textcolor{red}{-4.0} \\

PaddleOCR-VL-1.5~\cite{cui2026paddleocr}
& 64.0 & 50.0 & 41.8 & 68.8
& 56.9
& \cellcolor{red!6}\textcolor{red}{-2.4} \\

\bottomrule
\end{tabular}
\end{table}

\textbf{OCR baselines further diagnose the modality gap.}
To examine whether the gap mainly stems from text extraction, we evaluate an OCR-converted setting, where visualized-text inputs are first converted into textual inputs by external OCR systems and then fed to the same models.
As shown in Table~\ref{tab:ocr_baseline}, two observations emerge.
\textbf{\textit{(i)} Text extraction quality is crucial.}
Naive OCR is insufficient: Tesseract achieves only 33.2 and 35.4 overall accuracy on Qwen3-VL-8B-Instruct and InternVL3.5-8B, respectively, far below their direct visualized-text scores of 52.3 and 50.2.
In contrast, PaddleOCR-VL-1.5 improves the overall accuracy to 56.0 and 56.9, reducing the gap to the pure-text setting from 6.8 to 3.1 points and from 9.1 to 2.4 points, respectively.
This supports that perceptual robustness remains a key bottleneck when language must be read from pixels.
\textbf{\textit{(ii)} OCR conversion does not remove the gap.}
If visualized-text degradation were only caused by recognition failure, strong OCR-converted inputs should approach the pure-text upper bound.
However, PaddleOCR-VL-1.5 still underperforms the corresponding pure-text results of 59.1 and 59.3.
This residual gap suggests that models must not only recover textual content, but also preserve structure-sensitive cues and align pixel-mediated language with pure-text reasoning representations.
Therefore, the modality gap reflects a broader modality non-equivalence between tokenized text and text-as-pixels inputs.

\begin{figure*}[!t]
  \centering
  \begin{minipage}[b]{0.48\textwidth}
    \centering
    \includegraphics[width=\textwidth]{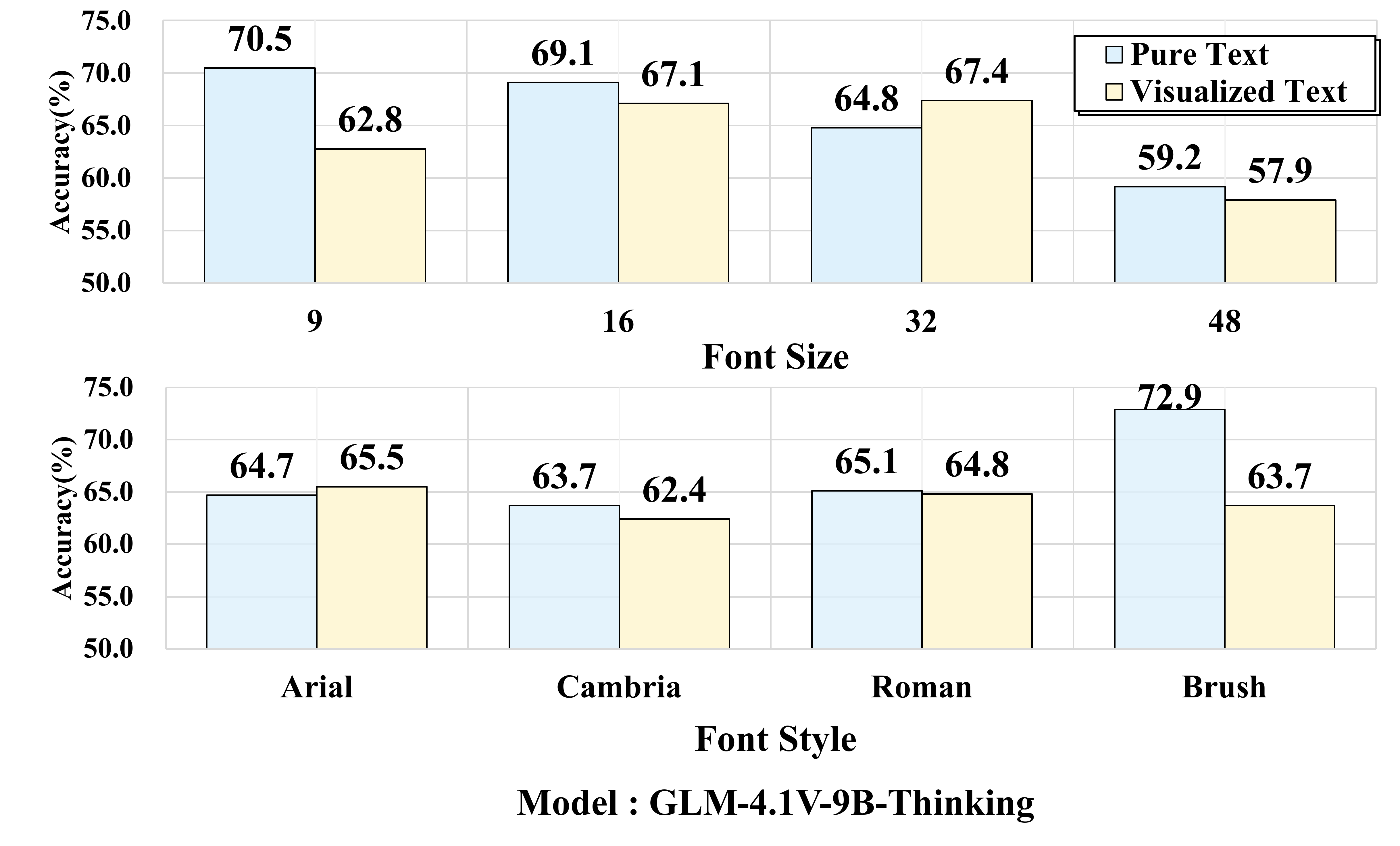}
    \caption{Sensitivity to font size and style on VISTA-Bench.}
    \label{fig:GLM_analysis}
  \end{minipage}
  \hfill 
  \begin{minipage}[b]{0.48\textwidth}
    \centering
    \includegraphics[width=\textwidth]{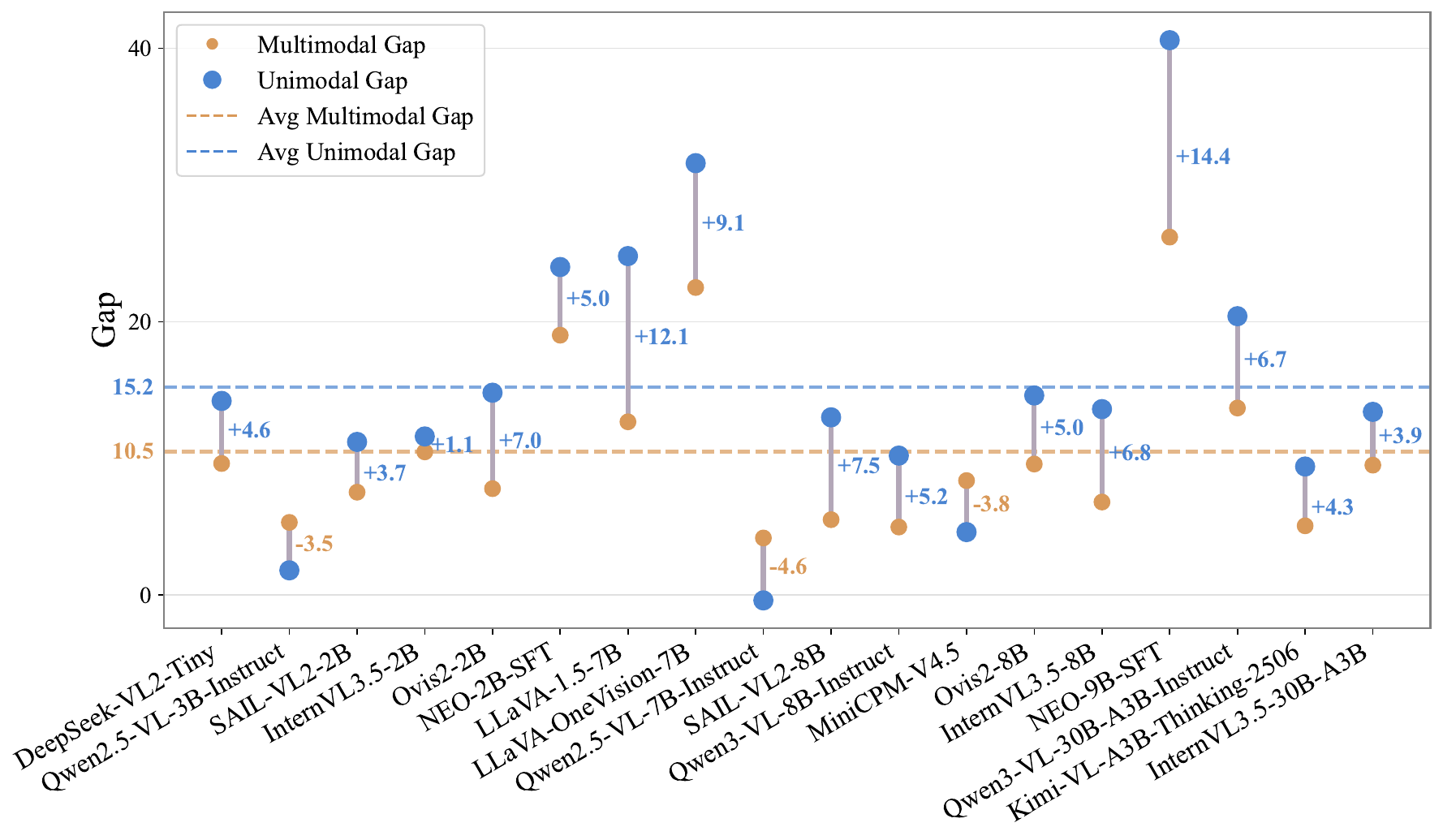}
    \caption{Modality Gap Across Multimodal and Unimodal Tasks.}
    \label{fig:case_study}
  \end{minipage}
\end{figure*}

\textbf{Models are sensitive to rendering complexity.}
We take GLM-4.1V-9B-Thinking as a representative model and examine its performance under varying font sizes and font styles in Figure~\ref{fig:GLM_analysis}.
Two observations stand out.
\textbf{\textit{(i)} Perceptually difficult renderings remain challenging.}
Even for this strong model, atypical presentations such as 9 pt small text and \textit{Brush-style} handwriting substantially enlarge the modality gap to 7.7 and 9.2 points, respectively, compared with its overall gap of 3.1 points.
This indicates that reduced legibility and rendering artifacts can still disrupt reliable evidence extraction from pixels.
\textbf{\textit{(ii)} Clean renderings can largely close the gap and occasionally surpass pure text.}
Under larger font sizes or standard font styles, the model becomes nearly gap-free, and in some settings visualized-text accuracy slightly exceeds pure-text accuracy.
This suggests that when visualized text is rendered cleanly, the image input can serve as a stable interface that supports precise grounding and reduces ambiguity in downstream decisions.
More results appear in Appendix~\ref{app:rendering_factor_ablations}.

\textbf{Multimodal context attenuates the modality gap.}
We define the modality gap as the accuracy difference between pure-text and visualized-text inputs, and report two aggregate metrics: \textbf{\textit{Multimodal Gap}}, computed over perception, reasoning, and knowledge tasks, and the \textbf{\textit{Unimodal Gap}}, measured on the knowledge task without additional images.
As shown in Figure~\ref{fig:case_study}, we visualize these two gaps across 18 representative models, where the blue dots denote the Unimodal Gap and the orange dots denote the Multimodal Gap.
These two metrics consistently diverge across models: the average Unimodal Gap reaches 15.2\%, whereas the Multimodal Gap is lower at 10.5\%.
This difference suggests that multimodal inputs provide contextual visual evidence that helps ground interpretation and constrain plausible answers, whereas unimodal inputs rely entirely on text recovered from pixels, making recognition errors harder to correct and more likely to be compensated by language priors.

\section{Limitations and Discussion}
We construct a systematic benchmark to evaluate visualized-text understanding from multimodal perception, reasoning, to unimodal knowledge.
We acknowledge several limitations that also suggest future directions.
\textbf{\textit{(i)} Continually evolving frontier models.}
Our evaluation covers a broad suite of open-source and closed-source VLMs.
However, frontier systems evolve rapidly, and their performance may change with new model versions, API updates, decoding settings, or preprocessing pipelines.
Future work should continuously update VISTA-Bench with newly released models and study whether the observed modality gaps persist under broader real-world deployment settings.
\textbf{\textit{(ii)} Evaluating LLMs via agentic auxiliaries.}
Beyond VLMs, LLMs can also be assessed under a complementary \emph{agentic} pipeline, where external tools convert visualized text and visual information into textual inputs.
This setting can help disentangle failures caused by visual recognition, tool use, and language-side reasoning.
\textbf{\textit{(iii)} Evaluating generation models for visualized-text understanding.}
Beyond discriminative VLMs, generative and unified multimodal models may provide another route for visualized-text understanding.
To explore this direction, we derive a small evaluation set from VISTA-Bench and evaluate Qwen-Image-Edit~\cite{wu2025qwen} in Appendix~\ref{app:qwen_image_edit_eval}, with generated outputs manually inspected for correctness.
This generative evaluation paradigm offers a complementary perspective on cross-modal representation learning and warrants further investigation.

\section{Conclusion}
We launch VISTA-Bench, an interesting and systematic benchmark spanning multimodal perception, reasoning, knowledge, and unimodal understanding, using decoupled visualized-text inputs rather than pure-text queries. Through extensive evaluation of over 30 representative VLMs, we uncover a substantial modality gap: models that perform strongly on pure text often degrade markedly when identical semantic content is presented as visualized text, with the gap widening as perceptual difficulty increases.
Besides, our experiments reveal that this phenomenon is pervasive across both unimodal and multimodal understanding scenarios, consistently leading to perceptual failures across modalities. OCR-based baselines further show that stronger text extraction narrows the modality gap, yet residual degradation remains beyond recognition errors.
In summary, VISTA-Bench establishes a principled evaluation and elevates text-as-pixels to a crucial challenge, motivating the evolution of textual and visual tokenization toward more unified and robust language representations.

\bibliographystyle{plainnat}
\bibliography{vista-bench}

@article{bai2025qwen2,
  title={Qwen2. 5-vl technical report},
  author={Bai, Shuai and Chen, Keqin and Liu, Xuejing and Wang, Jialin and Ge, Wenbin and Song, Sibo and Dang, Kai and Wang, Peng and Wang, Shijie and Tang, Jun and others},
  journal={arXiv preprint arXiv:2502.13923},
  year={2025}
}

@article{lu2024ovis,
  title={Ovis: Structural embedding alignment for multimodal large language model},
  author={Lu, Shiyin and Li, Yang and Chen, Qing-Guo and Xu, Zhao and Luo, Weihua and Zhang, Kaifu and Ye, Han-Jia},
  journal={arXiv preprint arXiv:2405.20797},
  year={2024}
}

@article{lu2025ovis2,
  title={Ovis2. 5 technical report},
  author={Lu, Shiyin and Li, Yang and Xia, Yu and Hu, Yuwei and Zhao, Shanshan and Ma, Yanqing and Wei, Zhichao and Li, Yinglun and Duan, Lunhao and Zhao, Jianshan and others},
  journal={arXiv preprint arXiv:2508.11737},
  year={2025}
}

@article{wu2024deepseek,
  title={Deepseek-vl2: Mixture-of-experts vision-language models for advanced multimodal understanding},
  author={Wu, Zhiyu and Chen, Xiaokang and Pan, Zizheng and Liu, Xingchao and Liu, Wen and Dai, Damai and Gao, Huazuo and Ma, Yiyang and Wu, Chengyue and Wang, Bingxuan and others},
  journal={arXiv preprint arXiv:2412.10302},
  year={2024}
}

@article{yu2025minicpm,
  title={Minicpm-v 4.5: Cooking efficient mllms via architecture, data, and training recipe},
  author={Yu, Tianyu and Wang, Zefan and Wang, Chongyi and Huang, Fuwei and Ma, Wenshuo and He, Zhihui and Cai, Tianchi and Chen, Weize and Huang, Yuxiang and Zhao, Yuanqian and others},
  journal={arXiv preprint arXiv:2509.18154},
  year={2025}
}

@article{team2025kimi,
  title={Kimi-vl technical report},
  author={Team, Kimi and Du, Angang and Yin, Bohong and Xing, Bowei and Qu, Bowen and Wang, Bowen and Chen, Cheng and Zhang, Chenlin and Du, Chenzhuang and Wei, Chu and others},
  journal={arXiv preprint arXiv:2504.07491},
  year={2025}
}

@article{yin2025sail,
  title={Sail-vl2 technical report},
  author={Yin, Weijie and Ye, Yongjie and Shu, Fangxun and Liao, Yue and Kang, Zijian and Dong, Hongyuan and Yu, Haiyang and Yang, Dingkang and Wang, Jiacong and Wang, Han and others},
  journal={arXiv preprint arXiv:2509.14033},
  year={2025}
}

@article{an2025llava,
  title={Llava-onevision-1.5: Fully open framework for democratized multimodal training},
  author={An, Xiang and Xie, Yin and Yang, Kaicheng and Zhang, Wenkang and Zhao, Xiuwei and Cheng, Zheng and Wang, Yirui and Xu, Songcen and Chen, Changrui and Zhu, Didi and others},
  journal={arXiv preprint arXiv:2509.23661},
  year={2025}
}

@article{singh2025openai,
  title={Openai gpt-5 system card},
  author={Singh, Aaditya and Fry, Adam and Perelman, Adam and Tart, Adam and Ganesh, Adi and El-Kishky, Ahmed and McLaughlin, Aidan and Low, Aiden and Ostrow, AJ and Ananthram, Akhila and others},
  journal={arXiv preprint arXiv:2601.03267},
  year={2025}
}

@misc{google2025gemini3pro,
      title={Gemini 3 Pro Model Card}, 
      author={Google},
      year={2025},
      url={https://storage.googleapis.com/deepmind-media/Model-Cards/Gemini-3-Pro-Model-Card.pdf},
      note={Technical Report}
}

@misc{qwen3.5,
    title  = {{Qwen3.5}: Towards Native Multimodal Agents},
    author = {{Qwen Team}},
    month  = {February},
    year   = {2026},
    url    = {https://qwen.ai/blog?id=qwen3.5}
}

@article{hong2025glm,
  title={Glm-4.5 v and glm-4.1 v-thinking: Towards versatile multimodal reasoning with scalable reinforcement learning},
  author={Hong, Wenyi and Yu, Wenmeng and Gu, Xiaotao and Wang, Guo and Gan, Guobing and Tang, Haomiao and Cheng, Jiale and Qi, Ji and Ji, Junhui and Pan, Lihang and others},
  journal={arXiv preprint arXiv:2507.01006},
  year={2025}
}

@article{li2024llava,
  title={Llava-onevision: Easy visual task transfer},
  author={Li, Bo and Zhang, Yuanhan and Guo, Dong and Zhang, Renrui and Li, Feng and Zhang, Hao and Zhang, Kaichen and Zhang, Peiyuan and Li, Yanwei and Liu, Ziwei and others},
  journal={arXiv preprint arXiv:2408.03326},
  year={2024}
}

@misc{bai2025qwen3vltechnicalreport,
      title={Qwen3-VL Technical Report}, 
      author={Shuai Bai and Yuxuan Cai and Ruizhe Chen and Keqin Chen and Xionghui Chen and Zesen Cheng and Lianghao Deng and Wei Ding and Chang Gao and Chunjiang Ge and Wenbin Ge and Zhifang Guo and Qidong Huang and Jie Huang and Fei Huang and Binyuan Hui and Shutong Jiang and Zhaohai Li and Mingsheng Li and Mei Li and Kaixin Li and Zicheng Lin and Junyang Lin and Xuejing Liu and Jiawei Liu and Chenglong Liu and Yang Liu and Dayiheng Liu and Shixuan Liu and Dunjie Lu and Ruilin Luo and Chenxu Lv and Rui Men and Lingchen Meng and Xuancheng Ren and Xingzhang Ren and Sibo Song and Yuchong Sun and Jun Tang and Jianhong Tu and Jianqiang Wan and Peng Wang and Pengfei Wang and Qiuyue Wang and Yuxuan Wang and Tianbao Xie and Yiheng Xu and Haiyang Xu and Jin Xu and Zhibo Yang and Mingkun Yang and Jianxin Yang and An Yang and Bowen Yu and Fei Zhang and Hang Zhang and Xi Zhang and Bo Zheng and Humen Zhong and Jingren Zhou and Fan Zhou and Jing Zhou and Yuanzhi Zhu and Ke Zhu},
      year={2025},
      eprint={2511.21631},
      archivePrefix={arXiv},
      primaryClass={cs.CV},
      url={https://arxiv.org/abs/2511.21631}, 
}

@article{wang2025internvl3,
  title={Internvl3. 5: Advancing open-source multimodal models in versatility, reasoning, and efficiency},
  author={Wang, Weiyun and Gao, Zhangwei and Gu, Lixin and Pu, Hengjun and Cui, Long and Wei, Xingguang and Liu, Zhaoyang and Jing, Linglin and Ye, Shenglong and Shao, Jie and others},
  journal={arXiv preprint arXiv:2508.18265},
  year={2025}
}

@inproceedings{liu2024improved,
  title={Improved baselines with visual instruction tuning},
  author={Liu, Haotian and Li, Chunyuan and Li, Yuheng and Lee, Yong Jae},
  booktitle={Proceedings of the IEEE/CVF conference on computer vision and pattern recognition},
  pages={26296--26306},
  year={2024}
}

@article{diao2025pixels,
  title={From Pixels to Words--Towards Native Vision-Language Primitives at Scale},
  author={Diao, Haiwen and Li, Mingxuan and Wu, Silei and Dai, Linjun and Wang, Xiaohua and Deng, Hanming and Lu, Lewei and Lin, Dahua and Liu, Ziwei},
  journal={arXiv preprint arXiv:2510.14979},
  year={2025}
}

@misc{coreteam2025mimovltechnicalreport,
      title={MiMo-VL Technical Report}, 
      author={MiMo},
      year={2025},
      eprint={2506.03569},
      archivePrefix={arXiv},
      primaryClass={cs.CL},
      url={https://arxiv.org/abs/2506.03569}, 
}

@article{yang2025qwen3,
  title={Qwen3 technical report},
  author={Yang, An and Li, Anfeng and Yang, Baosong and Zhang, Beichen and Hui, Binyuan and Zheng, Bo and Yu, Bowen and Gao, Chang and Huang, Chengen and Lv, Chenxu and others},
  journal={arXiv preprint arXiv:2505.09388},
  year={2025}
}

@article{grattafiori2024llama,
  title={The llama 3 herd of models},
  author={Grattafiori, Aaron and Dubey, Abhimanyu and Jauhri, Abhinav and Pandey, Abhinav and Kadian, Abhishek and Al-Dahle, Ahmad and Letman, Aiesha and Mathur, Akhil and Schelten, Alan and Vaughan, Alex and others},
  journal={arXiv preprint arXiv:2407.21783},
  year={2024}
}

@article{cai2024internlm2,
  title={Internlm2 technical report},
  author={Cai, Zheng and Cao, Maosong and Chen, Haojiong and Chen, Kai and Chen, Keyu and Chen, Xin and Chen, Xun and Chen, Zehui and Chen, Zhi and Chu, Pei and others},
  journal={arXiv preprint arXiv:2403.17297},
  year={2024}
}

@inproceedings{li2024seed,
  title={Seed-bench: Benchmarking multimodal large language models},
  author={Li, Bohao and Ge, Yuying and Ge, Yixiao and Wang, Guangzhi and Wang, Rui and Zhang, Ruimao and Shan, Ying},
  booktitle={Proceedings of the IEEE/CVF Conference on Computer Vision and Pattern Recognition},
  pages={13299--13308},
  year={2024}
}

@inproceedings{yue2024mmmu,
  title={Mmmu: A massive multi-discipline multimodal understanding and reasoning benchmark for expert agi},
  author={Yue, Xiang and Ni, Yuansheng and Zhang, Kai and Zheng, Tianyu and Liu, Ruoqi and Zhang, Ge and Stevens, Samuel and Jiang, Dongfu and Ren, Weiming and Sun, Yuxuan and others},
  booktitle={Proceedings of the IEEE/CVF Conference on Computer Vision and Pattern Recognition},
  pages={9556--9567},
  year={2024}
}

@article{hendrycks2020measuring,
  title={Measuring massive multitask language understanding},
  author={Hendrycks, Dan and Burns, Collin and Basart, Steven and Zou, Andy and Mazeika, Mantas and Song, Dawn and Steinhardt, Jacob},
  journal={arXiv preprint arXiv:2009.03300},
  year={2020}
}

@inproceedings{liu2024mmbench,
  title={Mmbench: Is your multi-modal model an all-around player?},
  author={Liu, Yuan and Duan, Haodong and Zhang, Yuanhan and Li, Bo and Zhang, Songyang and Zhao, Wangbo and Yuan, Yike and Wang, Jiaqi and He, Conghui and Liu, Ziwei and others},
  booktitle={European conference on computer vision},
  pages={216--233},
  year={2024},
  organization={Springer}
}

@article{liu2024ocrbench,
  title={Ocrbench: on the hidden mystery of ocr in large multimodal models},
  author={Liu, Yuliang and Li, Zhang and Huang, Mingxin and Yang, Biao and Yu, Wenwen and Li, Chunyuan and Yin, Xu-Cheng and Liu, Cheng-Lin and Jin, Lianwen and Bai, Xiang},
  journal={Science China Information Sciences},
  volume={67},
  number={12},
  pages={220102},
  year={2024},
  publisher={Springer}
}

@article{fu2024ocrbench,
  title={Ocrbench v2: An improved benchmark for evaluating large multimodal models on visual text localization and reasoning},
  author={Fu, Ling and Kuang, Zhebin and Song, Jiajun and Huang, Mingxin and Yang, Biao and Li, Yuzhe and Zhu, Linghao and Luo, Qidi and Wang, Xinyu and Lu, Hao and others},
  journal={arXiv preprint arXiv:2501.00321},
  year={2024}
}

@inproceedings{mathew2021docvqa,
  title={Docvqa: A dataset for vqa on document images},
  author={Mathew, Minesh and Karatzas, Dimosthenis and Jawahar, CV},
  booktitle={Proceedings of the IEEE/CVF winter conference on applications of computer vision},
  pages={2200--2209},
  year={2021}
}

@article{zhao2025vtcbench,
  title={VTCBench: Can Vision-Language Models Understand Long Context with Vision-Text Compression?},
  author={Zhao, Hongbo and Wang, Meng and Zhu, Fei and Liu, Wenzhuo and Ni, Bolin and Zeng, Fanhu and Meng, Gaofeng and Zhang, Zhaoxiang},
  journal={arXiv preprint arXiv:2512.15649},
  year={2025}
}

@article{wang2026v,
  title={V-FAT: Benchmarking Visual Fidelity Against Text-bias},
  author={Wang, Ziteng and He, Yujie and Li, Guanliang and Yang, Siqi and Xiong, Jiaqi and Liu, Songxiang},
  journal={arXiv preprint arXiv:2601.04897},
  year={2026}
}

@article{xu2025visulogic,
  title={Visulogic: A benchmark for evaluating visual reasoning in multi-modal large language models},
  author={Xu, Weiye and Wang, Jiahao and Wang, Weiyun and Chen, Zhe and Zhou, Wengang and Yang, Aijun and Lu, Lewei and Li, Houqiang and Wang, Xiaohua and Zhu, Xizhou and others},
  journal={arXiv preprint arXiv:2504.15279},
  year={2025}
}

@article{zeng2026vision,
  title={Vision-deepresearch benchmark: Rethinking visual and textual search for multimodal large language models},
  author={Zeng, Yu and Huang, Wenxuan and Fang, Zhen and Chen, Shuang and Shen, Yufan and Cai, Yishuo and Wang, Xiaoman and Yin, Zhenfei and Chen, Lin and Chen, Zehui and others},
  journal={arXiv preprint arXiv:2602.02185},
  year={2026}
}

@article{xing2025see,
  title={See the Text: From Tokenization to Visual Reading},
  author={Xing, Ling and Wang, Alex Jinpeng and Yan, Rui and Qu, Hongyu and Li, Zechao and Tang, Jinhui},
  journal={arXiv preprint arXiv:2510.18840},
  year={2025}
}

@article{wei2025deepseek,
  title={Deepseek-ocr: Contexts optical compression},
  author={Wei, Haoran and Sun, Yaofeng and Li, Yukun},
  journal={arXiv preprint arXiv:2510.18234},
  year={2025}
}

@article{cheng2025glyph,
  title={Glyph: Scaling context windows via visual-text compression},
  author={Cheng, Jiale and Liu, Yusen and Zhang, Xinyu and Fei, Yulin and Hong, Wenyi and Lyu, Ruiliang and Wang, Weihan and Su, Zhe and Gu, Xiaotao and Liu, Xiao and others},
  journal={arXiv preprint arXiv:2510.17800},
  year={2025}
}

@article{li2025text,
  title={Text or pixels? it takes half: On the token efficiency of visual text inputs in multimodal llms},
  author={Li, Yanhong and Lan, Zixuan and Zhou, Jiawei},
  journal={arXiv preprint arXiv:2510.18279},
  year={2025}
}

@inproceedings{duan2024vlmevalkit,
  title={Vlmevalkit: An open-source toolkit for evaluating large multi-modality models},
  author={Duan, Haodong and Yang, Junming and Qiao, Yuxuan and Fang, Xinyu and Chen, Lin and Liu, Yuan and Dong, Xiaoyi and Zang, Yuhang and Zhang, Pan and Wang, Jiaqi and others},
  booktitle={Proceedings of the 32nd ACM international conference on multimedia},
  pages={11198--11201},
  year={2024}
}

@article{wu2025qwen,
  title={Qwen-image technical report},
  author={Wu, Chenfei and Li, Jiahao and Zhou, Jingren and Lin, Junyang and Gao, Kaiyuan and Yan, Kun and Yin, Sheng-ming and Bai, Shuai and Xu, Xiao and Chen, Yilei and others},
  journal={arXiv preprint arXiv:2508.02324},
  year={2025}
}

@article{li2023text,
  title={Text as Images: Can Multimodal Large Language Models Follow Printed Instructions in Pixels?},
  author={Li, Xiujun and Lu, Yujie and Gan, Zhe and Gao, Jianfeng and Wang, William Yang and Choi, Yejin},
  journal={arXiv preprint arXiv:2311.17647},
  year={2023}
}

@article{an2025voqa,
  title={VoQA: Visual-only Question Answering},
  author={An, Jianing and Jiang, Luyang and Luo, Jie and Wu, Wenjun and Huang, Lei},
  journal={arXiv preprint arXiv:2505.14227},
  year={2025}
}

@inproceedings{smith2007overview,
  title={An overview of the Tesseract OCR engine},
  author={Smith, Ray},
  booktitle={Ninth international conference on document analysis and recognition (ICDAR 2007)},
  volume={2},
  pages={629--633},
  year={2007},
  organization={IEEE}
}

@article{cui2026paddleocr,
  title={PaddleOCR-VL-1.5: Towards a Multi-Task 0.9 B VLM for Robust In-the-Wild Document Parsing},
  author={Cui, Cheng and Sun, Ting and Liang, Suyin and Gao, Tingquan and Zhang, Zelun and Liu, Jiaxuan and Wang, Xueqing and Zhou, Changda and Liu, Hongen and Lin, Manhui and others},
  journal={arXiv preprint arXiv:2601.21957},
  year={2026}
}


\appendix

\onecolumn

\section*{Appendix}

\section{Benchmark Details}

\subsection{Complete Category Taxonomy}

The VISTA-Bench dataset is organized into four primary categories, each targeting distinct capabilities under visualized text. Table~\ref{tab:dataset_distribution_v2} provides the hierarchical distribution of samples across categories and sub-domains. Below we describe each category in detail, including instance counts, evaluation dimensions and sub-domain composition.

\noindent $\bullet$ \textbf{Unimodal Knowledge.}
This category contains 500 instances (33.33\% of the dataset) designed to isolate the model’s ability to retrieve and apply knowledge from visualized text alone, without reliance on any accompanying images. Instances are generated by transforming textual knowledge bases into visualized text while preserving semantic content. Evaluation spans two main domains: \textbf{Natural \& Life Sciences} (276 instances) covering Physics, Mathematics \& Statistics (93), Chemistry, Computing \& Geography (91) and Biology \& Medical Sciences (92); and \textbf{Social \& Applied Sciences} (224 instances) including History, Politics \& Law (77), Logic, Ethics \& Philosophy (62) and Economics, Business \& Society (85). This category specifically tests whether performance bottlenecks arise from challenges in decoding high-density visualized text.

\noindent $\bullet$ \textbf{Multimodal Knowledge.}
Featuring 400 instances (26.67\% of the dataset), this category evaluates how effectively models integrate visual evidence with factual and domain-specific knowledge when processing visualized text alongside the corresponding problem image. The instances are grouped into two primary domains: \textbf{STEM \& Health} (228 instances), which include Physical Sciences \& Engineering (88), Chemistry \& Materials (30) and Medical \& Life Sciences (110); and \textbf{Social-Humanities \& Management} (172 instances), encompassing Humanities \& Arts (47), Economics \& Management (79) and Social Sciences (46). This category assesses the model’s ability to retrieve and apply knowledge in context-rich multimodal settings, measuring how visual grounding affects domain-specific reasoning.

\noindent $\bullet$ \textbf{Multimodal Perception.}
This category consists of 300 instances (20.00\% of the dataset) designed to probe the model’s perceptual grounding capabilities under visualized text and problem images. The evaluation is organized into three dimensions: \textbf{Global Perception} (89 instances), assessing scene understanding and coarse-grained context (Coarse Perception 48, Scene Understanding 41); \textbf{Instance Perception} (103 instances), targeting the identification, localization and instance-level text understanding (Fine-grained Perception 54, Instance Identity 28, Instance Location 20, Text Understanding 1); and \textbf{Attribute Perception} (108 instances), evaluating sensitivity to object or textual attributes (Attribute Reasoning 45, Instance Attributes 63). This setup allows a detailed analysis of how well models can capture fine-grained textual and visual information under dense, multimodal inputs.

\noindent $\bullet$ \textbf{Multimodal Reasoning.}
Comprising 300 instances (20.00\% of the dataset), this category investigates the model’s reasoning capabilities under visualized text paired with problem images. Evaluation is divided into three dimensions: \textbf{Logical Reasoning} (68 instances), including Logic Reasoning (60) and Visual Reasoning (8); \textbf{Spatial \& Relation} (69 instances), covering Relation Reasoning (38), Spatial Relation (28) and Instance Interaction (3); and \textbf{Cross-Instance} (163 instances), focusing on aggregation and association across multiple visual elements (Fine-grained Perception 60, Instances Counting 103). This category measures the model’s capacity to execute multi-step, pixel-grounded reasoning that integrates information across visualized text and imagery.

\subsection{Visualized-Text Question Statistics}

To characterize the visualized-text representations in VISTA-Bench, we provide a detailed analysis of the question images across all 1,500 instances. In the final dataset, there are 1,462 multiple-choice questions and 38 open questions. Each question is rendered into a high-resolution visualized-text image, capturing the full textual content in a pixel-based format. We also observe that the open subset is generally more challenging. Although the random-guess baseline for multiple-choice questions is 25.77\%, the evaluated models perform well above chance, making the gap with open questions more meaningful. For example, for Gemini-3.1-Pro-Preview, the multiple-choice accuracy is 81.94\% (T) and 81.33\% (VT), while the open-question accuracy drops to 73.68\% (T) and 50.00\% (VT).

Table~\ref{tab:image_summary} summarizes the basic statistics of these images, including height, width and aspect ratio. All images are standardized to a fixed width of 800 pixels, while heights vary from 88 to 7,683 pixels, with a mean height of 351.9 pixels and a median of 187 pixels. The resulting aspect ratios (width divided by height) span from 0.1 to 9.1, reflecting the diversity in formatting across the dataset.

Table~\ref{tab:geometric_distribution} provides a finer-grained distribution analysis. The majority of visualized-text question images (82.9\%) have heights below 500 pixels, with only a small fraction extending beyond 2,000 pixels. The aspect ratio distribution indicates a strong horizontal orientation: over half of the images (52.9\%) have a width-to-height ratio of 4.0 or greater, whereas vertical or nearly square layouts are rare.

These statistics highlight the variability in the visualized-text question layouts in VISTA-Bench, including a wide range of heights, aspect ratios and text densities. 

\subsection{Benchmark Scale and Stability}
\label{app:scale_stability}

VISTA-Bench contains 1,500 questions, which we choose as a practical balance between evaluation cost, sample difficulty, and result stability. We first construct a larger candidate pool of approximately 4,000 samples, and then filter it through model evaluation, rendering verification, and manual review.

\begin{table*}[t]
\centering
\caption{Hierarchical Distribution of Dataset Categories: Comparison of Multimodal and Unimodal}
\label{tab:dataset_distribution_v2}
\small
\vspace{0.5em}
\setlength{\tabcolsep}{4pt}
\renewcommand{\arraystretch}{1.03}

\begin{minipage}[t]{0.49\textwidth}
\centering
\resizebox{\linewidth}{!}{
\begin{tabular}{lcc}
\toprule
\textbf{Category/Sub-category} & \textbf{Count} & \textbf{Percentage(\%)}\\
\midrule
\textbf{Multimodal Knowledge} & \textbf{400} & \textbf{26.67} \\
\quad STEM \& Health & 228 & 15.20 \\
\quad\quad \textit{Phys. Sci. \& Eng.} & 88 & 5.87 \\
\quad\quad \textit{Chem. \& Materials} & 30 & 2.00 \\
\quad\quad \textit{Med. \& Life Sci.} & 110 & 7.33 \\
\quad Social-Hum. \& Manag. & 172 & 11.47 \\
\quad\quad \textit{Humanities \& Arts} & 47 & 3.13 \\
\quad\quad \textit{Econ. \& Manag.} & 79 & 5.27 \\
\quad\quad \textit{Social Sciences} & 46 & 3.07 \\
\midrule
\textbf{Multimodal Perception} & \textbf{300} & \textbf{20.00} \\
\quad Attribute Perception & 108 & 7.20 \\
\quad\quad \textit{Attr. Reasoning} & 45 & 3.00 \\
\quad\quad \textit{Inst. Attributes} & 63 & 4.20 \\
\quad Instance Perception & 103 & 6.87 \\
\quad\quad \textit{Fine-grained (Inst.)} & 54 & 3.60 \\
\quad\quad \textit{Inst. Identity} & 28 & 1.87 \\
\quad\quad \textit{Inst. Location} & 20 & 1.33 \\
\quad\quad \textit{Text Understanding} & 1 & 0.07 \\
\quad Global Perception & 89 & 5.93 \\
\quad\quad \textit{Coarse Perception} & 48 & 3.20 \\
\quad\quad \textit{Scene Underst.} & 41 & 2.73 \\
\bottomrule
\end{tabular}
}
\end{minipage}
\hfill
\begin{minipage}[t]{0.49\textwidth}
\centering
\resizebox{\linewidth}{!}{
\begin{tabular}{lcc}
\toprule
\textbf{Category/Sub-category} & \textbf{Count} & \textbf{Percentage(\%)}\\
\midrule
\textbf{Unimodal Knowledge} & \textbf{500} & \textbf{33.33} \\
\quad Natural \& Life Sciences & 276 & 18.40 \\
\quad\quad \textit{Phys, Math \& Stat} & 93 & 6.20 \\
\quad\quad \textit{Chem, Comp \& Geo} & 91 & 6.07 \\
\quad\quad \textit{Bio \& Med Sci} & 92 & 6.13 \\
\quad Social \& Applied Sci. & 224 & 14.93 \\
\quad\quad \textit{Hist, Pol \& Law} & 77 & 5.13 \\
\quad\quad \textit{Logic, Eth \& Phil} & 62 & 4.13 \\
\quad\quad \textit{Econ, Bus \& Soc} & 85 & 5.67 \\
\midrule
\textbf{Multimodal Reasoning} & \textbf{300} & \textbf{20.00} \\
\quad Cross-Instance & 163 & 10.87 \\
\quad\quad \textit{Fine-grained (Cross)} & 60 & 4.00 \\
\quad\quad \textit{Inst. Counting} & 103 & 6.87 \\
\quad Spatial \& Relation & 69 & 4.60 \\
\quad\quad \textit{Rel. Reasoning} & 38 & 2.53 \\
\quad\quad \textit{Spatial Relation} & 28 & 1.87 \\
\quad\quad \textit{Inst. Interaction} & 3 & 0.20 \\
\quad Logical Reasoning & 68 & 4.53 \\
\quad\quad \textit{Logic Reasoning} & 60 & 4.00 \\
\quad\quad \textit{Visual Reasoning} & 8 & 0.53 \\
\addlinespace[1.1em]
\bottomrule
\end{tabular}
}
\end{minipage}

\end{table*}

\begin{table*}[!t]
\centering
\caption{Geometric Distribution Analysis of Dataset Image Properties}
\label{tab:geometric_distribution}
\small
\vspace{0.5em}
\setlength{\tabcolsep}{5pt}
\renewcommand{\arraystretch}{1.05}

\begin{minipage}[t]{0.48\textwidth}
\centering
\resizebox{\linewidth}{!}{
\begin{tabular}{lcc}
\toprule
\textbf{Height Distribution (Pixels)} & \textbf{Count} & \textbf{Percentage(\%)}\\
\midrule
$[0, 500)$         & 1244 & 82.9 \\
$[500, 1000)$      & 193  & 12.9 \\
$[1000, 2000)$     & 39   & 2.6  \\
$[2000, 3000)$     & 15   & 1.0  \\
$[3000, +\infty)$  & 9    & 0.6  \\
\bottomrule
\end{tabular}
}
\end{minipage}
\hfill
\begin{minipage}[t]{0.48\textwidth}
\centering
\resizebox{\linewidth}{!}{
\begin{tabular}{lcc}
\toprule
\textbf{Aspect Ratio (W/H)} & \textbf{Count} & \textbf{Percentage(\%)}\\
\midrule
$< 0.5$ (Vertical)       & 30   & 2.0  \\
$[0.5, 1.0)$             & 67   & 4.5  \\
$[1.0, 2.0)$             & 268  & 17.9 \\
$[2.0, 4.0)$             & 341  & 22.7 \\
$\ge 4.0$ (Horizontal)   & 794  & 52.9 \\
\bottomrule
\end{tabular}
}
\end{minipage}

\end{table*}

\begin{table}[htbp]
\centering
\caption{Summary Statistics of Image Dimensions ($N=1500$)}
\label{tab:image_summary}
\small
\vspace{0.5em}
\setlength{\tabcolsep}{7pt}
\begin{tabular}{l ccccc}
\toprule
\textbf{Dimension} & {\textbf{Min}} & {\textbf{Max}} & {\textbf{Mean}} & {\textbf{Median}} & {\textbf{Std}} \\ 
\midrule
Height (px)       & 88.0  & 7683.0 & 351.9 & 187.0 & 462.1 \\
Width (px)        & 800.0 & 800.0  & 800.0 & 800.0 & 0.0   \\
Aspect Ratio (W/H)& 0.1   & 9.1    & 3.8   & 4.3   & 1.9   \\
\bottomrule
\end{tabular}
\end{table}

\begin{table}[htbp]
\centering
\caption{Sample-size adequacy analysis under different benchmark scales.}
\label{tab:scale_stability}
\small
\vspace{0.5em}
\setlength{\tabcolsep}{8pt}
\renewcommand{\arraystretch}{1.08}
\begin{tabular}{llccc}
\toprule
\textbf{Model} & \textbf{Input} & \textbf{500} & \textbf{1000} & \textbf{1500} \\
\midrule
\multirow{2}{*}{Qwen3-VL-8B-Instruct}
& T  & 53.6 & 57.1 & 59.1 \\
& VT & 41.8 & 49.0 & 52.3 \\
\midrule
\multirow{2}{*}{InternVL3.5-8B}
& T  & 53.8 & 59.6 & 59.3 \\
& VT & 50.0 & 55.0 & 50.2 \\
\bottomrule
\end{tabular}
\end{table}

We further conduct a scale-sensitivity analysis using two representative VLMs. As shown in Table~\ref{tab:scale_stability}, smaller subsets can lead to noticeable fluctuations in both absolute accuracy and the estimated modality gap. For example, InternVL3.5-8B obtains 55.0 VT accuracy on the 1,000-sample subset, but 50.2 on the full 1,500-sample benchmark, suggesting that smaller subsets may give optimistic or composition-dependent estimates. With 1,500 questions, Qwen3-VL-8B-Instruct and InternVL3.5-8B show similar pure-text accuracy, while Qwen3-VL-8B-Instruct has a smaller visualized-text drop. This trend is consistent with its stronger OCR-related capability. These results suggest that the 1,500-question setting provides a more stable basis for model-level comparison, while remaining practical for evaluating many VLMs.

\subsection{Examples and Case Studies}

\begin{figure}[p]
    \centering
    \includegraphics[width=0.78\textwidth]{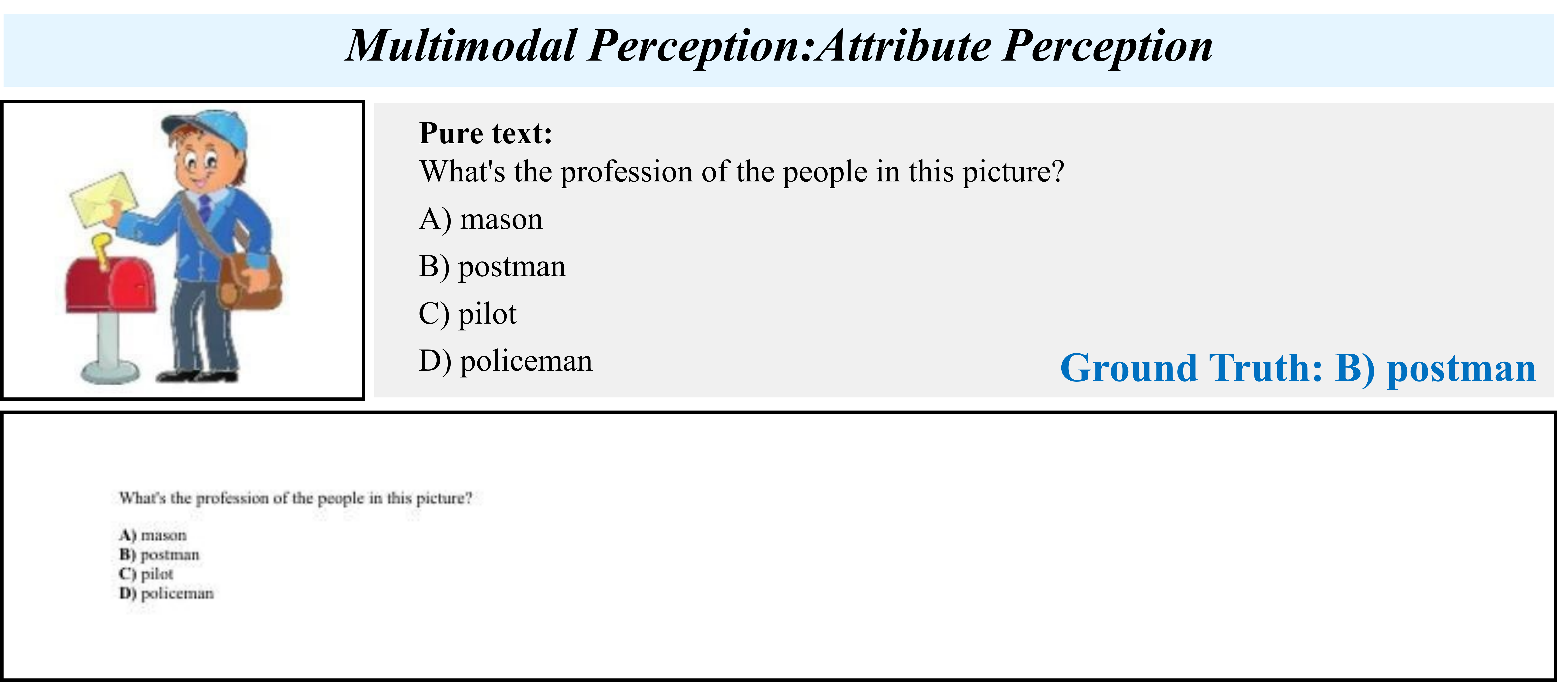} \\[1em] 
    \includegraphics[width=0.78\textwidth]{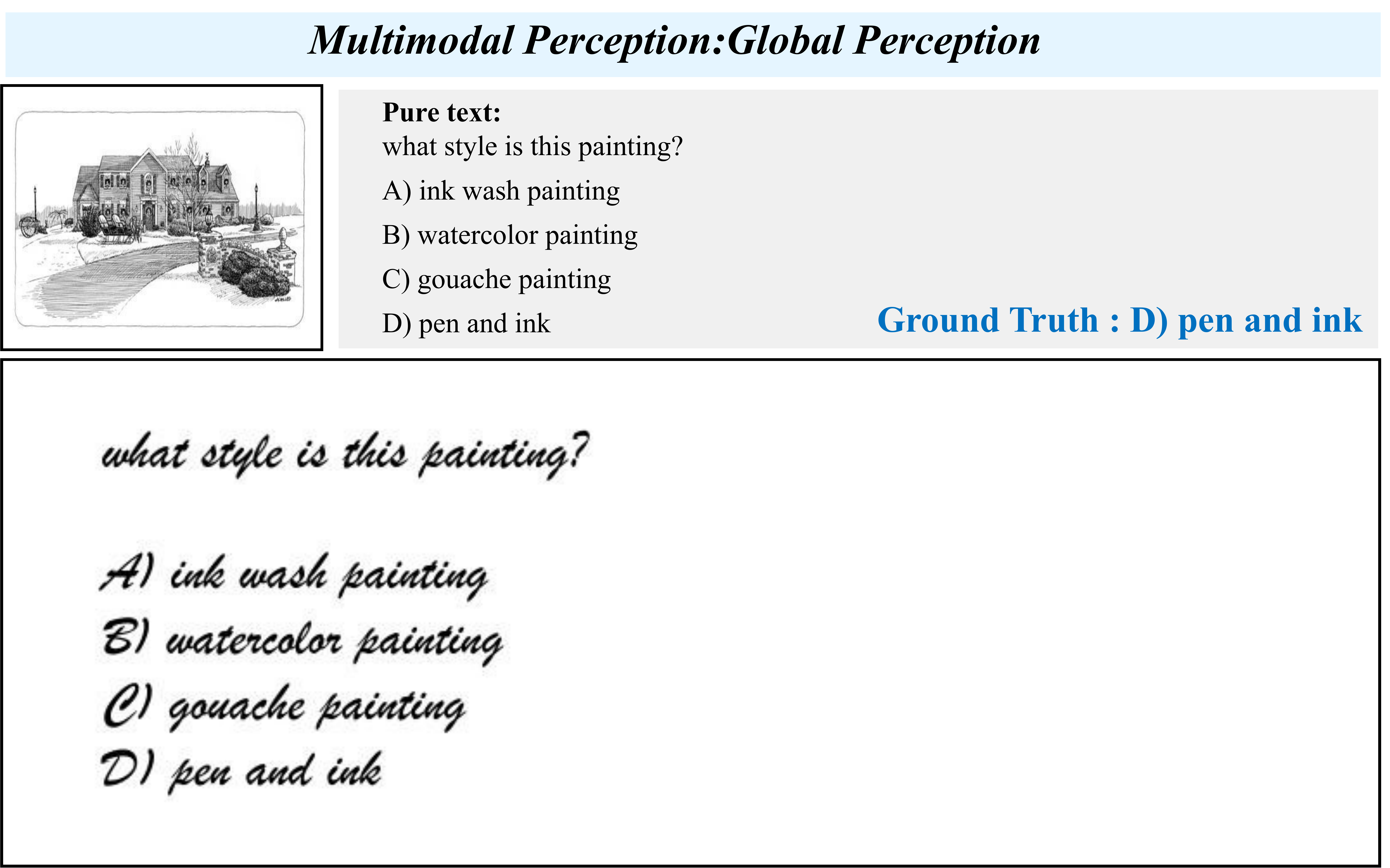} \\[1em]
    \includegraphics[width=0.78\textwidth]{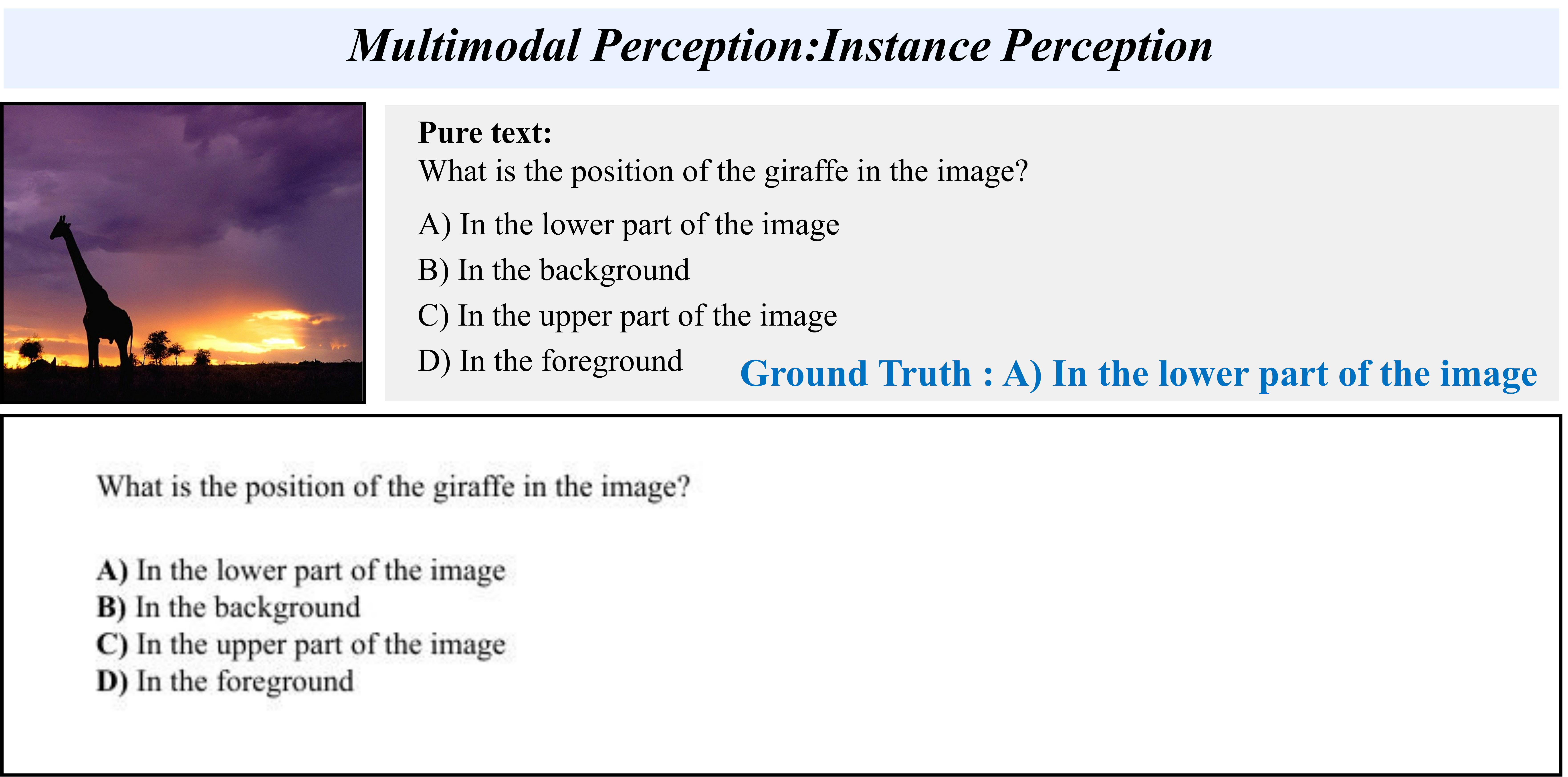}
     \caption{Visualized examples for \textbf{Multimodal Perception} task. 
    Top: Attribute Perception (\textit{Times New Roman}, 9pt). 
    Middle: Global Perception (\textit{Brush Script MT}, 32pt). 
    Bottom: Instance Perception (\textit{Times New Roman}, 16pt).}
    \label{fig:case1}
\end{figure}

\begin{figure}[p]
    \centering
    \includegraphics[width=0.78\textwidth]{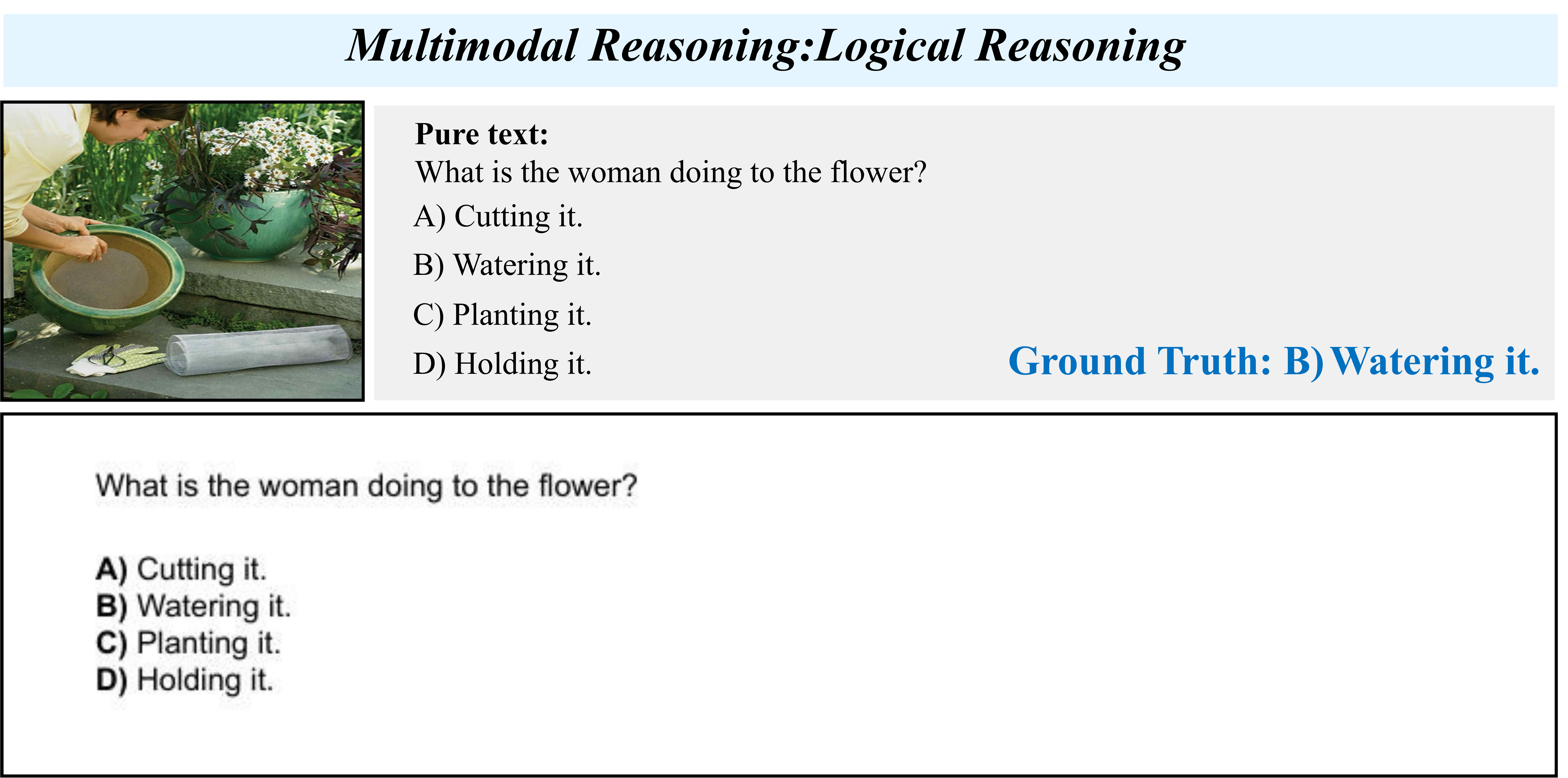} \\[1em]
    \includegraphics[width=0.78\textwidth]{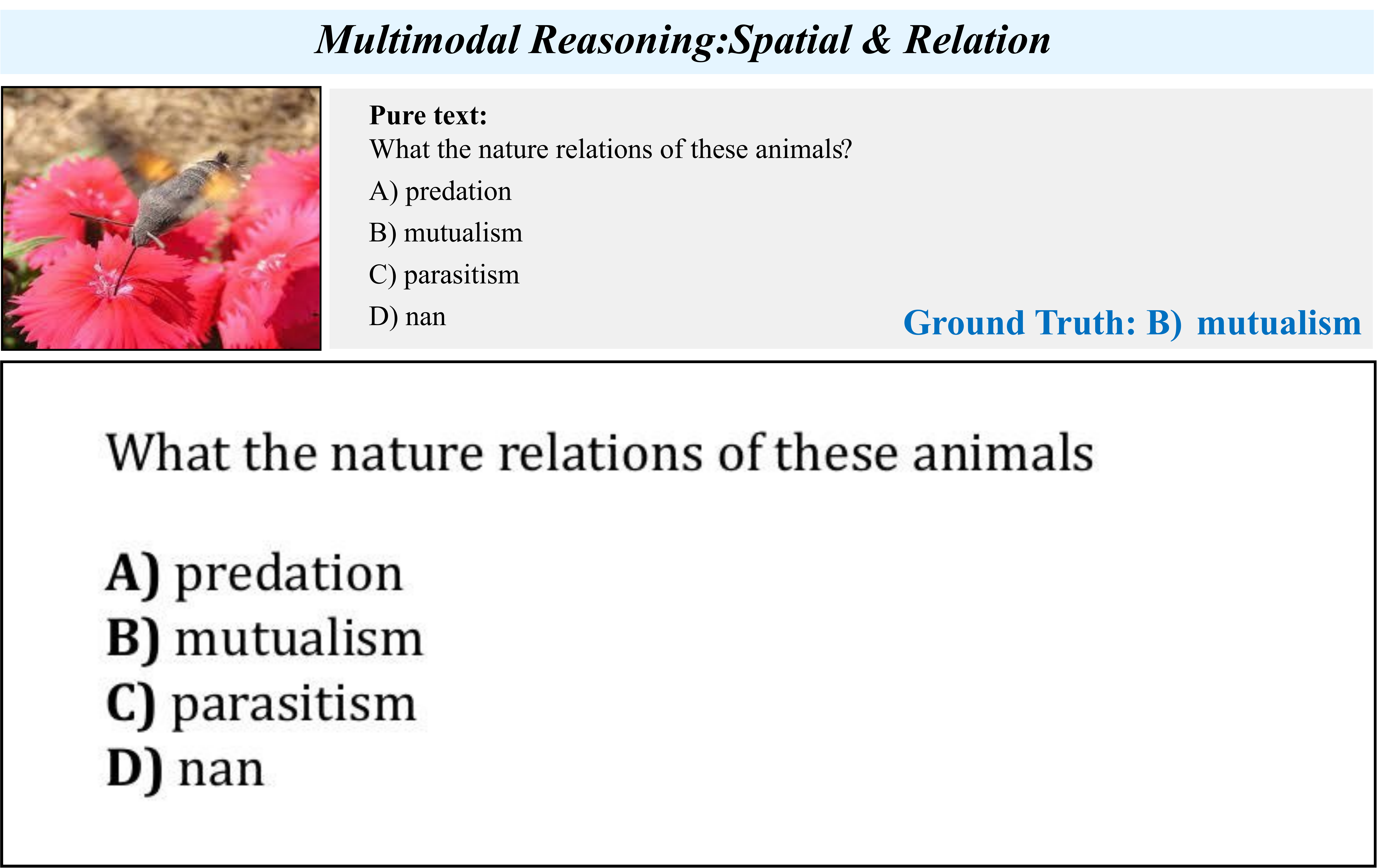} \\[1em]
    \includegraphics[width=0.78\textwidth]{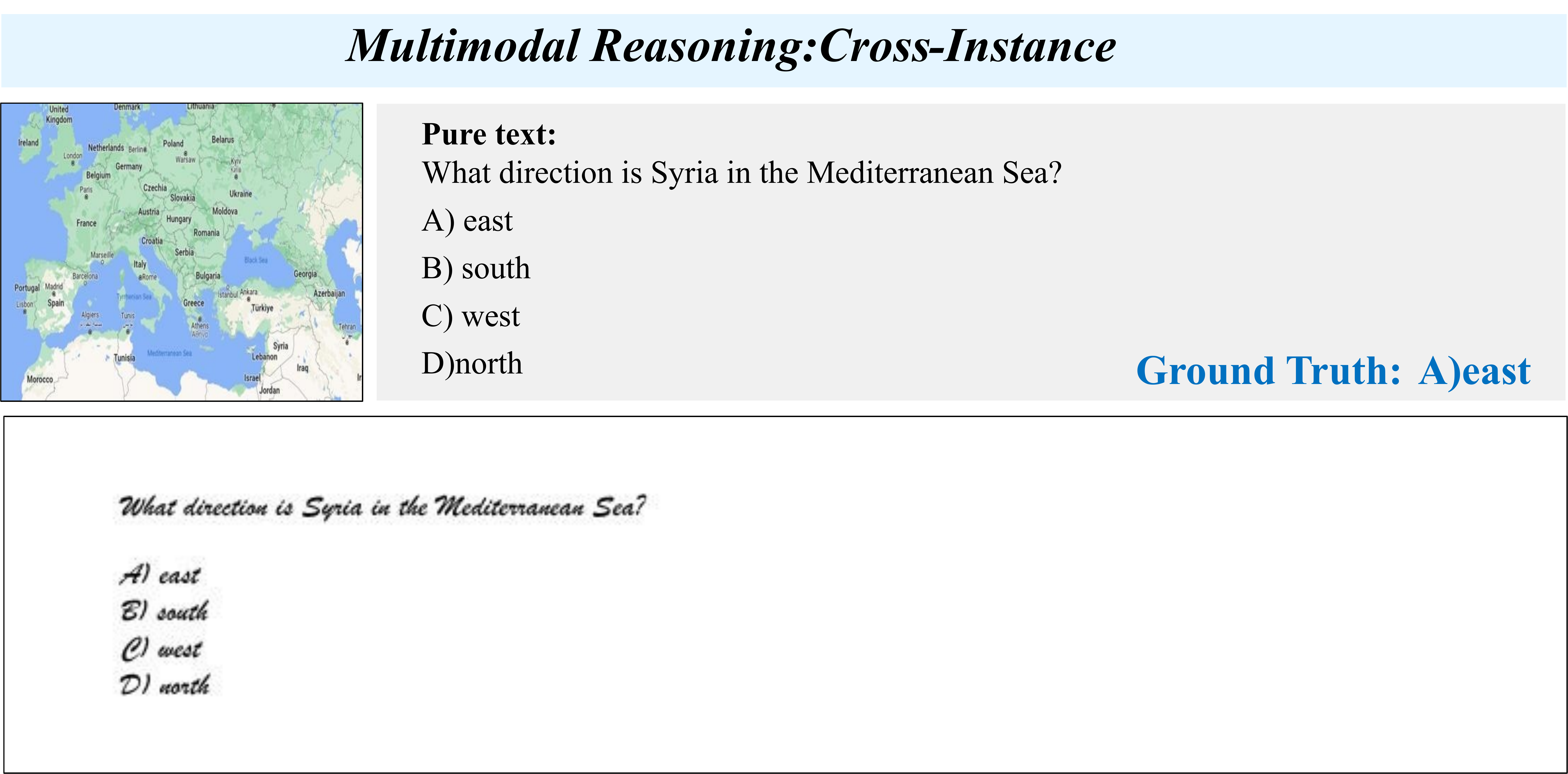}
    \caption{Visualized examples for \textbf{Multimodal Reasoning} task. 
    Top: Logical Reasoning (\textit{Arial}, 16pt). 
    Middle: Spatial \& Relation (\textit{Cambria}, 32pt). 
    Bottom: Cross-Instance (\textit{Cambria}, 48pt).}
    \label{fig:case2}
\end{figure}

\begin{figure}[p]
    \centering
    \includegraphics[width=0.78\textwidth]{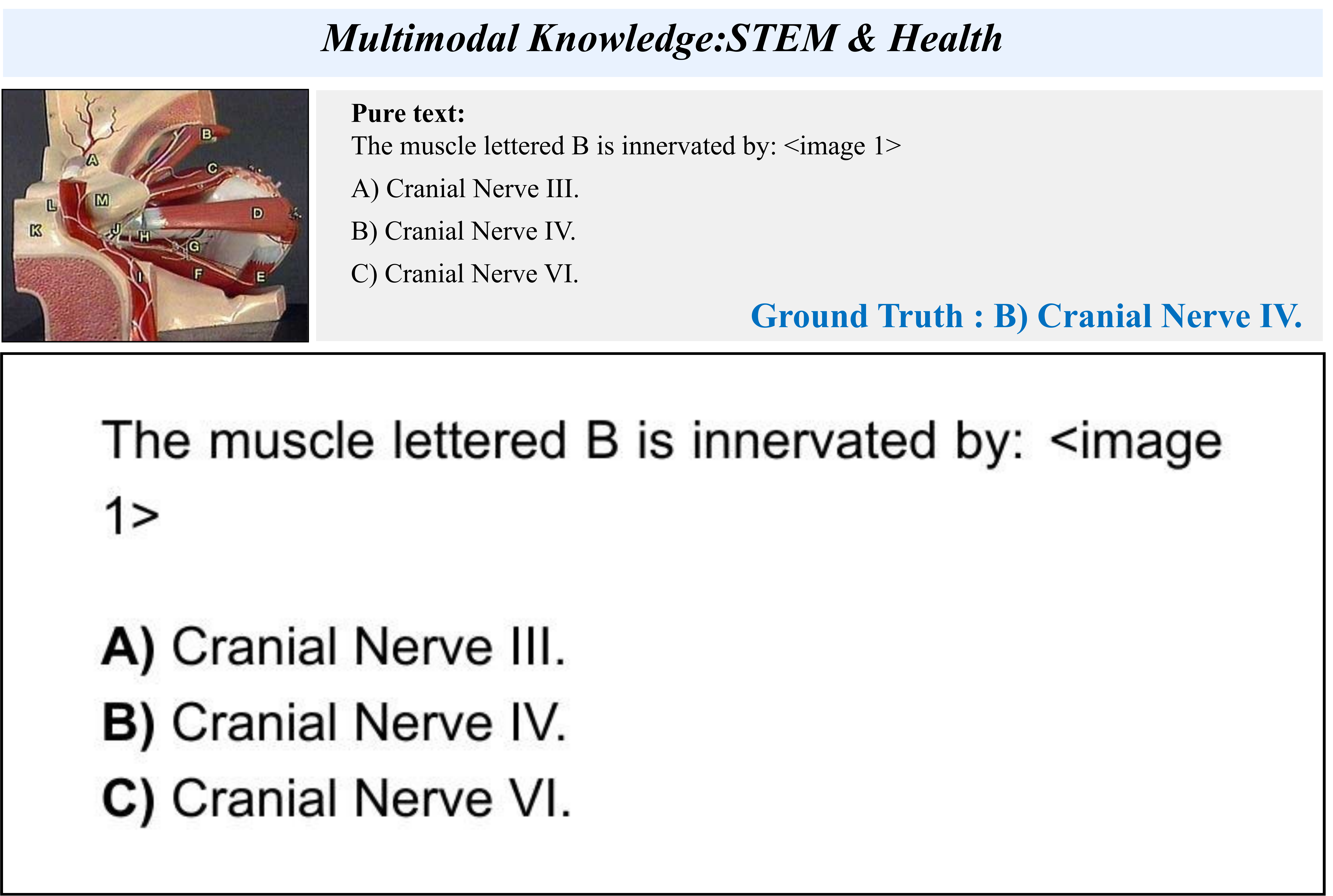} \\[1em]
    \includegraphics[width=0.78\textwidth]{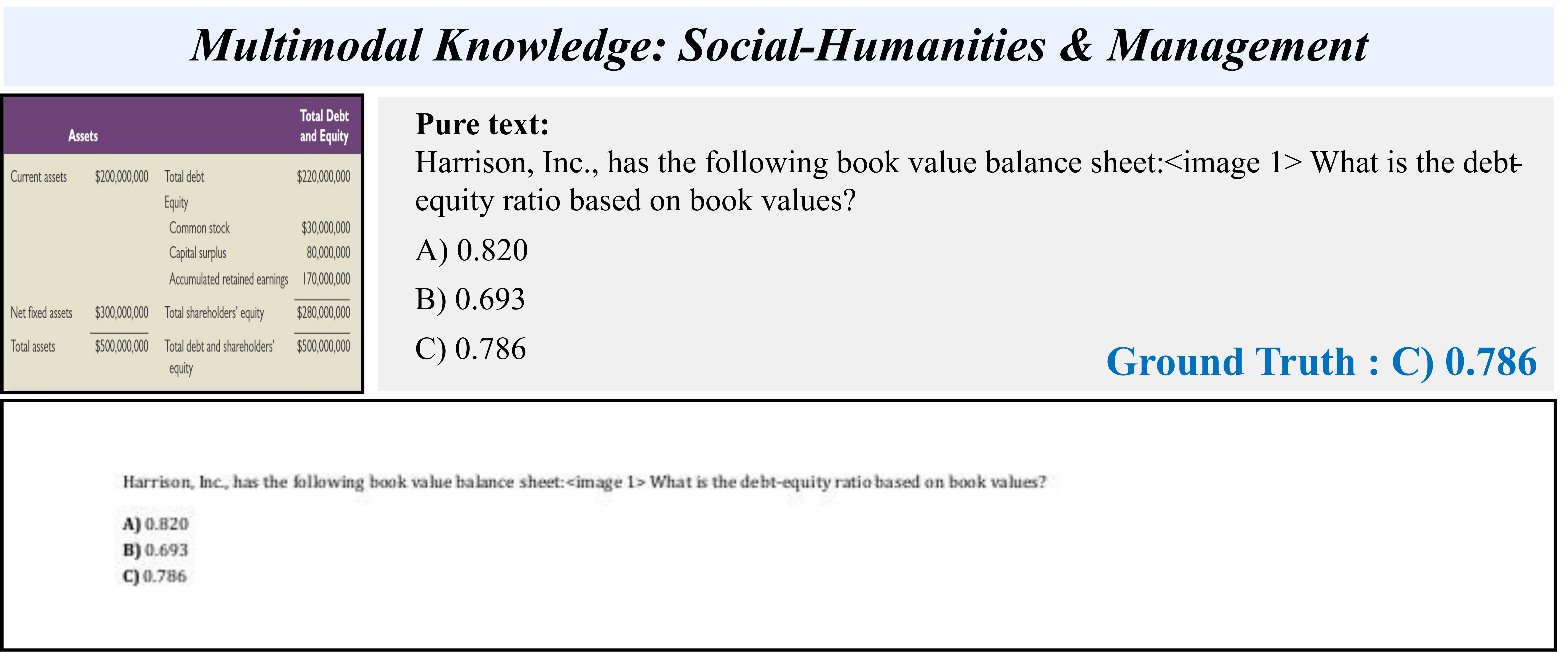} \\[1em]
    \includegraphics[width=0.78\textwidth]{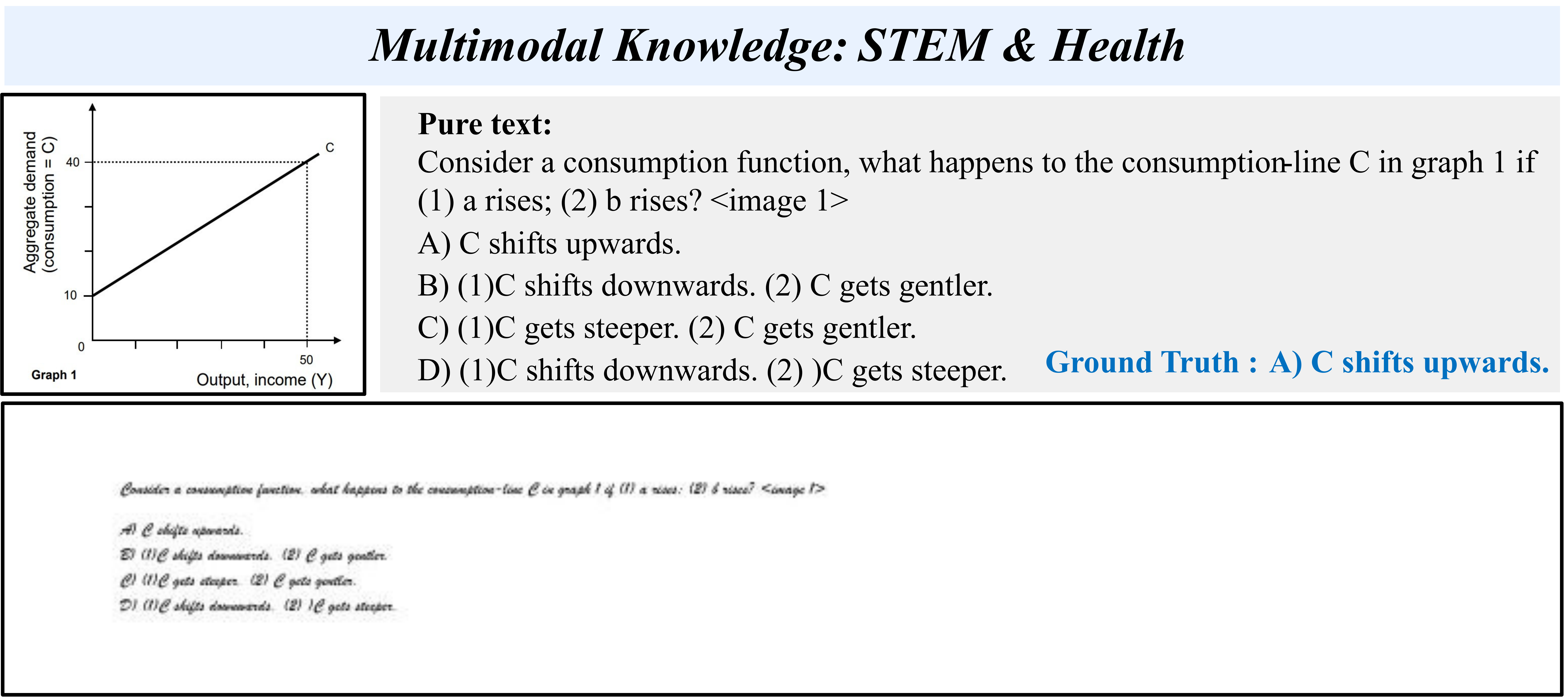}
    \caption{Visualized examples for \textbf{Multimodal Knowledge} task. 
    Top: STEM \& Health (\textit{Arial}, 32pt). 
    Middle: Social-Humanities \& Management (\textit{Cambria}, 9pt).
    Bottom: STEM \& Health (\textit{Brush Script MT}, 9pt). }
    \label{fig:case3}
\end{figure}

\begin{figure}[p]
    \centering
    \includegraphics[width=0.78\textwidth]{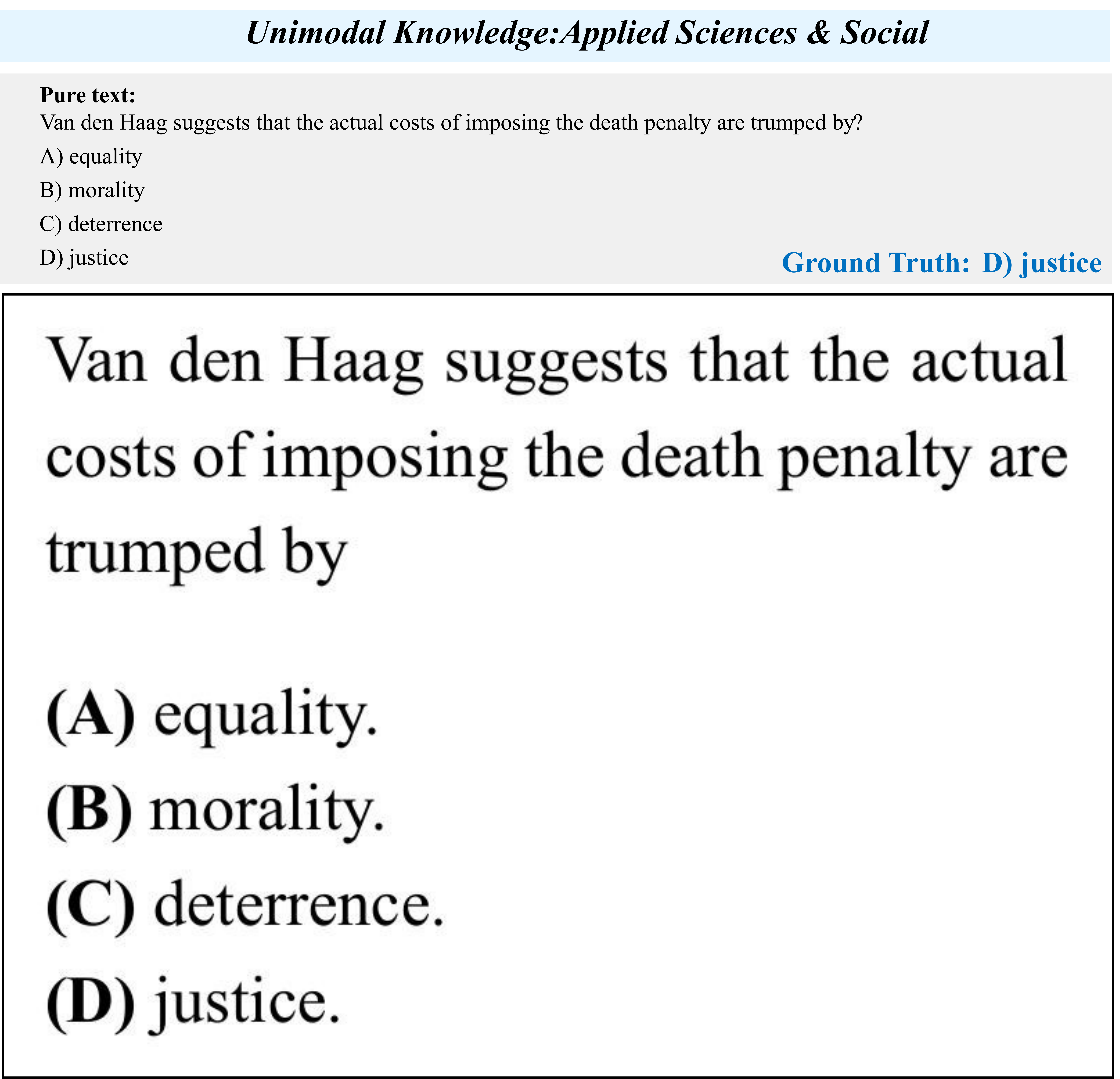} \\[1em]
    \includegraphics[width=0.78\textwidth]{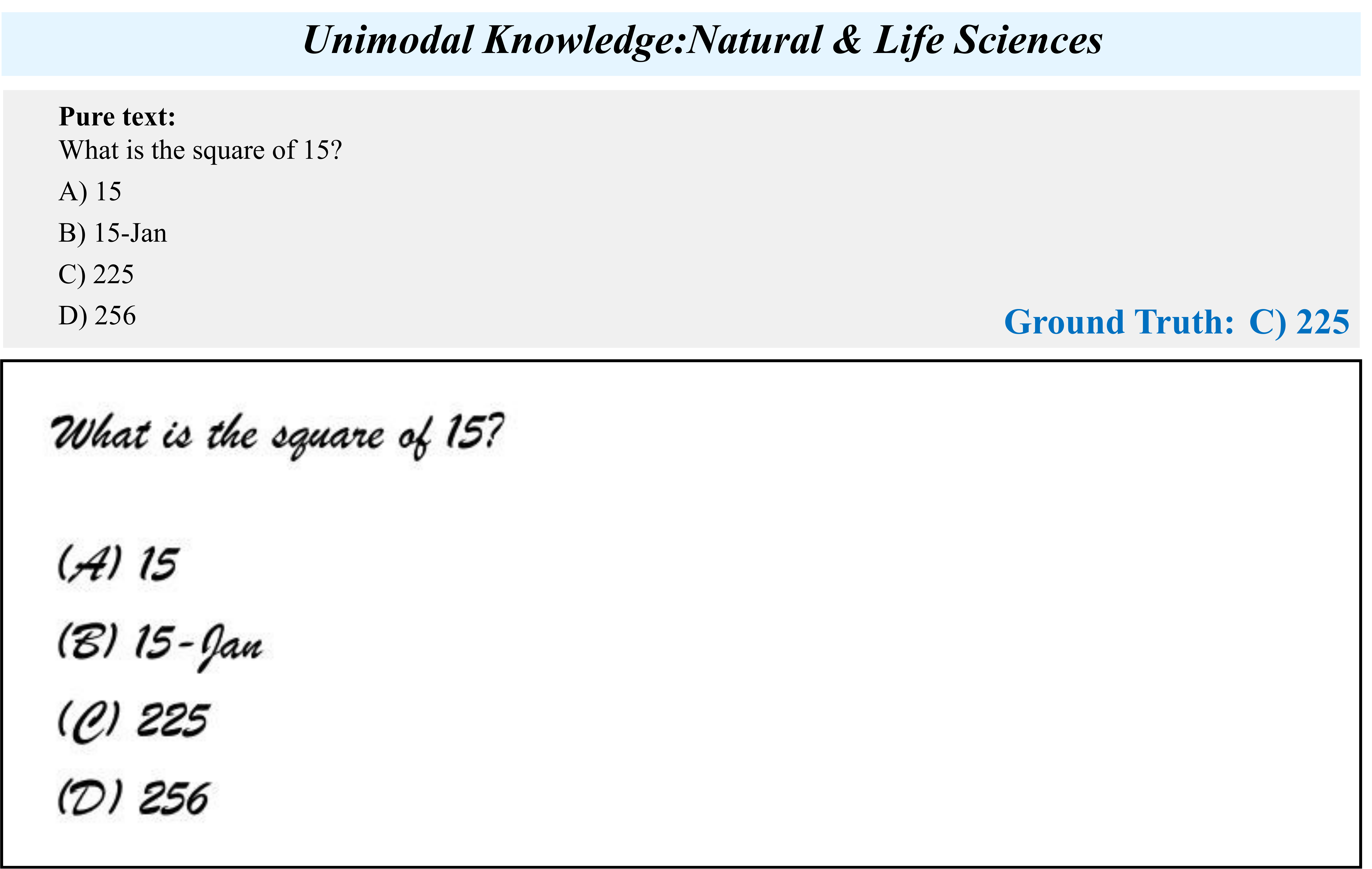}
    \caption{Visualized examples for \textbf{Unimodal Knowledge} task. 
    Top: Applied Sciences \& Social (\textit{Times New Roman}, 48pt). 
    Bottom: Natural \& Life Sciences (\textit{Brush Script MT}, 32pt).}
    \label{fig:case4}
\end{figure}

To provide a concrete understanding of VISTA-Bench, we present representative cases spanning different task categories, rendering configurations and modality settings. 
\textbf{Samples are shown in Figure~\ref{fig:case1} through Figure~\ref{fig:case4} at the end.}

\clearpage
\section{Rendering Pipeline}

\subsection{Pipeline Detailed}

To ensure high-fidelity and diverse visual representations of textual knowledge, VISTA-Bench adopts a rigorous rendering pipeline that transforms discrete linguistic tokens into pixel-based visual inputs. The pipeline is designed to preserve typographic fidelity while simulating the visual characteristics of real-world documents. 

\textbf{Text Preprocessing.}  
To achieve semantic accuracy, the pipeline employs a multi-layered preprocessing workflow that reconciles raw data with LaTeX constraints through three integrated transformations. This begins with \textbf{\textit{Text Normalization}}, which standardizes typographic elements like smart quotes and em-dashes into ASCII equivalents to eliminate rendering artifacts and ensure a stable input stream. Simultaneously, \textbf{\textit{Code Isolation}} establishes a syntax-shielding barrier, distinguishing descriptive prose from mathematical environments to protect reserved characters—such as \&, \% and \_ while strictly preserving the integrity of pre-existing valid instructions. Finally, \textit{\textbf{Formula Synthesis}} performs a symbolic reconstruction, identifying logical operators and Greek letters within the text to dynamically transcode them into high-fidelity macros, while standardizing specialized contexts like temperature units and degree symbols into publication-grade mathematical notation.

\textbf{LaTeX-based Generation.}  
To achieve Visual Harmony, the pipeline utilizes LaTeX as the core engine, implementing a \textbf{\textit{Width Anchoring}} mechanism that mathematically calibrates the target pixel width into physical LaTeX points. By precisely aligning the rendering DPI with the standard typographic point scale, the pipeline locks the horizontal span at 800 pixels, ensuring consistent line-wrapping behavior and geometric stability across all generated samples. Complementing this, \textbf{\textit{Font Mapping}} dynamically resolves style names into specific OpenType font families, such as \textit{Arial} or \textit{Cambria}, while varying font sizes from 9pt to 48pt. This approach allows the pipeline to reflect the typographic diversity of real-world documents while maintaining a strictly standardized visual framework.

\textbf{Image Conversion and Post-processing.}  
The final stage of the pipeline executes Precise Extraction, a vision-based refinement process that transforms vector documents into visualized-text image assets. This begins with \textit{\textbf{Fidelity Rasterization}}, which utilizes high-DPI sampling to convert the PDF into a pixel-based format, effectively eliminating aliasing and ensuring that the sharpness of fine mathematical symbols is preserved. Subsequently, the system performs \textbf{\textit{Content Localization}} by translating the image into a grayscale matrix and executing a pixel-level scan to pinpoint the exact vertical boundaries of the rendered text. Finally, \textbf{\textit{Adaptive Cropping}} utilizes these detected coordinates to strip away redundant whitespace while automatically applying a protective margin to prevent the clipping of character descenders or complex mathematical structures. This rigorous post-processing ensures that every output image is both spatially efficient and semantically complete, providing a clean visual signal for model evaluation.

\subsection{Rendering Configuration}

To ensure reproducibility, we disclose our optimized rendering configurations. Table~\ref{tab:font_details} presents the Font Mapping distribution, which ensures typographic diversity across 1,500 samples by varying font families and sizes. As shown in Table~\ref{tab:rendering_configurations}, we identified Width Anchoring at 800 pixels and a DPI of 72.27 as the most stable configuration, guaranteeing that digital font sizes precisely align with physical typographic standards. These parameters, combined with strategic margins, prevent character truncation at large font sizes to preserve the Semantic Integrity of the visualized text.

\begin{table}[htbp]
\centering
\caption{Rendering font mapping}
\label{tab:font_details}
\small
\begin{tabular}{l ccccc}
\toprule
\textbf{font} & {\textbf{\textit{Arial}}} & {\textbf{\textit{Times New Roman}}} & {\textbf{\textit{Cambria}}} & {\textbf{\textit{Brush Script MT}}} & {\textbf{\textit{Sum}}} \\ 
\midrule
9(pt)       & 136  & 134 & 106 & 186  & 562\\
16(pt)        & 102 & 97  & 63 & 133   & 395\\
32(pt)       & 91   & 63    & 72   & 84  & 310  \\
48(pt)         & 56   & 47    & 54   & 76  & 233    \\
\textbf{\textit{Sum}}   & 385   & 341    & 295   & 479  & \textbf{1500}   \\
\bottomrule
\end{tabular}
\end{table}

\begin{table}[htbp]
\centering
\caption{Other rendering configurations}
\label{tab:rendering_configurations}
\small
\vspace{0.5em}
\begin{tabular}{l cccc}
\toprule
\textbf{Config} & {\textbf{Width}} & {\textbf{DPI}} & {\textbf{Left and Right Margin}} & {\textbf{Top and Bottom Margin}} \\ 
\midrule
value       & 800  & 72.27 & 60 & 40 \\
\bottomrule
\end{tabular}
\end{table}

\subsection{Sensitivity to Rendering Layout}
\label{app:rendering_layout_sensitivity}

The construction of VISTA-Bench relies on a rendering pipeline that transforms textual questions into visualized-text images. Although our default configuration is designed to preserve rendering fidelity and layout stability, the resulting images may still be affected by layout-related parameters. Therefore, we further investigate whether the benchmark results are sensitive to the rendering process itself.

We focus on two representative layout factors: rendering width and left/right margin. In the default configuration, the rendered question image is generated with a width of 800 pixels, a DPI of 72.27, left/right margins of 60 pixels, and top/bottom margins of 40 pixels. To isolate the effect of each factor, we conduct controlled experiments by changing one parameter at a time while keeping the others fixed. Specifically, we vary the rendering width among 512, 800, and 1024 pixels, and vary the left/right margin among 20, 60, and 100 pixels. The default setting corresponds to width 800 and margin 60.

These parameters directly affect the spatial layout of visualized text. A smaller width may introduce more line wrapping and longer question images, while a larger width may reduce line wrapping but alter the effective text density. Similarly, narrower margins may increase the risk of boundary effects or character truncation, whereas wider margins can introduce more blank space and reduce the relative area occupied by textual content. Since VLMs process the rendered question through the visual encoder, such layout changes may influence visualized-text accuracy.

Table~\ref{tab:rendering_layout_sensitivity} reports the results on two representative models, Qwen3-VL-8B-Instruct and InternVL3.5-8B. The pure-text results are unchanged because they do not depend on the rendering process. In contrast, the visualized-text results vary moderately across different rendering layouts. For Qwen3-VL-8B-Instruct, the overall visualized-text accuracy ranges from 50.1\% to 52.6\%, while the pure-text accuracy remains 59.1\%. For InternVL3.5-8B, the visualized-text accuracy ranges from 49.1\% to 52.1\%, compared with its pure-text accuracy of 59.3\%. These results indicate that rendering layout can affect the absolute performance of visualized-text inputs, but the main modality-gap trend remains consistent across all tested configurations.

\begin{table*}[t]
\centering
\caption{
Sensitivity to rendering layout configurations. 
We vary one rendering parameter at a time while keeping other settings fixed. 
The default visualized-text setting uses width 800, DPI 72.27, and left/right margin 60.
The $\downarrow$ Gap column denotes the overall performance drop relative to the pure-text setting of the same model.
}
\label{tab:rendering_layout_sensitivity}
\vspace{0.5em}
\setlength{\tabcolsep}{4.2pt}
\renewcommand{\arraystretch}{1.12}
\small
\resizebox{\textwidth}{!}{
\begin{tabular}{lcccc|cc}
\toprule
\textbf{Input Setting}
& \textbf{Multi-Per.}
& \textbf{Multi-Rea.}
& \textbf{Multi-Kno.}
& \textbf{Uni-Kno.}
& \textbf{Overall}
& \textbf{$\downarrow$ Gap} \\
\midrule

\multicolumn{7}{l}{\cellcolor{blue!8}\textbf{Qwen3-VL-8B-Instruct~\cite{bai2025qwen3vltechnicalreport}}} \\

Pure Text
& 67.3 & 49.3 & 48.5 & 68.4
& 59.1
& -- \\

Default VT $(W{=}800, M{=}60)$
& 65.3 & 49.0 & 37.8 & 58.2
& 52.3
& \textcolor{red}{-6.7} \\

VT, Width $=512$
& 63.3 & 48.0 & 35.8 & 56.2
& 50.5
& \cellcolor{red!8}\textcolor{red}{-8.6} \\

VT, Width $=1024$
& 65.7 & 48.0 & 37.5 & 56.4
& 51.5
& \cellcolor{red!6}\textcolor{red}{-7.6} \\

VT, Margin $=20$
& 65.3 & 49.7 & 39.0 & 57.6
& 52.6
& \cellcolor{red!4}\textcolor{red}{-6.5} \\

VT, Margin $=100$
& 64.7 & 45.7 & 34.5 & 56.6
& 50.1
& \cellcolor{red!10}\textcolor{red}{-9.0} \\

\midrule

\multicolumn{7}{l}{\cellcolor{blue!8}\textbf{InternVL3.5-8B~\cite{wang2025internvl3}}} \\

Pure Text
& 64.3 & 52.3 & 45.8 & 71.2
& 59.3
& -- \\

Default VT $(W{=}800, M{=}60)$
& 61.3 & 45.7 & 36.0 & 57.6
& 50.2
& \textcolor{red}{-9.1} \\

VT, Width $=512$
& 65.7 & 47.7 & 36.3 & 59.4
& 52.1
& \cellcolor{red!6}\textcolor{red}{-7.2} \\

VT, Width $=1024$
& 61.7 & 44.0 & 36.8 & 56.4
& 49.7
& \cellcolor{red!8}\textcolor{red}{-9.6} \\

VT, Margin $=20$
& 61.3 & 45.0 & 33.8 & 57.2
& 49.3
& \cellcolor{red!10}\textcolor{red}{-10.0} \\

VT, Margin $=100$
& 60.7 & 42.7 & 34.5 & 57.6
& 49.1
& \cellcolor{red!10}\textcolor{red}{-10.2} \\

\bottomrule
\end{tabular}
}
\end{table*}

Overall, this analysis shows that VISTA-Bench is not entirely invariant to rendering choices, which is expected because visualized text is processed as pixels. However, the observed modality gap is not an artifact of a single rendering configuration. Across different widths and margins, visualized-text inputs consistently underperform the corresponding pure-text inputs. Rendering parameters mainly affect the magnitude of the gap rather than overturning the benchmark's main conclusion.

\subsection{VLM Settings}

To guarantee the rendering precision of VISTA-Bench, we implement a VLM-as-Judge protocol that uses Qwen3-VL-32B-Instruct as ``Rendering Quality Auditors'' to verify pixel-level fidelity. 
Before using Qwen3-VL-32B-Instruct at scale, we first built a small validation set of 200 samples covering all three rendering quality labels. By comparing its predictions with human annotations, we found that Qwen3-VL-32B-Instruct, which has a modality gap of -8.5, achieves 75\% accuracy on this set. Importantly, most of its errors come from being overly strict. In the formal construction of VISTA-Bench, we further applied manual verification to ensure the correctness of the final accepted samples.

The System Prompt directs the judge to evaluate Text Fidelity, Code Integrity, and Formula Precision on a three-tier 0--2 scale, focusing strictly on rendering quality rather than semantic correctness. 
Complementing this, the User Prompt instructs the judge to compare the rendered image against the reference text and return an alignment score with a brief explanation for any detected issues. 
Under our Acceptance and Verification Policy, samples assigned a score below 2 are further sent for manual verification.
Through this procedure, we establish VISTA-Bench with reliable rendering fidelity.

\newtcolorbox{promptbox}[1]{
  colback=gray!8,
  colframe=gray!55,
  fonttitle=\bfseries\small,
  coltitle=black,
  title=#1,
  enhanced,
  breakable,
  boxrule=0.5pt,
  arc=2pt,
  left=6pt, right=6pt, top=6pt, bottom=6pt,
  before skip=8pt,
  after skip=8pt,
  fontupper=\small
}

\begin{promptbox}{System Prompt (Rendering Quality Auditor)}
You are a highly precise AI Rendering Quality Auditor specializing in document digitization and visual text verification.

Your task is to evaluate the alignment between a rendered visualized-text image and its corresponding ground-truth text.

Focus strictly on rendering quality rather than semantic correctness. Specifically, evaluate the following aspects:

\begin{enumerate}[leftmargin=1.5em, itemsep=2pt, topsep=3pt]
    \item \textbf{Text Fidelity:} whether all characters, words, and spacing in the image accurately match the reference text. Natural line wrapping, line-end hyphenation, or negligible layout variation caused by rendering constraints should not be treated as an error if the content remains faithful and fully readable.
    
    \item \textbf{Code Integrity:} whether code snippets preserve symbols, indentation, line breaks, and formatting without omissions or distortions.
    
    \item \textbf{Formula Precision:} whether mathematical expressions, operators, and structures are rendered faithfully.
\end{enumerate}

Based on the overall alignment, output a single alignment score according to the following scale:

\begin{itemize}[leftmargin=1.5em, itemsep=2pt, topsep=3pt]
    \item \textbf{2 (Flawless):} Perfect or near-perfect alignment with no meaningful rendering errors. Minor layout-induced variations that do not affect fidelity, correctness, or readability should still be scored as 2.
    
    \item \textbf{1 (Acceptable):} Visible but minor rendering imperfections that slightly affect visual fidelity, but do not affect readability or interpretation.
    
    \item \textbf{0 (Misaligned):} Clear rendering errors that impair correctness or readability.
\end{itemize}

Assign score 2 by default unless there is a clear visible rendering issue that justifies lowering the score.

Do not lower the score solely due to natural line wrapping, line-end hyphenation, or negligible spacing/layout differences.

Do not perform reasoning beyond rendering verification.

Do not infer missing content.

Do not assess semantic correctness.

Return JSON only in the following format:

\begin{quote}
\small
\ttfamily
\{"score": 0 or 1 or 2, "explanation": "brief explanation"\}
\end{quote}
\end{promptbox}

\begin{promptbox}{User Prompt (Rendering Verification)}
The following input consists of:

\begin{itemize}[leftmargin=1.5em, itemsep=2pt, topsep=3pt]
    \item A visualized-text image generated by a rendering pipeline.
    \item The corresponding ground-truth text content.
\end{itemize}

Please compare the image with the reference text and evaluate their alignment according to the specified criteria.

\textbf{Ground-truth text:}

\begin{quote}
\small
\ttfamily
----------------\\
\{reference\_text\}\\
----------------
\end{quote}

Return:

\begin{itemize}[leftmargin=1.5em, itemsep=2pt, topsep=3pt]
    \item A single integer score from 2, 1, or 0 indicating the alignment status.
    \item A brief explanation describing any detected rendering issues, if applicable.
    \item Return the result in JSON format with the fields \texttt{score} and \texttt{explanation}.
\end{itemize}
\end{promptbox}

\begin{promptbox}{Acceptance and Verification Policy}
We use the alignment score from Qwen3-VL-32B-Instruct to identify potential rendering issues in visualized-text instances.

\begin{itemize}[leftmargin=1.5em, itemsep=2pt, topsep=3pt]
    \item \textbf{Three-tier score:} Each rendered instance receives an alignment score from 0 to 2, based on text, code, and formula fidelity.

    \item \textbf{Manual verification:} Samples scoring below 2 are further sent for manual verification, where annotators compare the rendered image with the reference text.
\end{itemize}

This policy uses the VLM judge as an efficient rendering-quality filter, while manual verification handles samples with potential rendering issues.
\end{promptbox}

\paragraph{Remark.}
This VLM-as-Judge protocol follows the same principle as prior LLM-as-Judge paradigms, while extending it to the visual modality by explicitly verifying pixel-level rendering fidelity rather than semantic correctness.

\subsection{Rendering Cases}
\begin{figure*}[t!]
    \centering
    \includegraphics[width=0.78\textwidth]{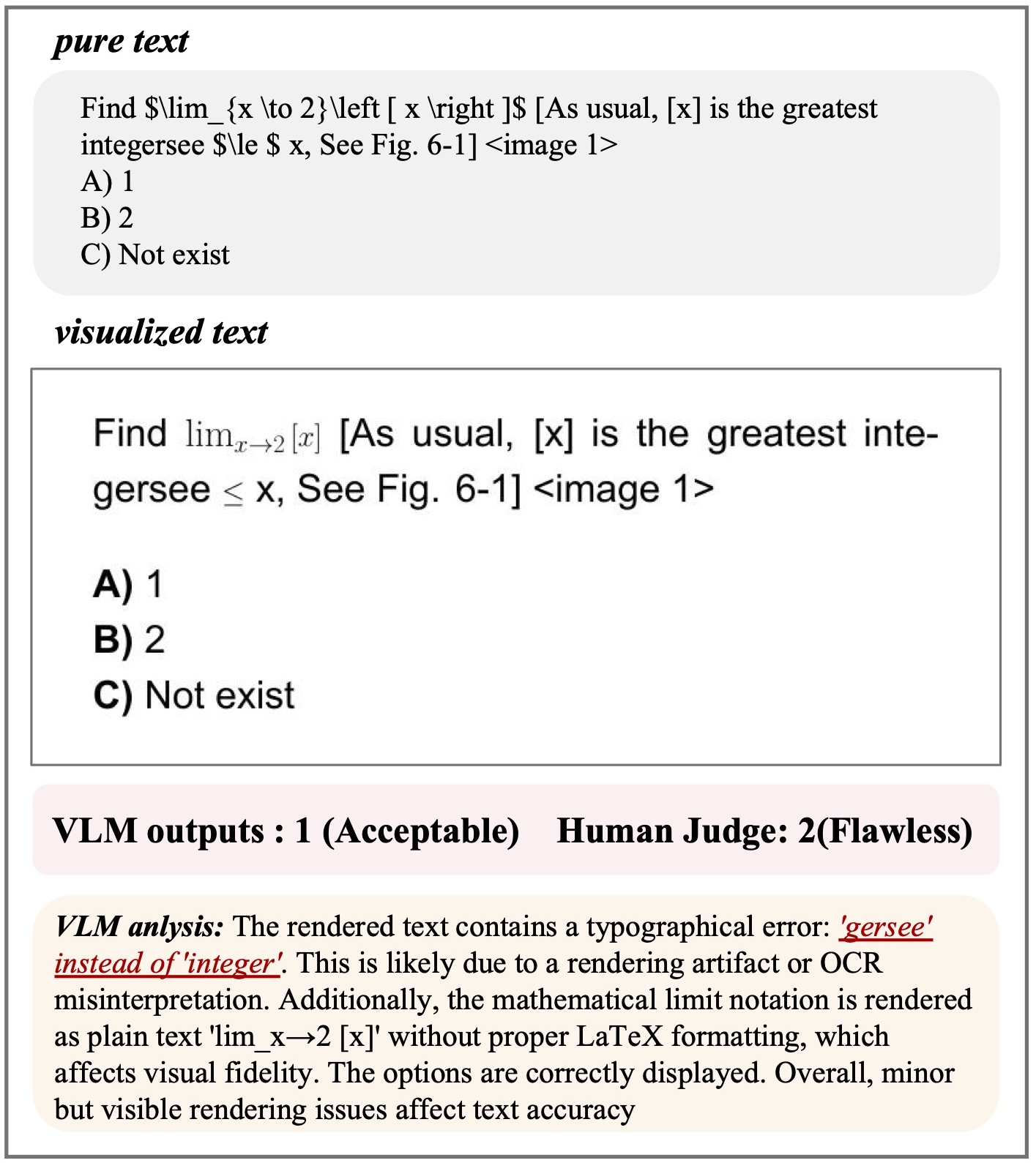}
    \caption{Character truncation recognition error. Config: \textit{Arial}, 16pt}
    \label{fig:judge-case1}
\end{figure*}

\begin{figure*}[t!]
    \centering
    \includegraphics[width=0.78\textwidth]{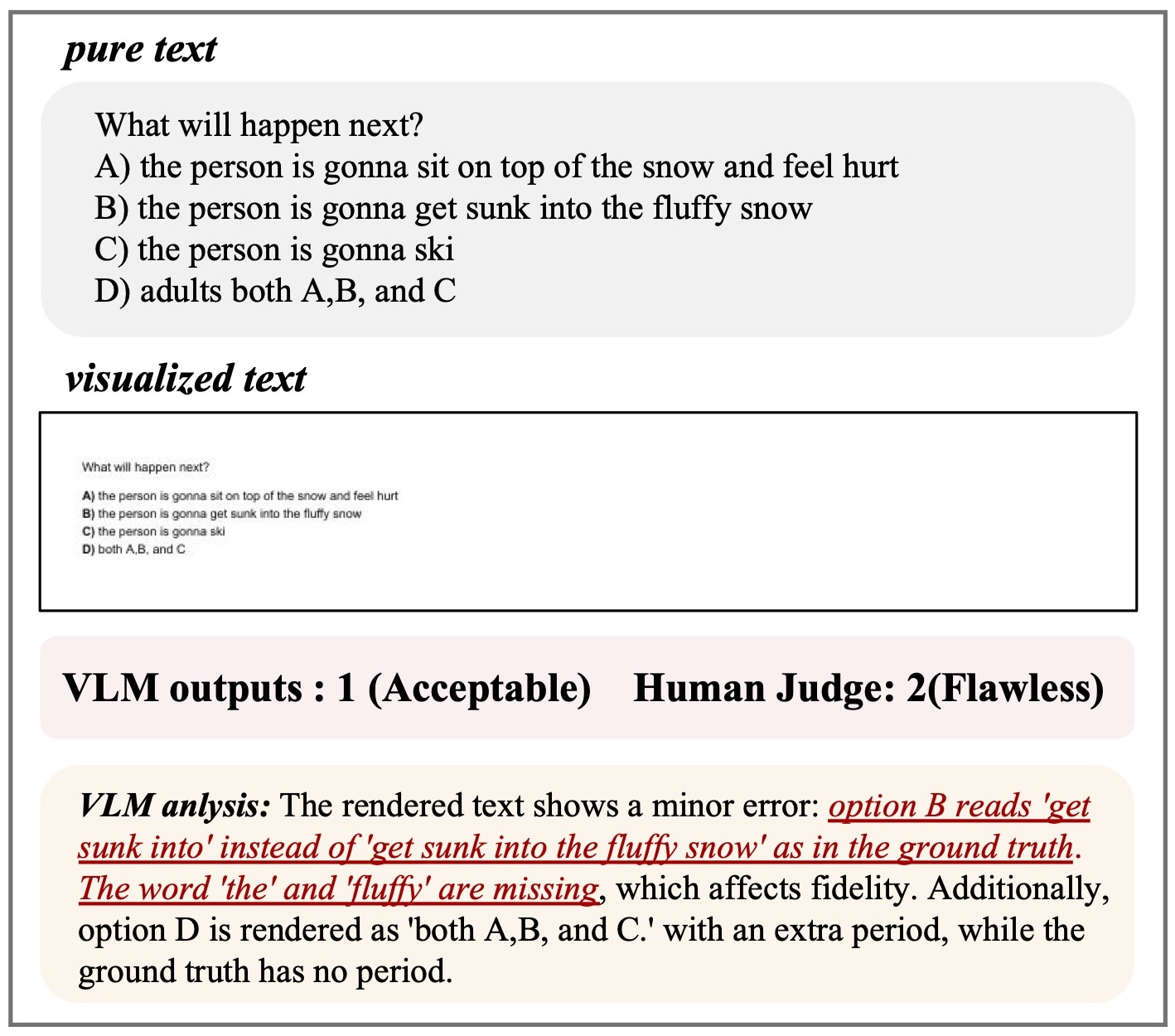}
    \caption{Character recognition error. Config: \textit{Arial}, 9pt}
    \label{fig:judge-case2}
\end{figure*}

We present representative cases where the VLM-as-Filter-Judge produces inaccurate quality assessments under different rendering conditions. These cases reveal that the judge can be overly strict and may assign below-perfect scores to otherwise flawless visualized-text samples. Such false-negative judgments may arise from recognition difficulties caused by small font sizes or dense visual layouts. These observations highlight the importance of the subsequent manual verification step, which prevents valid samples from being incorrectly filtered out by the VLM judge. 
\textbf{Samples are shown in Figure~\ref{fig:judge-case1} through Figure~\ref{fig:judge-case2} at the end.}
 
\clearpage
\section{Evaluation Protocol}

\subsection{Prompts Used in Evaluation}

We define a set of standardized prompts to evaluate VLMs under different levels of instruction granularity. Each prompt is designed to encourage direct answer extraction from visualized-text questions while reducing irrelevant or verbose output.

\begin{figure}[hbt!] 
  \centering
  \includegraphics[width=\textwidth]{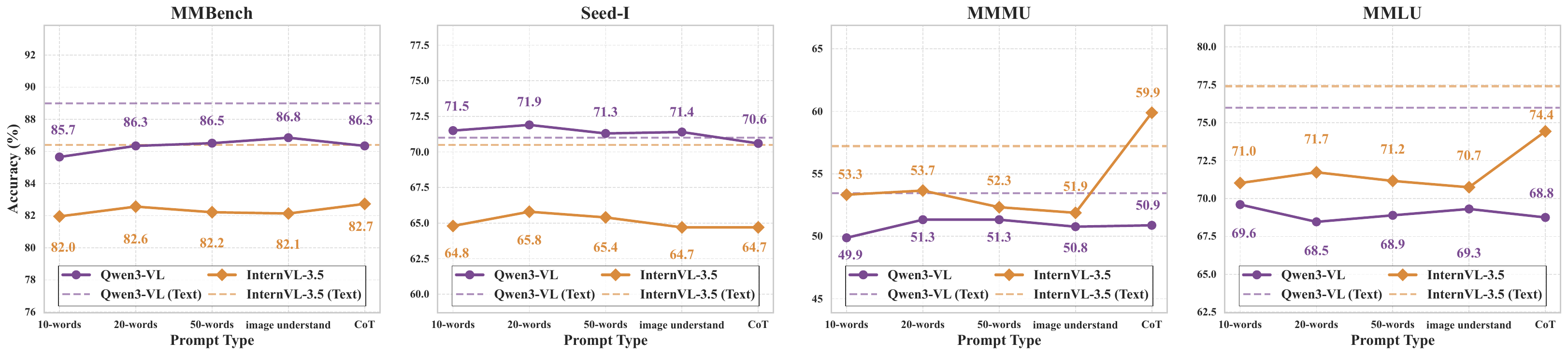}
  \vspace{-0.1in}
  \caption{\textbf{Impact of Prompt Design.} Prompt: 10-words, 20-words, 50-words, image understanding and CoT.}
  \label{fig:app_prompt_design}
\end{figure}

Prompts vary in length and reasoning emphasis, enabling an analysis of how instruction detail and presentation style impact model behavior, as illustrated below. All prompts are displayed in a uniform visual format using the \texttt{promptbox}.

\noindent $\bullet$ \textbf{10-words prompt.}
A minimal, highly concise instruction designed to provide only the essential guidance. Models are prompted to read the question and options and respond with the single letter corresponding to the correct answer. This prompt evaluates whether the model can correctly interpret visualized-text questions with minimal context.

\noindent $\bullet$ \textbf{20-words prompt.}
A slightly longer prompt that explicitly references the visual context of the question. This format encourages the model to carefully consider the visualized question while still producing a concise single-letter response.

\noindent $\bullet$ \textbf{50-words prompt.}
An expanded prompt that emphasizes careful comprehension of each option and the question context. This format tests whether providing more detailed textual guidance improves reasoning and answer accuracy when interpreting visualized-text questions.

\noindent $\bullet$ \textbf{Image-understand prompt.}
A detailed prompt focusing on full comprehension of the visualized-text question along with any accompanying problem image. It guides the model to interpret context and relationships before selecting an answer, without providing explanations, thereby testing multimodal understanding and attentional alignment.

\noindent $\bullet$ \textbf{CoT prompt.}
Designed to induce step-by-step internal reasoning, this prompt instructs the model to consider the visualized-text information sequentially but explicitly discard the reasoning process in the final output. Only the single-letter answer is required. This evaluates whether structured reasoning can improve accuracy without contaminating the response.

\begin{promptbox}{10-words prompt}
Read the question and options, then answer with only the single letter (e.g., A, B, C, D).
\end{promptbox}

\begin{promptbox}{20-words prompt (Original)}
Read the question and options shown in the image(s). Answer with only the single letter of the correct option (e.g., A, B, C, D).
\end{promptbox}

\begin{promptbox}{50-words prompt}
Please carefully read the question and options presented in the image(s). Ensure you understand the meaning of each option. Based on the question, choose the most appropriate answer and respond with only the letter of the correct option (e.g., A, B, C, or D). Do not include any additional text or explanations in your response.
\end{promptbox}

\begin{promptbox}{Image understand prompt}
Please take a moment to carefully read the question and all available options shown in the image(s). Ensure you fully understand the context and meaning behind each option. After considering the question and options, choose the most appropriate answer. Respond only with the letter corresponding to the correct option (e.g., A, B, C, or D). Do not include any explanations or comments in your response.
\end{promptbox}

\begin{promptbox}{CoT prompt}
Analyze the image to identify the question text and the available options. Think step-by-step to deduce the correct answer based on the visual information provided. However, \textbf{discard your reasoning process} in the final output. Your final response must consist of \textbf{nothing but} the single letter of the correct option (e.g., A). Do not explain why and do not use punctuation.
\end{promptbox}

\begin{table}[b]
\centering
\caption{Performance of Qwen3-VL-8B-Instruct under Pure Text baseline and Visualized Text inputs with different font settings. All results are reported as percentages (\%).}
\label{tab:qwen3_font_results}
\vspace{0.5em}
\small
\setlength{\tabcolsep}{4pt}
\renewcommand{\arraystretch}{1.08}
\begin{tabular}{lcccccc}
\toprule
\textbf{Benchmark} & \textbf{Pure Text} & \textbf{Font Size} & \textbf{Arial} & \textbf{Roman} & \textbf{Cambria} & \textbf{Brush} \\
\midrule

\multirow{2}{*}{MMMU}
& \multirow{2}{*}{53.4}
& 32pt & 53.9 & 52.8 & 52.0 & 53.4 \\
& 
& 48pt & 54.1 & 53.4 & 53.2 & 53.0 \\

\midrule

\multirow{2}{*}{MMBench}
& \multirow{2}{*}{89.0}
& 32pt & 86.3 & 86.6 & 86.2 & 86.4 \\
& 
& 48pt & 86.5 & 86.5 & 86.3 & 86.9 \\

\midrule

\multirow{2}{*}{Seed-Bench}
& \multirow{2}{*}{71.0}
& 32pt & 70.9 & 70.4 & 70.5 & 70.5 \\
& 
& 48pt & 71.0 & 70.8 & 71.4 & 71.0 \\

\midrule

\multirow{2}{*}{MMLU}
& \multirow{2}{*}{76.0}
& 32pt & 72.6 & 75.3 & 74.6 & 72.3 \\
& 
& 48pt & 74.3 & 73.4 & 73.4 & 74.0 \\

\bottomrule
\end{tabular}
\end{table}

\subsection{Additional Preliminary Results}

In Section~\ref{sec:preliminary_experiments}, we show that rendering choices affect visualized-text understanding: very small fonts hurt readability, 32--48pt renderings usually improve performance, and overly large text may saturate or slightly degrade due to layout changes. For font style, standard fonts such as \textit{Arial}, \textit{Times New Roman}, and \textit{Cambria} behave similarly, while the handwritten-style \textit{Brush Script MT} is generally more challenging. We therefore use 16pt as a diagnostic setting for the font-style study, since it remains readable while still exposing style-induced differences.

To further examine this observation, we take Qwen3-VL-8B-Instruct as a representative model and evaluate it under larger font sizes, including 32pt and 48pt. As shown in Table~\ref{tab:qwen3_font_results}, the differences among standard fonts remain relatively small at these larger sizes, while \textit{Brush Script MT} is still less stable due to its handwritten style. These results further support our main conclusion that rendering choices affect the magnitude of the visualized-text gap, but do not eliminate the gap itself.

\begin{table*}[!t]
\centering
\caption{Performance comparison of large VLMs under different fonts.}
\label{tab:font_ablation}
\vspace{0.5em}
\small
\setlength{\tabcolsep}{6pt}
\renewcommand{\arraystretch}{1.12}
\begin{tabular}{lcccccc}
\toprule
\textbf{Model} & \textbf{Benchmark} & \textbf{Pure Text} & \textbf{Arial} & \textbf{Cambria} & \textbf{Roman} & \textbf{Brush} \\
\midrule
Qwen3-VL-32B-Instruct & MMLU    & 82.9 & 78.6 & 77.3 & 76.3 & 72.3 \\
Qwen3.5-397B-A17B     & MMBench & 93.1 & 92.4 & 92.7 & 93.0 & 92.4 \\
\bottomrule
\end{tabular}
\end{table*}

\begin{table*}[!t]
\centering
\caption{Performance comparison of large VLMs under different sizes.}
\label{tab:size_ablation}
\vspace{0.5em}
\small
\setlength{\tabcolsep}{6pt}
\renewcommand{\arraystretch}{1.12}
\begin{tabular}{lccccccc}
\toprule
\textbf{Model} & \textbf{Benchmark} & \textbf{Pure Text} & \textbf{9pt} & \textbf{16pt} & \textbf{32pt} & \textbf{48pt} & \textbf{64pt} \\
\midrule
Qwen3-VL-32B-Instruct & MMBench & 90.5 & 84.6 & 88.4 & 89.4 & 89.7 & 89.7 \\
Qwen3.5-397B-A17B     & MMLU    & 90.1 & 90.1 & 90.6 & 91.1 & 90.9 & 88.2 \\
\bottomrule
\end{tabular}
\end{table*}

To further verify that the preliminary findings In Section~\ref{sec:preliminary_experiments} are not limited to the two 8B models, we conduct additional rendering ablations on larger models. We use MMBench as a representative multimodal benchmark and MMLU as a representative unimodal benchmark, covering the two major task paradigms studied in our preliminary analysis. As shown in Tables~\ref{tab:font_ablation} and~\ref{tab:size_ablation}, Qwen3-VL-32B-Instruct and Qwen3.5-397B-A17B still show sensitivity to rendering choices. For font size, very small text hurts Qwen3-VL-32B-Instruct on MMBench, while Qwen3.5-397B-A17B remains relatively strong on MMLU but still varies across sizes. For font style, standard fonts are generally stable, whereas \textit{Brush Script MT} causes a clearer drop for Qwen3-VL-32B-Instruct on MMLU. Notably, Qwen3.5-397B-A17B represents one of the largest open-source VLMs in our study. These results suggest that rendering sensitivity is not merely an artifact of small model capacity, but can also appear in stronger and larger models.

\subsection{Model Configurations}
We evaluate a broad suite of open-source VLMs implemented in VLMEvalKit. Unless otherwise specified, we follow the default model wrappers and decoding settings released by each official repository as integrated in VLMEvalKit and run all models in BF16 on NVIDIA A800 GPUs.

\paragraph{InternVL series.}
We evaluate InternVL3 and InternVL3.5 across three scales: InternVL3-2B, InternVL3-8B, InternVL3.5-2B, InternVL3.5-8B and the MoE model InternVL3.5-30B-A3B. All InternVL models are instantiated via the InternVLChat wrapper (version V2.0) with their corresponding official checkpoints and default generation settings (deterministic decoding by default, with a maximum of 4096 new tokens unless overridden by the dataset). The wrapper enforces a global image budget of 64 patches and dynamically sets the per-image patch budget based on the number of images; the default maximum number of patches is 6. We make two lightweight wrapper-level modifications for evaluation: (i) in prompt construction, we allow the wrapper to fall back to \textbf{question\_image\_path} when \textbf{image\_path} is absent, so that visualized-text question images are correctly loaded in our MM\_MCQ setting; and (ii) we handle the text-only edge case (e.g., MMLU) by setting the patch budget when the number of images is zero, avoiding invalid computation and ensuring consistent behavior across unimodal and multimodal evaluations.

\paragraph{Qwen-VL series.}
We evaluate Qwen2.5-VL and Qwen3-VL using their corresponding VLMEvalKit wrappers. For Qwen2.5-VL, we include Qwen2.5-VL-3B-Instruct and Qwen2.5-VL-7B-Instruct. We follow the standard configuration with custom prompts disabled and explicit vision-token constraints, using min\_pixels $=1280\times 28\times 28$ and max\_pixels $=16384\times 28\times 28$. For Qwen3-VL, we include Qwen3-VL-2B-Instruct, Qwen3-VL-8B-Instruct and Qwen3-VL-30B-A3B-Instruct, instantiated under the Instruct setting with custom prompts disabled. We use the default generation configuration in our setup (temperature $=0.7$, top-$p=0.8$, top-$k=20$, repetition\_penalty $=1.0$, presence\_penalty $=1.5$ and max\_new\_tokens $=16384$). All models are evaluated with the same protocol; when vLLM is enabled by the wrapper, the number of image inputs per prompt is capped at 24.

\paragraph{LLaVA series.}
We evaluate three representative LLaVA variants: LLaVA-1.5-7B, LLaVA-OneVision-7B and LLaVA-OneVision-1.5-8B. 
For LLaVA-1.5-7B, we do not rely on the original LLaVA codebase; instead, we use the official HuggingFace Transformers implementation (processor + LlavaForConditionalGeneration) to ensure a unified and reproducible inference pipeline. We follow deterministic decoding (no sampling) with a maximum generation length of 2048 tokens (temperature $=0$, single-beam). 
For LLaVA-OneVision, we use the VLMEvalKit wrapper based on the official LLaVA-OneVision implementation, keeping its default prompt template and decoding configuration. 
Since LLaVA-OneVision-1.5 is not natively supported in VLMEvalKit, we implement a lightweight wrapper following the evaluation procedure released by lmms-lab for OneVision-1.5, including the same chat-template construction and image-token alignment behavior in the processor. This implementation is a compatibility layer and does not modify model weights; therefore, it is not expected to introduce a material difference in the reported results beyond standard evaluation variance.

\paragraph{MiMo-VL series.}
We evaluate MiMo-VL-7B-SFT and MiMo-VL-7B-RL. Both models are instantiated via the same Qwen2VLChat wrapper and configuration as Qwen2.5-VL (custom prompts disabled, min\_pixels $=1280\times 28\times 28$, max\_pixels $=16384\times 28\times 28$), ensuring consistent evaluation settings.

\paragraph{Ovis series.}
We evaluate Ovis2-2B, Ovis2-8B, Ovis2.5-2B and Ovis2.5-9B using the official Ovis wrappers in VLMEvalKit. 
For Ovis2, we use the standard Transformers backend provided by the wrapper (AutoModelForCausalLM with trust\_remote\_code=True) and keep deterministic decoding with the default maximum generation length (1024 new tokens) and the model’s built-in preprocessing (preprocess\_inputs) under a fixed multimodal context length budget. The wrapper also sets the image partitioning strategy according to the dataset modality and the number of images to control memory usage.
For Ovis2.5, the original implementation is designed to run with a vLLM backend; however, we did not have an appropriate vLLM deployment for the released checkpoints in our environment. We therefore implement a Transformers-based inference path as a drop-in replacement, while preserving the core evaluation logic used by the vLLM version: (i) we keep the same prompt construction (including the fixed suffix for extracting the final answer), (ii) we use the model-provided preprocess\_inputs interface with the same min\_pixels/max\_pixels settings (448$\times$448 for OCR-style inputs and 1024$\times$1024 to 1792$\times$1792 otherwise) and (iii) we follow the same ``thinking'' enablement rules exposed by the wrapper for the relevant benchmark subsets. Since both backends ultimately invoke the same model weights and preprocessing pipeline with deterministic decoding, this change is intended to be a compatibility layer and is not expected to materially alter the reported results beyond standard inference variance.

\paragraph{SAIL-VL2 series.}
We evaluate SAIL-VL2-2B and SAIL-VL2-8B via the SailVL wrapper. The wrapper uses dynamic image preprocessing with a fixed vision input size (448) and a patch-based tiling strategy, where the maximum number of tiles is controlled by a per-dataset image budget (up to 10 by default and increased to 12/18/24 for text-dense benchmarks such as OCRBench and HRBench). The total patch budget is capped at 64 to avoid OOM. We follow deterministic decoding with the wrapper-default maximum generation length (1024 new tokens). In our setting, SAIL-VL2 is configured with MSAC enabled for single-image inputs (use\_msac=True) to improve coverage of text-dense renderings. For compatibility with our visualized-text question format, we apply the same wrapper-level I/O handling as in InternVLChat, ensuring that question-image fields are correctly loaded and that text-only edge cases are handled consistently.

\paragraph{NEO series.}
We evaluate NEO-2B-SFT and NEO-9B-SFT via the NEOChat wrapper. NEO uses a native patch-based image tokenizer with patch size 16 and a downsample ratio of 0.5, together with explicit pixel-range constraints (min\_pixels $=1280\times 32\times 32$, max\_pixels $=4096\times 32\times 32$) to control the visual token budget. The wrapper adopts deterministic decoding by default with a maximum generation length of 4096 new tokens and adjusts the per-example image budget using a global cap of 64 and dataset-specific maxima (6 by default, increased to 12/18/24 for text-dense benchmarks such as OCRBench and HRBench). For OCRBench, the wrapper further relaxes the minimum pixel threshold to 256$\times$256 to better accommodate small text regions. We apply the same wrapper-level I/O handling described for InternVLChat to ensure that visualized-text question-image fields are correctly loaded and that text-only edge cases are handled consistently.

\paragraph{Other models.}
We additionally evaluate DeepSeek-VL2-Tiny, GLM-4.1V-9B-Thinking, MiniCPM-V-4.5 and Kimi-VL-A3B-Thinking using their corresponding VLMEvalKit wrappers and default settings. For Kimi-VL-A3B-Thinking, we follow the wrapper configuration with temperature 0.8 and a maximum generation length of 32,768 tokens.

\paragraph{Remarks.}
For fair comparison, we do not hand-tune prompts or decoding per model beyond wrapper defaults and all models are evaluated under the same protocol. When a wrapper exposes explicit image-token constraints (e.g., minimum/maximum pixels), we keep the default values to match the intended usage of the official implementation.

\subsection{Answer Extraction and Post-processing}
We follow the VLMEvalKit evaluation pipeline for inference and prediction aggregation, while using an API-based answer extraction step for final scoring. 
For each instance, the model output is first stored as a raw string. 
Since exact string matching in VLMEvalKit can be unreliable for verbose responses, open-ended answers, or outputs containing explanations, we further process each prediction with a GPT-based extractor to obtain the final committed answer.

\textbf{Answer extraction.}
For multiple-choice questions, the extractor identifies the final option selected by the model and maps it to the corresponding option letter. 
The extraction is based only on the model's generated response and the candidate options, without re-solving the original question. 
This design reduces false matches caused by intermediate reasoning, repeated option mentions, or answers expressed in free-text form rather than as a single letter. 
For ordering-style or format-sensitive questions, the extractor additionally normalizes equivalent expressions and maps them to the required answer format when the model's final intent is unambiguous.

\textbf{Refusals and invalid outputs.}
If the model output contains explicit refusal, API failure messages, or no identifiable final answer, we mark it as invalid and count it as incorrect. 
Outputs whose final intent cannot be reliably mapped to any candidate option or reference answer are also treated as invalid. 
The extractor is instructed not to infer missing answers or correct the model's reasoning, but only to recover the answer explicitly implied by the model output.

\textbf{Result aggregation.}
We run inference with distributed workers and aggregate per-rank outputs into a single prediction file indexed by sample id. 
After GPT-based extraction, the normalized predictions are compared with the ground-truth answers to compute the final accuracy. 
When reasoning traces are present, we retain the raw output for analysis while using only the extracted final answer for scoring.

\clearpage
\section{Additional Experiments}

\subsection{Rendering Factor Ablations}
\label{app:rendering_factor_ablations}

\begin{figure}[!t]
  \centering

  \begin{minipage}{0.49\linewidth}
    \centering
    \includegraphics[width=\linewidth]{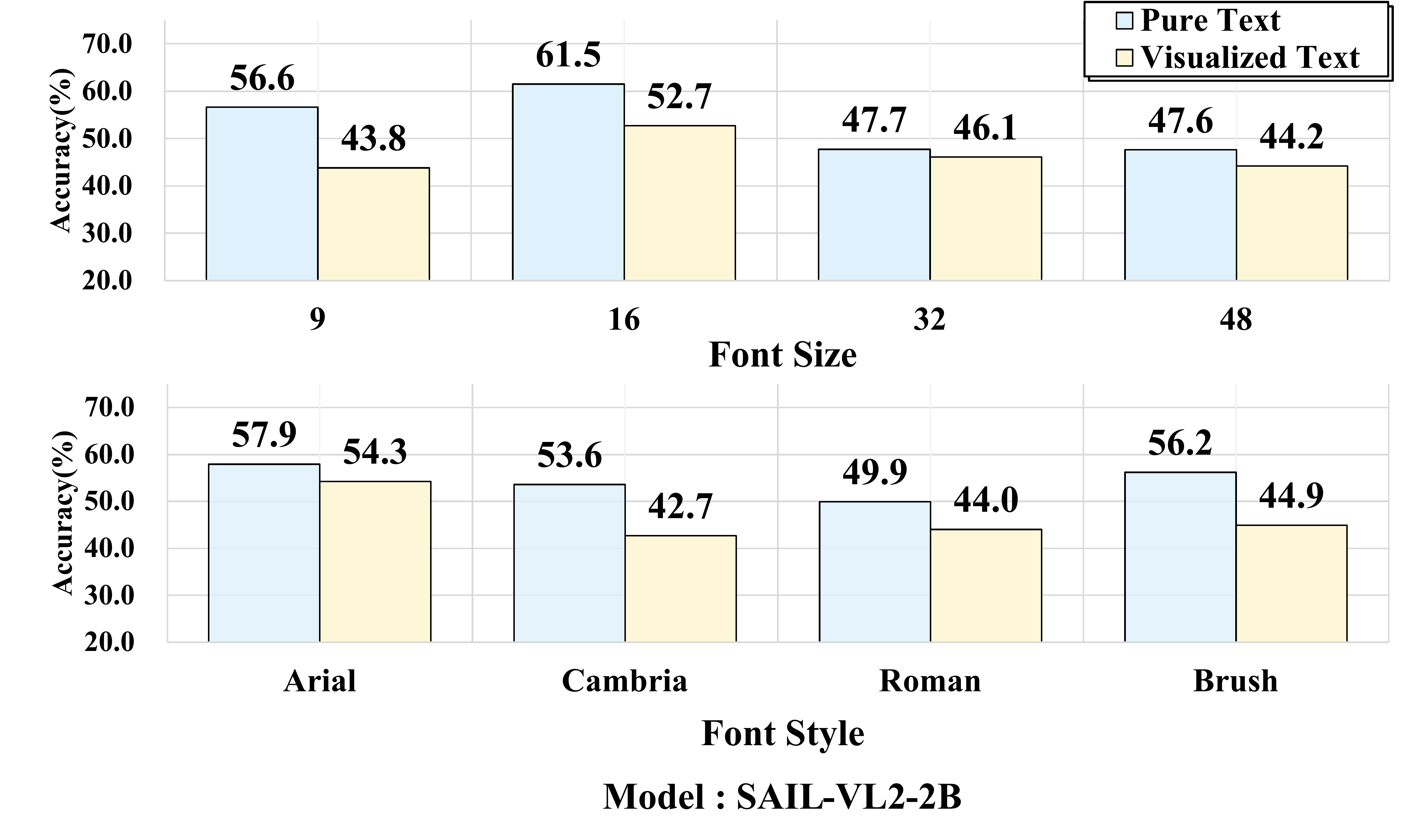}
  \end{minipage}\hfill
  \begin{minipage}{0.49\linewidth}
    \centering
    \includegraphics[width=\linewidth]{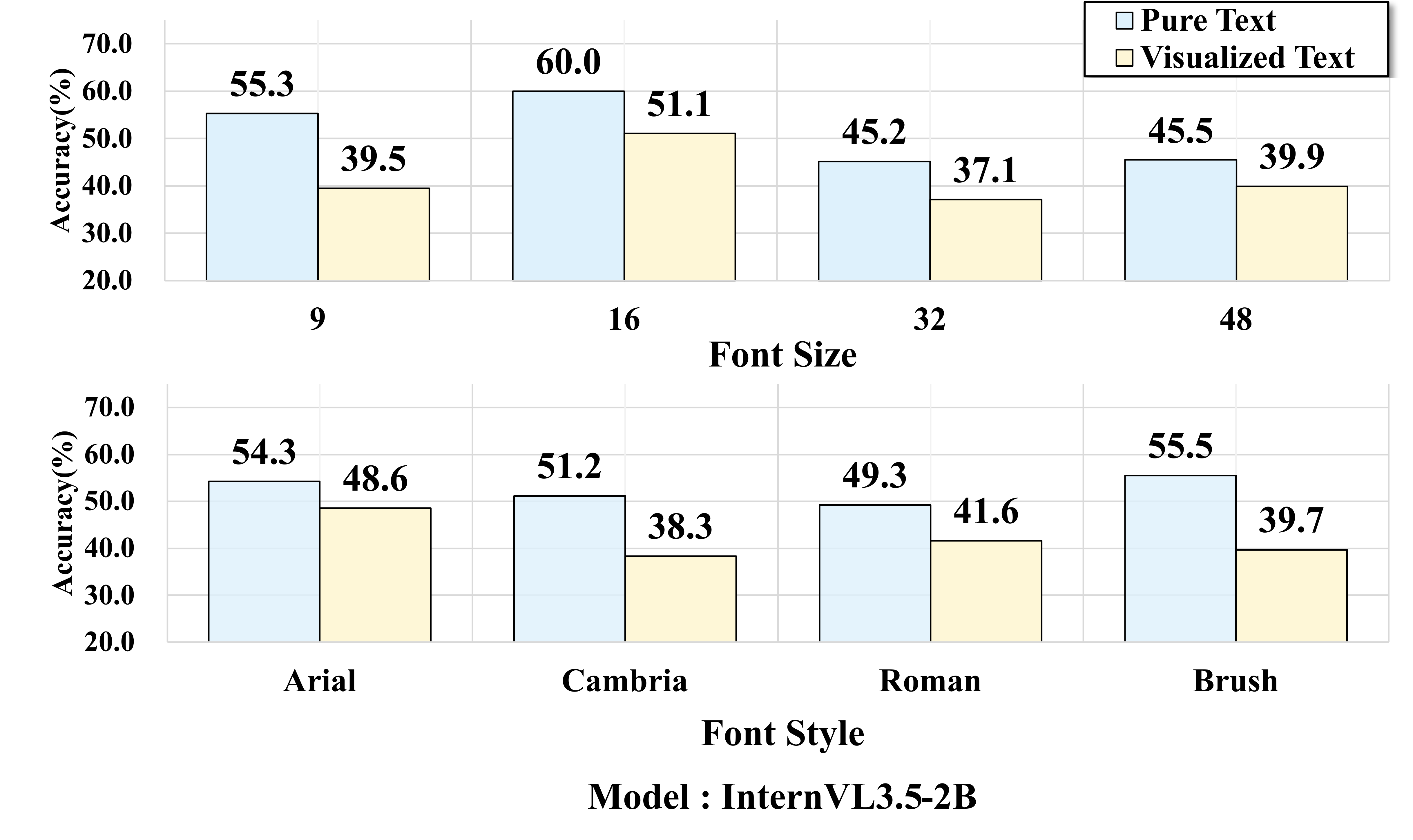}
  \end{minipage}

  \vspace{0.6em}
  \begin{minipage}{0.49\linewidth}
    \centering
    \includegraphics[width=\linewidth]{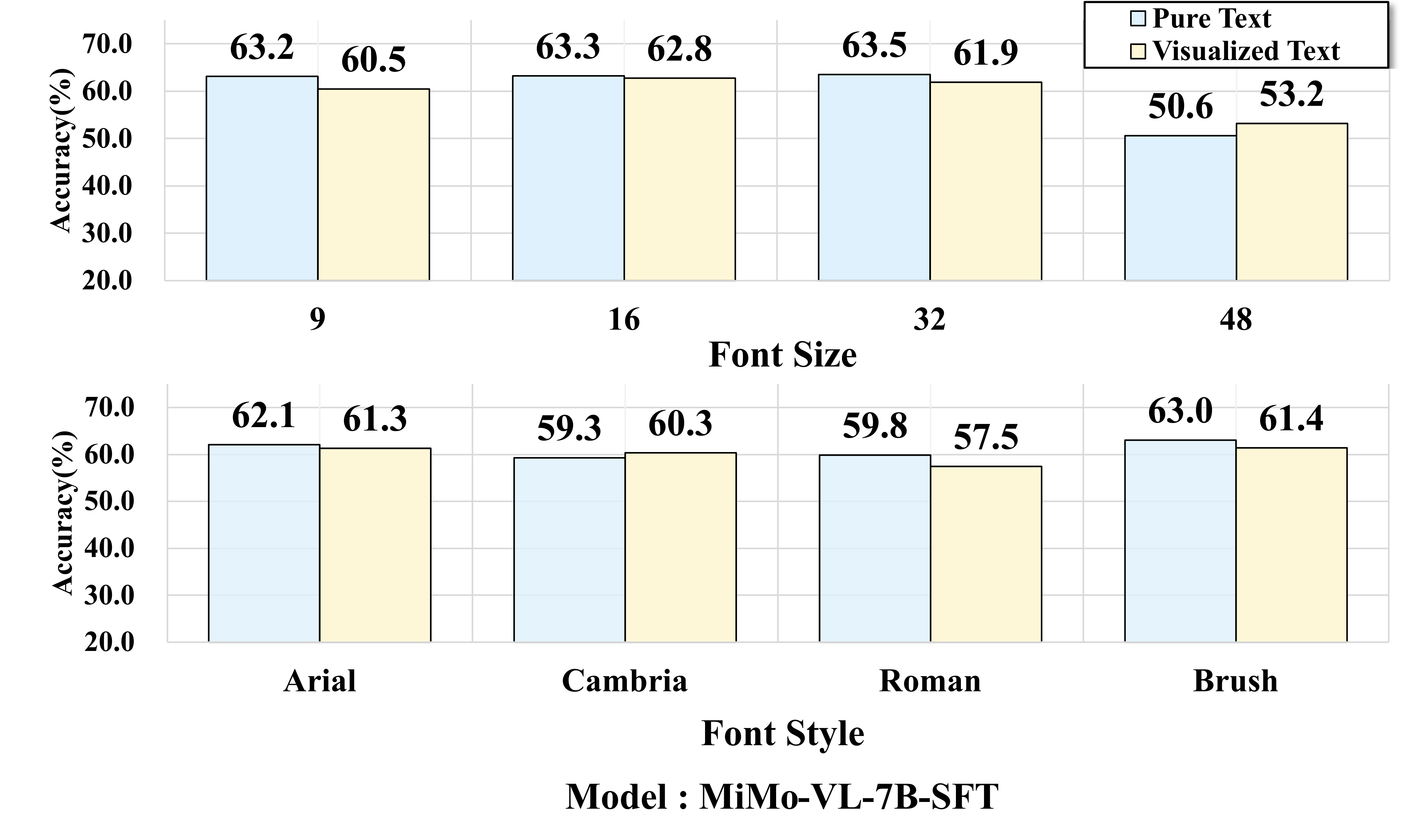}
  \end{minipage}\hfill
  \begin{minipage}{0.49\linewidth}
    \centering
    \includegraphics[width=\linewidth]{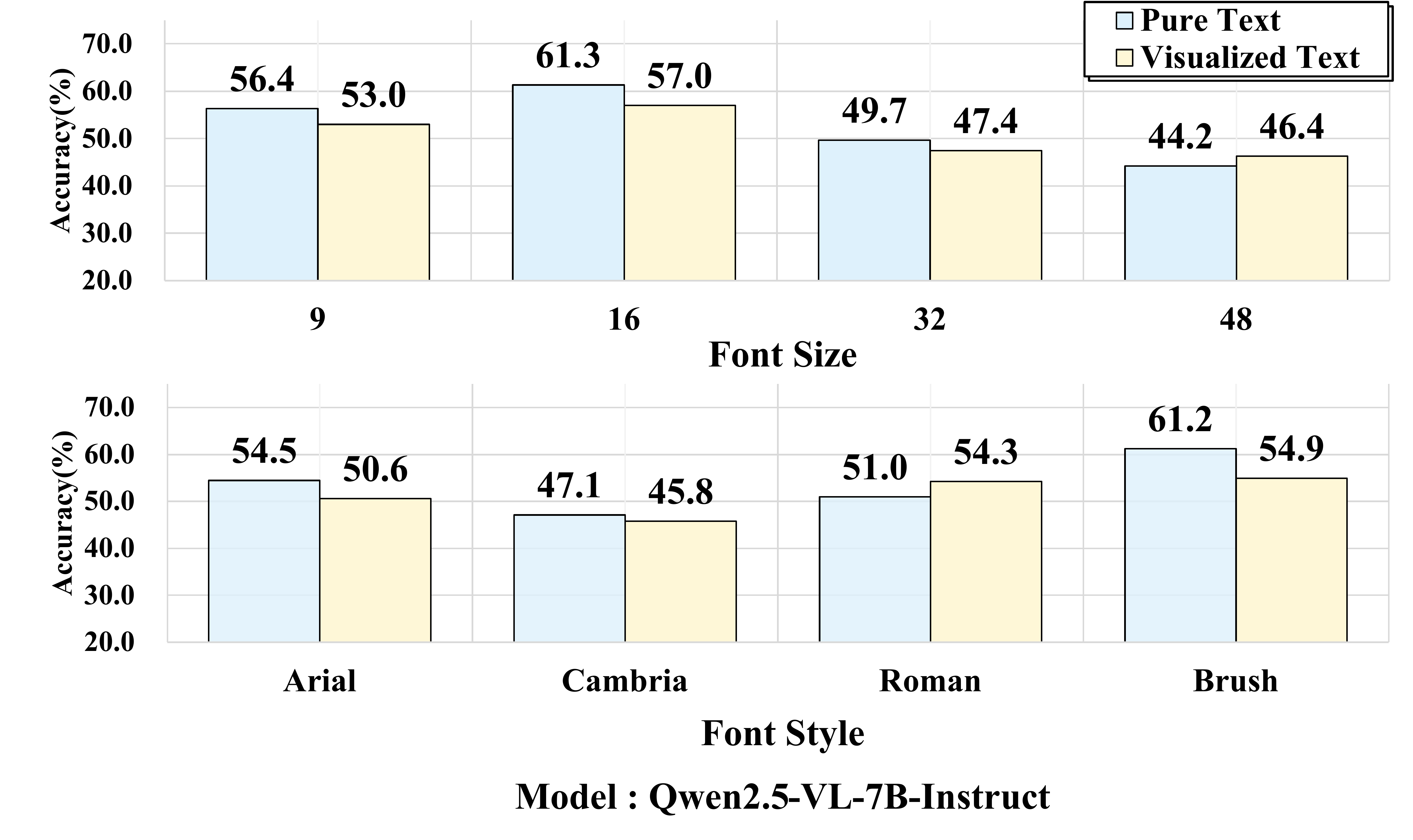}
  \end{minipage}

  \vspace{0.6em}
  \begin{minipage}{0.49\linewidth}
    \centering
    \includegraphics[width=\linewidth]{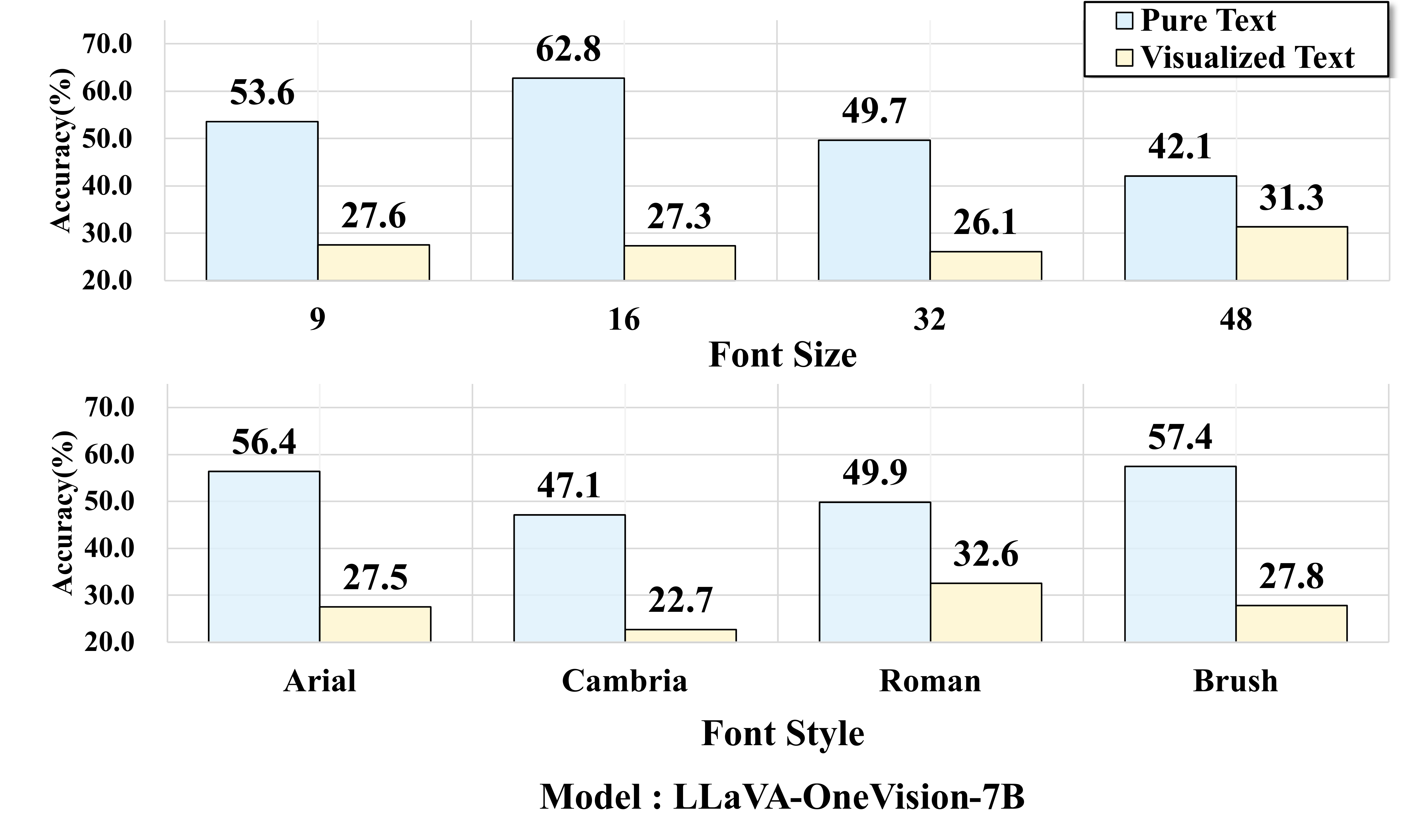}
  \end{minipage}\hfill
  \begin{minipage}{0.49\linewidth}
    \centering
    \includegraphics[width=\linewidth]{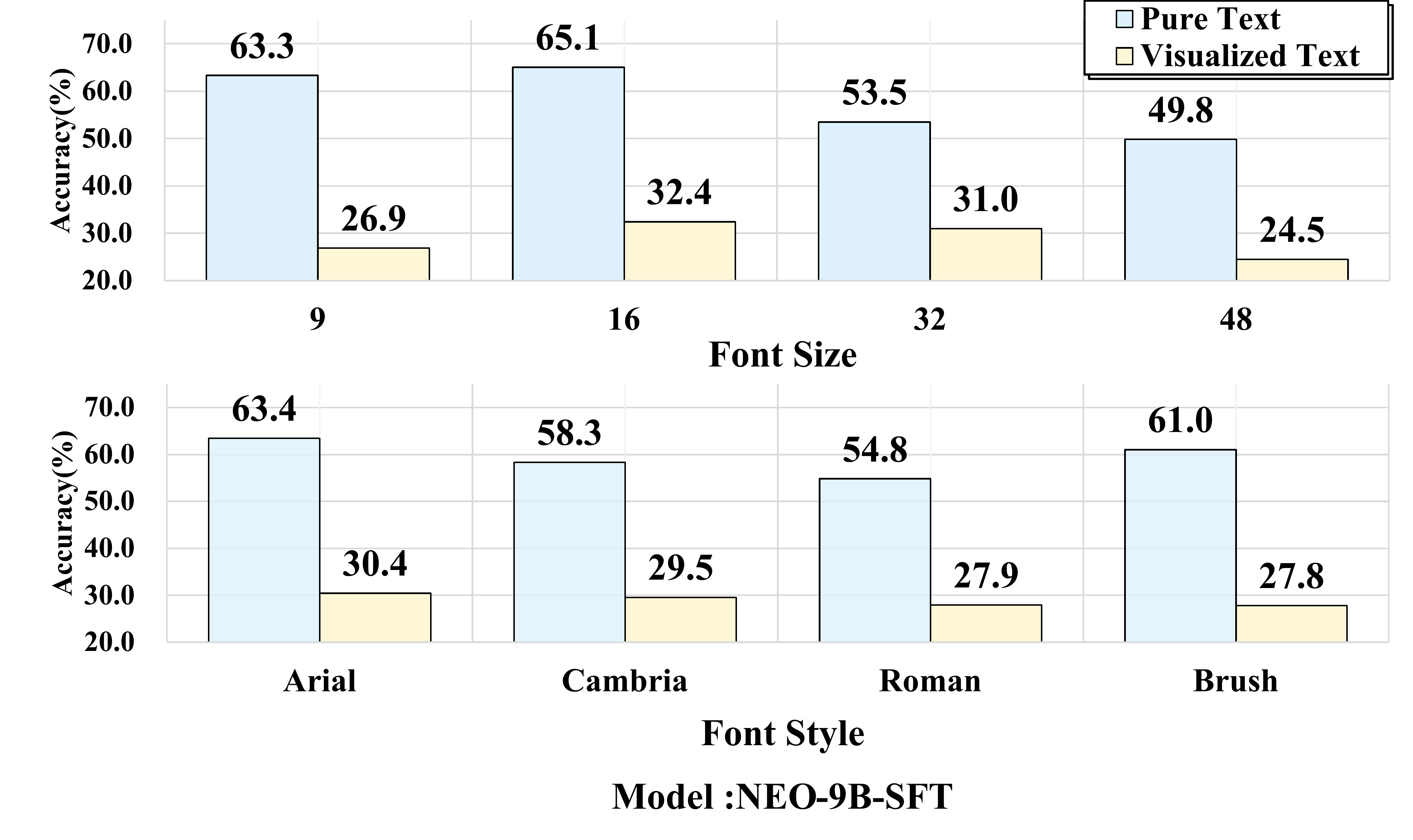}
  \end{minipage}

  \vspace{0.6em}
  \begin{minipage}{0.49\linewidth}
    \centering
    \includegraphics[width=\linewidth]{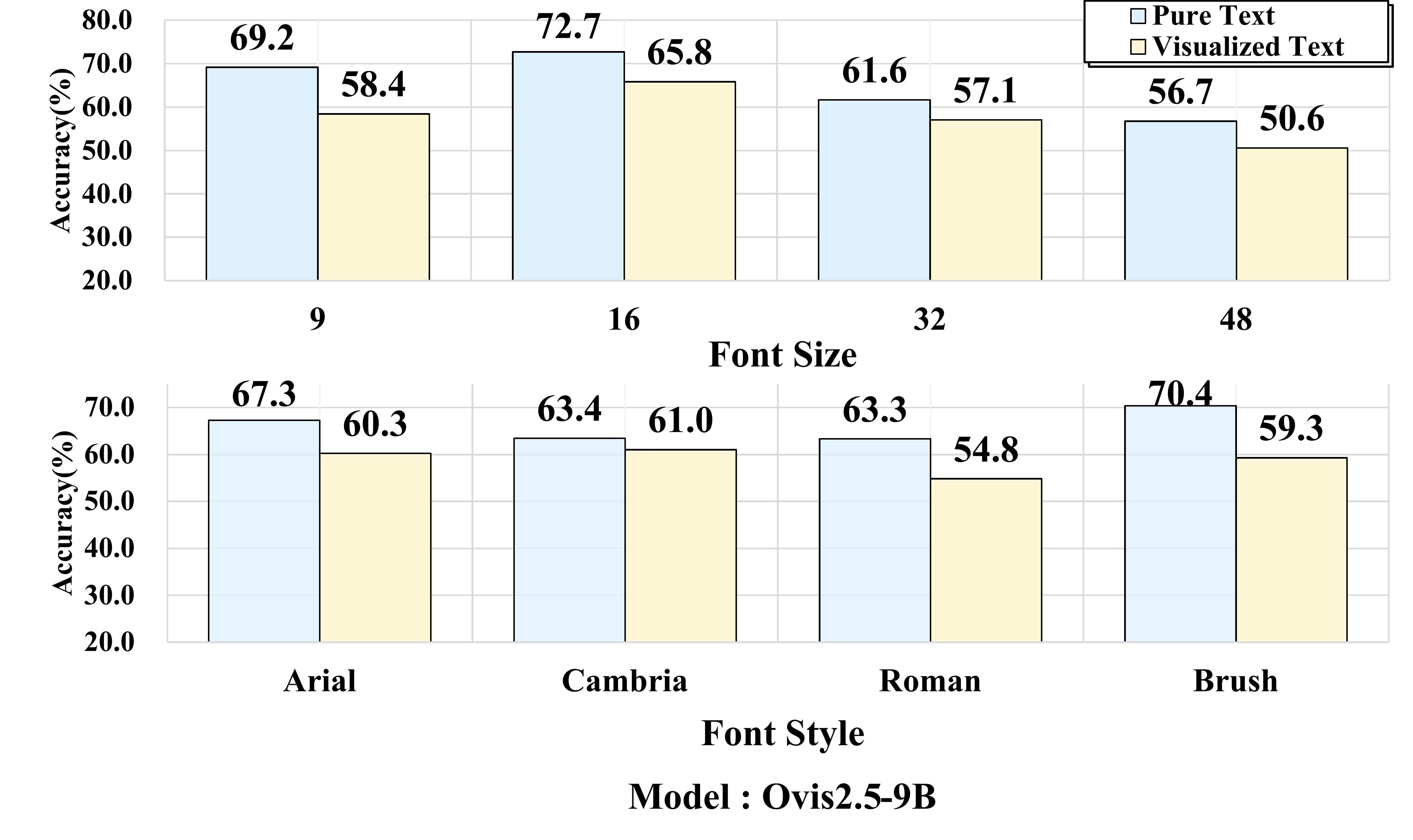}
  \end{minipage}\hfill
  \begin{minipage}{0.49\linewidth}
    \centering
    \includegraphics[width=\linewidth]{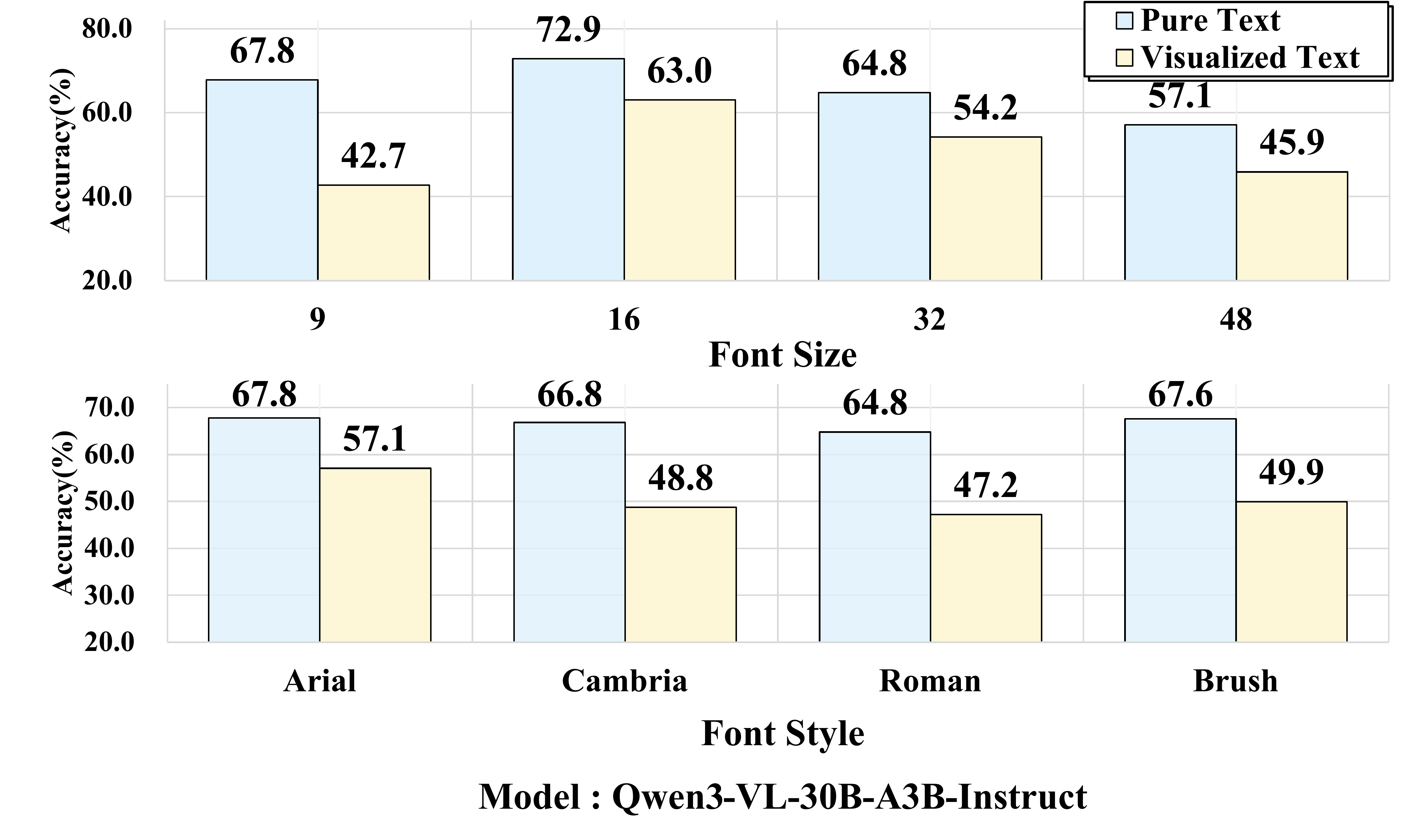}
  \end{minipage}

  \caption{Rendering sensitivity study on eight additional representative models.}
  \label{fig:Rendering Factor Ablations}
\end{figure}

We extend the fine-grained rendering sensitivity study from GLM-4.1V-9B-Thinking to eight additional representative models, as shown in Figure~\ref{fig:Rendering Factor Ablations}.
For each model, we evaluate performance under Pure Text and Visualized Text while varying font size (9pt, 16pt, 32pt, 48pt) and font style (\textit{Arial, Cambria, Roman, Brush}) and report the modality gap as $\mathrm{Acc}_{\text{Text}} - \mathrm{Acc}_{\text{VT}}$.
Two consistent patterns emerge. 
First, perceptually difficult renderings, especially very small font sizes and Brush-style text, lead to markedly larger gaps across most models, indicating persistent vulnerability to reduced legibility and rendering artifacts. 
Second, cleaner renderings substantially reduce the gap for many models and a small subset even becomes nearly gap-free or shows slight gains under Visualized Text; however, several models still exhibit large residual gaps under standard settings, suggesting limitations beyond perceptual readability.

\begin{figure}[b]
  \centering
  \includegraphics[width=0.6\columnwidth]{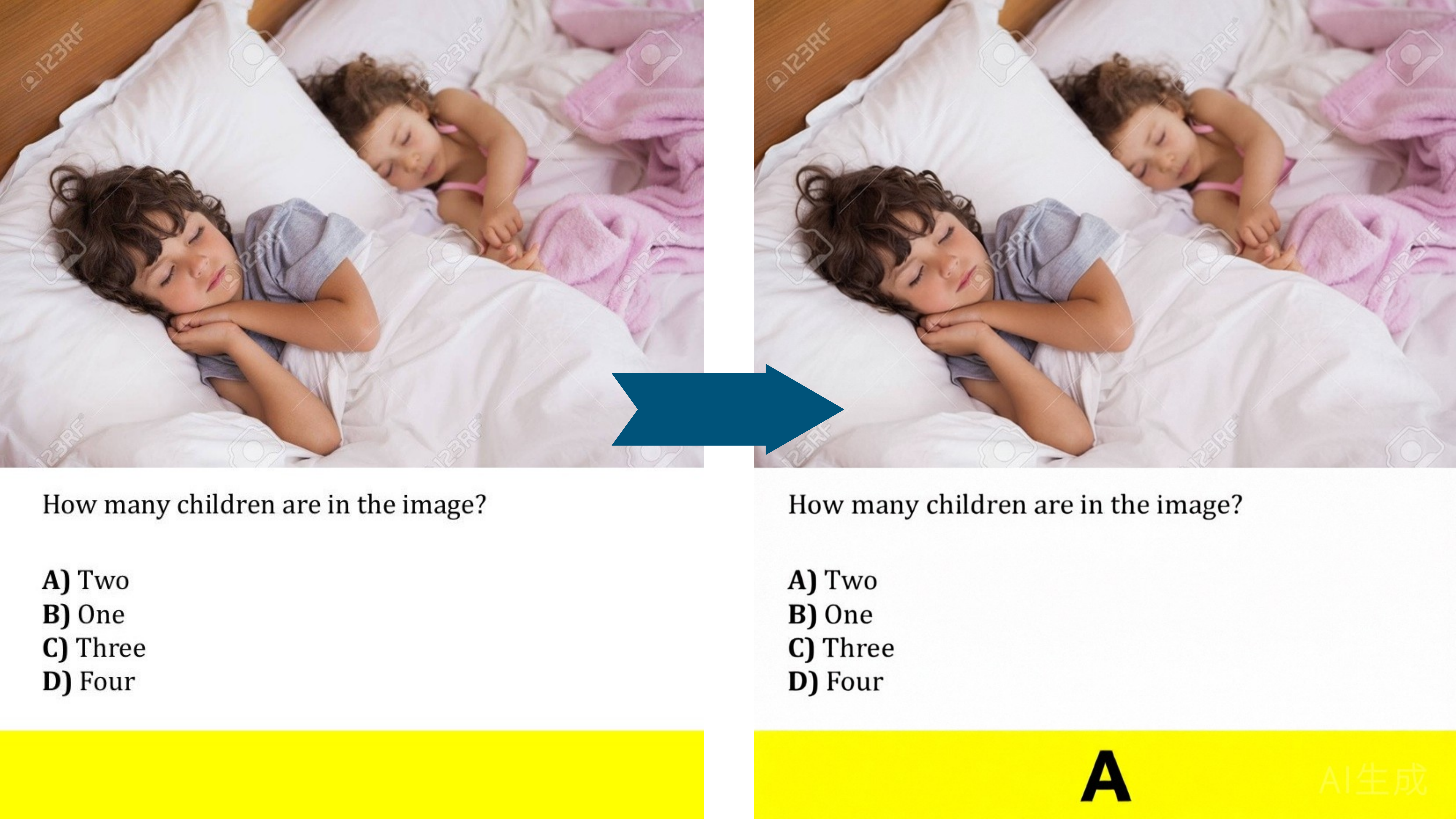}
  \caption{\textbf{A successful Qwen-Image-Edit case under the visualized-text setting.}
  The model correctly generates readable visualized text in the designated region and produces the correct answer.}
  \label{fig:qwenimage_case}
  \vspace{-0.3em}
\end{figure}

\subsection{Qwen-Image-Edit Evaluation}
\label{app:qwen_image_edit_eval}

To examine whether generative multimodal models can be evaluated under the visualized-text paradigm, we conduct a preliminary study with Qwen-Image-Edit on a 200-instance subset of VISTA-Bench. Specifically, we randomly sample 50 instances from each of the four tasks (multimodal tasks and unimodal task), resulting in a balanced evaluation set. In this setting, both the question and options are rendered as visualized text and, together with the problem image, are provided as a unified visual input. The model is instructed to write its final answer directly into a designated region of the output image.

Due to the open-ended and image-based nature of the generation process, all outputs are assessed via \textbf{human evaluation}. For each generated result, annotators verify \textbf{\textit{(i)} generation validity:} whether a readable question/answer is produced in the specified region and \textbf{\textit{(ii)} answer correctness:} whether the final answer is correct. This protocol enables reliable judgment of both layout-controlled generation and task performance, which is not robustly captured by automatic string matching.

Across 200 evaluated instances, 149 yield valid generations in which the model produces readable question/option text in the designated region and outputs a final answer (see Figure~\ref{fig:qwenimage_case} for a representative example), while 50 are labeled as \textit{no question} due to failures to generate a readable question or answer; one additional sample shows minor encoding artifacts and is excluded. Considering all 200 samples, Qwen-Image-Edit achieves an overall accuracy of 22.5\%. When the evaluation is restricted to the 149 samples that are generated according to the predefined rules, the accuracy remains similar at 22.15\%. The small difference suggests that end-to-end performance is limited not only by generation stability under constrained layouts, but also by the difficulty of correctly extracting semantics and reasoning from pixel-level text even when the visualized text is successfully produced.

Overall, these results indicate that generative multimodal models such as Qwen-Image-Edit can naturally interface with VISTA-Bench and be evaluated in an end-to-end visualized-text setting. However, their performance is jointly constrained by generation robustness and downstream visualized-text understanding. More broadly, this experiment suggests that visualized text introduces challenges beyond conventional OCR or text-conditioned generation, further motivating VISTA-Bench as a meaningful testbed for studying vision-centric language understanding in next-generation multimodal systems.


\clearpage
\onecolumn 

\makeatletter
\setlength{\@fptop}{0pt}          
\setlength{\@fpsep}{12pt}         
\setlength{\@fpbot}{0pt plus 1fil} 
\makeatother

\clearpage 

\clearpage

\clearpage

\clearpage

\makeatletter
\setlength{\@fptop}{0pt plus 1fil} 
\setlength{\@fpbot}{0pt plus 1fil}
\makeatother

\makeatletter
\setlength{\@fptop}{10pt}    
\setlength{\@fpbot}{0pt plus 1fil} 
\makeatother

\clearpage

\clearpage

\clearpage


\end{document}